\documentclass[10pt,journal,compsoc]{IEEEtran}
\ifCLASSOPTIONcompsoc
  \usepackage[nocompress]{cite}
\else
  \usepackage{cite}
\fi
\usepackage[table,svgnames]{xcolor}
\usepackage[top=0.4in,bottom=0.42in,left=0.4in,right=0.4in]{geometry}

\ifCLASSINFOpdf
\else
\fi

\usepackage{amsmath,amssymb,amsfonts}
\usepackage[ruled,linesnumbered]{algorithm2e}
\usepackage{array}
\usepackage{pifont}
\usepackage{textcomp}
\usepackage{stfloats}
\usepackage{url}
\usepackage{graphicx}
\usepackage{utfsym}
\usepackage{float}
\usepackage{overpic}
\usepackage{cancel}
\usepackage{threeparttable}
\usepackage{colortbl}

\usepackage{booktabs}
\usepackage{multirow}
\usepackage{makecell}
\usepackage{arydshln}
\usepackage[labelformat=simple]{subcaption}
\usepackage[colorlinks,linkcolor=red,citecolor=green]{hyperref}

\makeatletter
\newcommand{\thickhline}{%

    \noalign {\ifnum 0=`}\fi \hrule height 1pt
    \futurelet \reserved@a \@xhline
}
\newcolumntype{I}{!{\vrule width 1pt}}

\begin{document}
\title{RPCANet$^{++}$: Deep Interpretable Robust PCA \\for Sparse Object Segmentation}

\author{Fengyi~Wu, Yimian~Dai, Tianfang~Zhang, Yixuan~Ding, Jian~Yang, Ming-Ming Cheng, Zhenming~Peng
\IEEEcompsocitemizethanks{
\IEEEcompsocthanksitem F. Wu, Y. Ding, and Z. Peng are with the School of Information and Communication Engineering and the Laboratory of Imaging Detection and Intelligent Perception, University of Electronic Science and Technology of China, Chengdu, China. (E-mail: wufengyi98$@$163.com; 2840046d$@$student.gla.ac.uk; zmpeng$@$uestc.edu.cn).

\IEEEcompsocthanksitem Y. Dai, M. Cheng, and J. Yang are with PCA Lab, VCIP, College of Computer Science, Nankai University, Tianjin 300350, China. (e-mail: yimian.dai$@$gmail.com;
cmm$@$nankai.edu.cn; csjyang$@$nankai.edu.cn).

\IEEEcompsocthanksitem Tianfang Zhang is with the Department of Automation, Tsinghua University, Beijing, China. (e-mail: sparkcarleton$@$gmail.com).

\IEEEcompsocthanksitem This research was supported by NSFC (
No. 61775030, No.61571096,
No. 62301261, 
No. U24A20330, No. 62361166670, 
No. 62225604),
Shenzhen Science and Technology Program (JCYJ20240813114237048), 
and Natural Science Foundation of Sichuan Province of China (Grant No. 2025ZNSFSC0522).
}
\thanks{Manuscript submitted at June, 2025.}
}

\IEEEtitleabstractindextext{%
\begin{abstract}
Robust principal component analysis (RPCA) decomposes an observation matrix into low-rank background and sparse object components. This capability has enabled its application in tasks ranging from image restoration to segmentation. However, traditional RPCA models suffer from computational burdens caused by matrix operations, reliance on finely tuned hyperparameters, and rigid priors that limit adaptability in dynamic scenarios. To solve these limitations, we propose RPCANet\textsuperscript{++}, a sparse object segmentation framework that fuses the interpretability of RPCA with efficient deep architectures. Our approach unfolds a relaxed RPCA model into a structured network comprising a Background Approximation Module (BAM), an Object Extraction Module (OEM), and an Image Restoration Module (IRM). To mitigate inter-stage transmission loss in the BAM, we introduce a Memory-Augmented Module (MAM) to enhance background feature preservation, while a Deep Contrast Prior Module (DCPM) leverages saliency cues to expedite object extraction. Extensive experiments on diverse datasets demonstrate that RPCANet\textsuperscript{++} achieves state-of-the-art performance under various imaging scenarios. We further improve interpretability via visual and numerical low-rankness and sparsity measurements. By combining the theoretical strengths of RPCA with the efficiency of deep networks, our approach sets a new baseline for reliable and interpretable sparse object segmentation. Codes are available at our \href{https://fengyiwu98.github.io/rpcanetx}{Project Webpage}.
\end{abstract}

}

\maketitle

\IEEEdisplaynontitleabstractindextext

%
\IEEEpeerreviewmaketitle

\ifCLASSOPTIONcompsoc
\IEEEraisesectionheading{\section{Introduction}\label{sec:introduction}}
\else
\label{sec:introduction}
\fi
Over the past decades, robust principal component analysis (RPCA), as an extension of PCA, has received extensive attention due to its robust representation of outliers. An observed matrix $\mathbf{D}$ comprises a low-rank matrix with redundant features and a sparse matrix with distinct objects. Such decomposition structure benefits massive research fields \cite{bouwmans-2018-rpca} such as low-level vision tasks like image restoration \cite{ji-2011-robust} and denoising \cite{zhao-2014-hyperspectral} or high-level vision tasks such as fore/background subtraction \cite{liu-2015-background}, image classification \cite{zhang2011image}, and is particularly useful for various practical image segmentation tasks (e.g., defect detection, vessel subtraction, and infrared small target segmentation): by regrading the background with redundant information as $\mathbf{B}$ and segmented object as the sparse components $\mathbf{O}$ as:
\begin{equation}
    \mathbf{D} = \mathbf{B} + \mathbf{O}
\label{eq_org}
\end{equation}

Such models are formulated as either convex \cite{wright-2009-robust,zhou-2010-pcp} or non-convex \cite{netrapalli2014non,zhang-2018-nram,wang2021surface} optimization problems, typically solved using augmented Lagrangian methods \cite{candes-2011-rpca}, proximal gradient descent \cite{yi-2016-fast}, or alternating direction minimization \cite{yuan-2009-sparse}. To enhance image segmentation performance, these models incorporate various constraints, including object constraints \cite{cui2012groupsparse,peng2016salient,zhang-2019-nolc}, background constraints \cite{zhu2020tnlrs,xie2016weighted,xia2019vessel}, and field priors \cite{shen2012unified,zhou2012moving,zhu2019infrared}, among others. However, despite their theoretical appeal, existing models face two key limitations:

\begin{figure*}[!t]
\setlength{\abovecaptionskip}{0.1cm}
\setlength{\belowcaptionskip}{0cm}
\centering
	\includegraphics[width=0.95\linewidth]{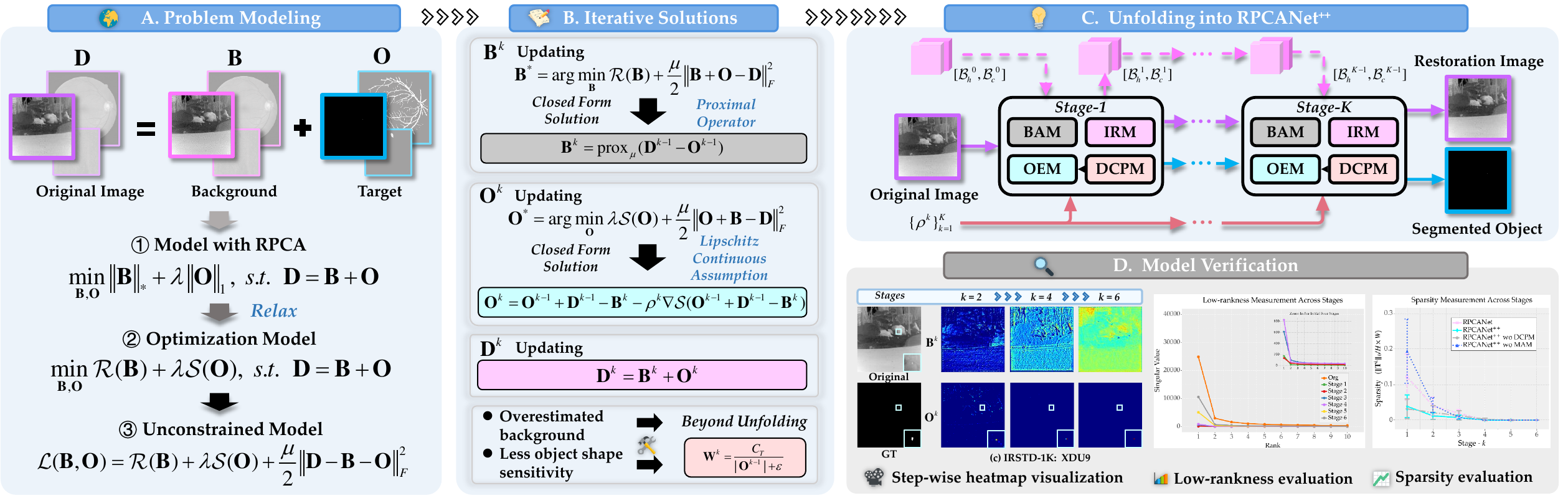}
   \caption{\textbf{Overview} of the proposed \textbf{RPCANet$^{++}$} architecture. \textbf{A}. Model the given image within a relaxed RPCA scheme and transform it into an unconstrained optimization problem. \textbf{B}. Iteratively solves the model above with closed-form solutions; Consider two high-level issues with corresponding solutions. \textbf{C}. Unfold the solutions in a deep unfolding framework; typically, RPCANet$^{++}$ are assisted with memory-augmented modules and deep target priors. \textbf{D}. Visual and numerical model verifications via post-hoc techniques present overall interpretability.}
\label{fig:short}
\vspace{-0.3cm}
\end{figure*}

$\bullet$ \textbf{Computational cost to convergence:} Repeated use of costly matrix or tensor operations slows convergence and hinders real-time deployment, especially in memory-constrained settings.

$\bullet$ \textbf{Limited generalizability:} 
(1) Heavy reliance on manually tuned hyperparameters leads to performance drops across diverse scenarios;  
(2) Rigid priors restrict adaptability to different domains, limiting broader applicability.

Recent progress in deep neural networks (DNNs) and the availability of large-scale public datasets have advanced object segmentation with greater adaptability to emerging data. Architectures such as FPNs \cite{lin-2017-fpn}, U-Nets \cite{ronneberger-2015-unet}, Transformers \cite{vaswani-2017-transformer}, and segment anything model \cite{kirillov2023sam} are widely adopted. However, their empirical designs often compromise interpretability, rendering them “black boxes” whose outputs may be unreliable without domain-specific context.

To address the interpretability challenges of DNNs, deep unfolding networks (DUNs) have emerged as a distinctive class of model-based approaches. By unrolling iterative optimization algorithms into structured deep networks, DUNs have recently found applications across diverse inverse problems, including compressive sensing \cite{sun-2016-admmnet,shen2025hunet}, as well as image denoising \cite{ren-2021-denoise,zheng2021deep}, super-resolution \cite{zhang-2020-dunsuper,zhou2023memory}, and image restoration \cite{kong-2022-restoration, fu-2024-restoration}.

Despite growing interest in deep unfolded RPCA models \cite{cai-2021-lrpca}, their application to segmentation remains limited. Adapting optimization-based frameworks to high-level vision tasks poses several challenges—particularly in handling key matrix operations such as singular value decomposition (SVD) and soft-thresholding (ST) under dynamic conditions. While prior studies have proposed variants of SVD/ST and improved initialization strategies \cite{jou-2024-rpca,cai-2021-lrpca}, efficiently managing complex matrix computations and selecting learnable parameters remains an open issue \cite{zhou-2023-lrsp}. Furthermore, although emerging datasets offer ground truth for sparse components, effective supervision of the intermediate low-rank component remains largely unsolved.

Furthermore, RPCA can be viewed as a decoupling task, where improving low-rank background estimation enhances object detection, and vice versa. However, in DUNs, this reciprocal benefit is often weakened by stage-wise transmission loss \cite{zhang-2023-physics}, leading to background misestimation and object omission. While inter-stage operations such as summation or concatenation help mitigate these issues \cite{zhang-2023-ctnet}, they remain insufficient for adaptively preserving background features. In addition, unlike general restoration tasks, segmentation-focused optimization methods typically introduce saliency priors \cite{dai-2017-ript} to highlight object regions and speed up convergence. Inspired by such strategies, we propose incorporating a domain factor to guide convergence and enhance segmentation performance.

Motivated by the challenges discussed above, we propose RPCANet$^{++}$—an interpretable segmentation framework that bridges deep unfolding with high-level vision operations. As shown in Fig.~\ref{fig:short}, RPCANet$^{++}$ unfolds a relaxed RPCA model into three modules: Object Extraction (OEM), Background Approximation (BAM), and Image Restoration (IRM). Instead of relying on costly matrix computations and handcrafted parameters, we employ theoretically constrained neural networks, treating the low-rank background as a latent element. Through object-background merging, the task jointly performs sparse object extraction and image restoration. Unlike conventional segmentation frameworks, RPCANet$^{++}$ avoids feature loss from repeated downsampling and adapts well to diverse domains. To address feature degradation across stages, we introduce a Memory-Augmented Module (MAM) that reinforces background information adaptively. In parallel, we design a Deep Contrast Prior Module (DCPM), inspired by reweighted optimization, to guide object allocation. To verify that our model adheres to RPCA principles, we employ stage-wise low-rankness and sparsity metrics (Fig.~\ref{fig:short}) to evaluate interpretability. RPCANet$^{++}$ thus integrates optimization-driven transparency with high-level learning flexibility, delivering reliable and generalizable segmentation. The main contributions are summarized as follows:

\begin{enumerate}[]
\item We propose RPCANet$^{++}$, a novel sparse object segmentation model that unfolds the traditional RPCA framework into a deep network, combining the interpretability of model-driven methods with the generalizability of data-driven learning.
\item We design a background approximation module using nonlinear proximal networks to estimate background content without costly matrix operations, supported by a cross-stage memory augmentation mechanism.
\item Sparse objects are extracted through a Lipschitz-constrained neural representation, complemented by an adaptive local contrast prior inspired by traditional saliency cues to enhance segmentation accuracy.
\item Beyond segmentation, we integrate low-rank background and sparse components to complete an image restoration task, leveraging the robustness of the decomposition-based structure.
\item We validate RPCANet$^{++}$ via quantitative metrics and interpretability analyses. Experiments across diverse datasets demonstrate strong generalization and consistent outperformance over state-of-the-art methods.
\end{enumerate}

A precursor to this work was presented at a conference \cite{wu-2024-rpcanet}. This article extends that foundation with substantial advancements: 1) To solve the background transmission loss, we introduce a memory-augmented module in the BAM to enhance the feature restoration. 2) Besides, inspired by prior assistants in mainstream optimization methods, we design a saliency-inspired prior DCPM to accelerate object extraction; 3) We provide a more profound analysis of RPCANet$^{++}$ with novel metrics like the measurement of low rankness and sparsity, increasing the post hoc \cite{an-2024-interpre} interpretability of our module. 4) We provide more results on the sparse object segmentation tasks, including infrared small target detection (IRSTD) \cite{dai-2021-acm}, vessel segmentation (VS \cite{drive-2014,hoover-2000-stare,chase_db1-2020}, and defect detection (DD) \cite{dong-2019-neuseg,song-2020-MCITF} to prove the generality of our method. We hope this efficient and interpretable architecture can be a new baseline in future sparse object segmentation research.

\section{Related Work}
\label{sec:2}
\subsection{RPCA for Sparse Object Segmentation}
Ever since the introduction of the RPCA model \cite{wright-2009-robust}, it has been extensively applied across various domains \cite{bouwmans-2018-rpca}. For low-level image processing tasks, most RPCA-based methods focus on inverse problems \cite{liu2020smooth,chang2020weighted,zhou2014low}, where sparse elements are treated as noise to be removed.

\begin{figure}
\setlength{\abovecaptionskip}{0pt}
\setlength{\belowcaptionskip}{0pt}
    \centering
    \includegraphics[width=1\linewidth]{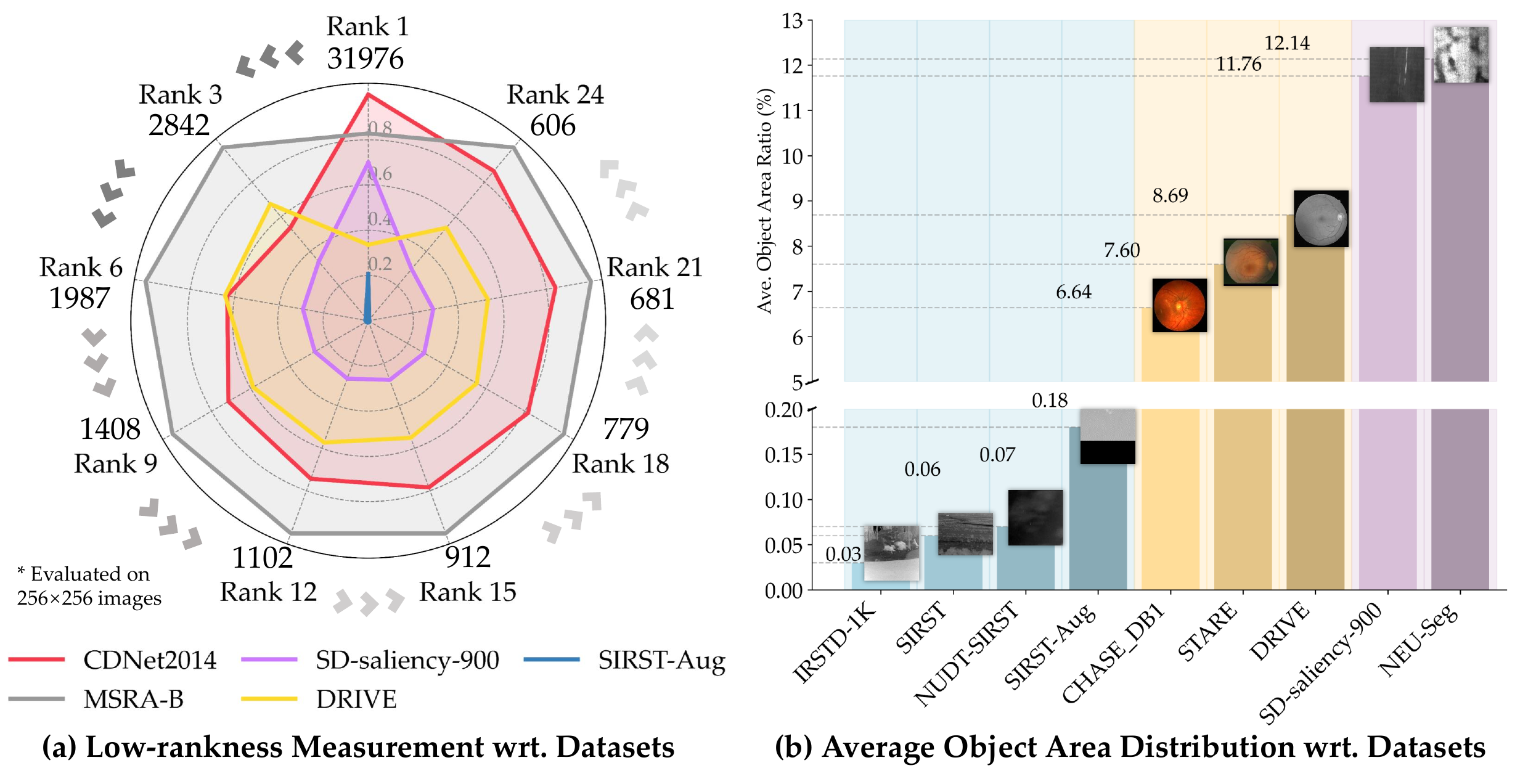}
    \caption{\textbf{(a)} \textbf{Low-rankness evaluation} of typical object segmentation tasks solved via RPCA, including \textcolor{Crimson}{BFS} (CDNet2014 \cite{wang2014cdnet2014}), \textcolor{gray}{SOD} (MSRA-B \cite{jiang2013msrab}), \textcolor{MediumPurple}{DD} (SD-saliency-900 \cite{song-2020-MCITF}), \textcolor{Goldenrod}{VS} (DRIVE \cite{staal-2004-drive}), and \textcolor{CadetBlue}{IRSTD} (SIRST-Aug \cite{zhang-2023-agpc}). Grayscale datasets exhibit low-rank properties, evidenced by the rapid decay of singular values in early ranks (detailed in Section \ref{sec:4.1.3}). The largest singular value is annotated around each rank, with the last three tasks showing superior low-rankness (SIRST-Aug being the "lowest-rank"). \textbf{(b)} \textbf{Objects' average area of each dataset:} Summary of nine sparse object segmentation datasets (\textcolor{CadetBlue}{IRSTD}, \textcolor{Goldenrod}{VS}, and \textcolor{MediumPurple}{DD}) with diverse average object area distributions analyzed in this study.}
    \label{fig:dataset}
\vspace{-0.5cm}
\end{figure}

In contrast, high-level vision tasks reinterpret sparse elements as informative outliers—grayscale regions highlighting foreground objects \cite{jou-2024-rpca}. This perspective has driven progress in foreground/background separation (BFS) \cite{liu-2015-background,Ebadi2018foreground,Javed2017foreground} and salient object detection (SOD) \cite{shen2012unified,peng2016salient,borji2019salient}. However, while such images often exhibit low-rank structure (Fig.~\ref{fig:dataset}(a)), the foreground regions are typically not truly sparse; their average area frequently exceeds 20\% \cite{li2014secrets,fan2018salient,jiang2013msrab}, and even surpasses 40\% in some datasets \cite{zhang2015salient,wang2021salient}. Consequently, classical RPCA methods falter under these conditions, prompting a shift toward using sparse components as downstream priors—ultimately complicating high-level segmentation.

Interestingly, recent studies leveraging RPCA-based segmentation frameworks have revealed that when objects are genuinely sparse—occupying a minimal area and exhibiting well-defined contours relative to the low-rank background—optimization techniques prove more effective than filter- or contrast-based methods. Consequently, researchers have refined RPCA models by incorporating domain-specific features. For example, in defect detection (DD, $\sim$10\%; Fig. \ref{fig:dataset}(b)), low-rank and sparse decomposition provide a foundation for implementing texture features, intrinsic priors, and energy constraints \cite{song-2020-MCITF,wang-2020-defect,ahmed2020sparse,qin2024metallic}. In medical imaging ($\sim$5\%), RPCA-based approaches focus on segmenting sparse vessels (VS) and enhancing them through intensity and edge-preservation priors \cite{lee-2018-rpca,qin2019accurate,FU-20201-cdrpca}. Additionally, for infrared small target detection, where sparse objects constitute less than 0.15\% of the image area \cite{ying-2025-rgbt-tiny}, RPCA frameworks have gained prominence. Since the pioneering work of \cite{gao-2013-ipi}, subsequent studies have advanced this field by improving background correlation \cite{wang-2017-tvpcp,liu-2021-asttv,wu-2023-4d} and incorporating sparse target constraints \cite{dai-2017-ript,zhang-2019-pstnn,kong-2022-logtfnn}.

In summary, RPCA provides a robust backbone for sparse object segmentation, with researchers integrating domain priors to enhance convergence speed and detection performance. Nevertheless, existing methods face significant challenges, including convergence inefficiencies and initialization complexities \cite{song-2020-MCITF}, which hinder their broader applicability due to computational demands and limited generalizability. In this article, the proposed RPCANet$^{++}$ aims to solve these challenges by integrating the interpretable RPCA model in a learnable and dynamic way to segment sparse objects.

\subsection{Deep Unfolding Networks}

\begin{figure*}[!ht]
\setlength{\abovecaptionskip}{1pt}
\setlength{\belowcaptionskip}{0cm}
\centering
\includegraphics[scale=0.57]{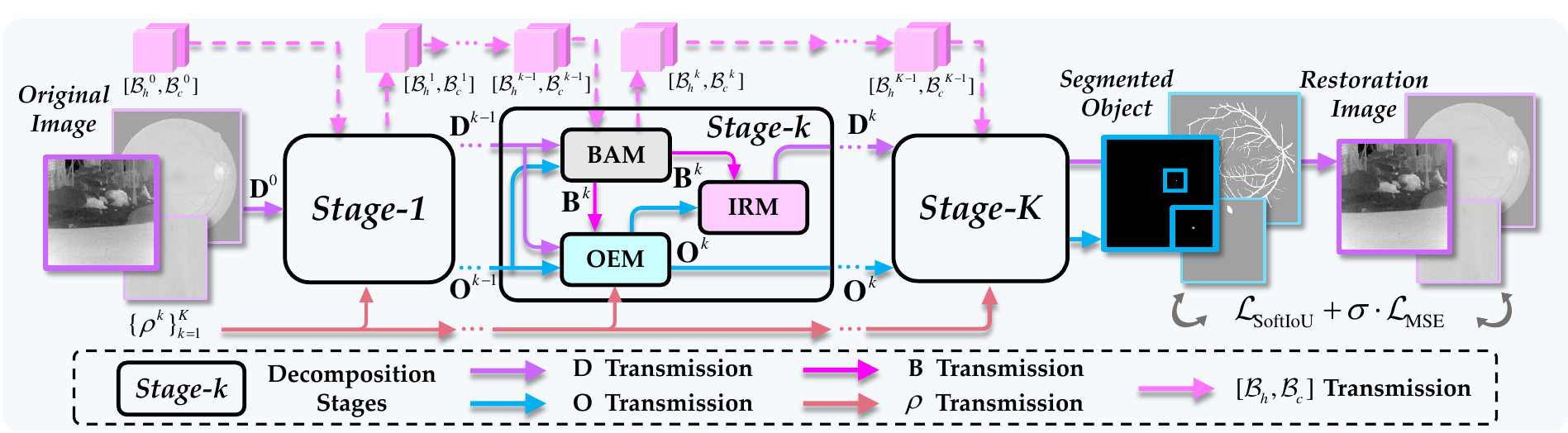}
\caption{\textbf{RPCANet$^{++}$ framework} unfolds iterative model-driven closed-form equations in deep network design and comprises corresponding $K$ stages. Transmissive elements are presented in different colors: $\mathbf{D}$ for the restoration image, $\mathbf{B}$ for the low rank background, $\mathbf{O}$ for the sparse object matrix, $\rho$ for the learnable parameter, and $[\mathcal{B}_h,\mathcal{B}_c]$ for the latent background features. The construction of each stage is depicted in Fig. \ref{fig:detail}.}
\label{fig:overall}
\vspace{-0.4cm}
\end{figure*}
Deep unfolding networks (DUN), also known as deep unrolling, integrate neural networks into iterative updates in optimization schemes and have achieved notable success in image inverse problems. Gregor and LeCun \cite{lecun-2010-lista} pioneered such a technique in sparse coding within a learned iterative shrinkage-thresholding algorithm (LISTA). Inspired by this work, Zhang and Ghanem designed ISTA-Net \cite{zhang-2018-istanet}, which replaces sparse transformation with nonlinear neural networks. Similar to the “deep unrolling" pattern, traditional iterative procedures such as alternating direction methods of multipliers (ADMM) \cite{sun-2016-admmnet}, approximate message passing (AMP) \cite{zhang-2021-ampnet}, and inertial proximal algorithm for nonconvex optimization (iPiano) \cite{su-2020-ipiano} are unrolled into deep networks. Considering the merit of interpretability and efficacy, scholars attempted to apply the DUN scheme in low-rank and sparse decomposition (or RPCA) problems, which shows great adaptive capacity in clutter suppression~\cite{solomon-2020-corona}, medical image restoration \cite{huang-2021-lrps}, multi-spectral and hyper-spectral image fusion \cite{yan-2023-fusion}, etc. Optimization-based sparse object segmentation task, as an RPCA-driven method, can theoretically fit into the deep unfolding paradigm.

From another perspective, although DUNs have proven effective and interpretable in various image inverse tasks, passing single-channel outputs across independent stages can lead to information degradation \cite{zhou-2023-lrsp,zhang-2023-physics}. To address this, prior works proposed strategies such as cascading accumulation \cite{zhang-2023-ctnet}, contextual memory blocks \cite{chen-2020-mem}, and inter/intra-stage memory fusion \cite{song-2021-madun}. Inspired by these efforts, we adopt a memory-augmented method to mitigate background transmission loss. Furthermore, DUNs' architectural flexibility \cite{you-2021-istanetpp} allows for integrating task-oriented priors \cite{zhang-2021-prior,yang-2022-prior,wei-2022-prior,imran-2022-prior} to enhance iterative optimization. In our design, drawing from local contrast mechanisms \cite{yu-2020-cdc} and the acceleration effects of priors in traditional optimization \cite{dai-2017-ript}, we embed a Deep Contrast Prior Module (DCPM) to improve segmentation accuracy while maintaining interpretability.


\section{Methodology}
\label{sec:3}

\subsection{Formulation of Optimization Problem}
\label{sec:3.1}
In the context of segmentation-oriented RPCA tasks, our objective is to estimate low-rank background $\mathbf{B}\in \mathbb{R}{^{m \times n}}$ and extract the sparse object matrix $\mathbf{O}\in \mathbb{R}{^{m \times n}}$. For an image $\mathbf{D}\in \mathbb{R}{^{m \times n}}$, we transform the segmentation model (\ref{eq_org}) into the following optimization framework:
\begin{equation}
\setlength{\abovedisplayskip}{2pt}
\setlength{\belowdisplayskip}{2pt}
    \min \limits_{\mathbf{B},\mathbf{O}} rank(\mathbf{B}) + \lambda \left\| \mathbf{O} \right\|_0 \quad s.t.~\mathbf{D} = \mathbf{B} + \mathbf{O} \enspace,
\label{eq_RPCA}
\end{equation}
where we signify $\lambda$ as a trade-off coefficient, and the term ${\left\|  \cdot  \right\|_0}$ denotes the $l_0$-norm, which is defined as the count of non-zero elements within a matrix.

However, addressing (\ref{eq_RPCA}) presents an NP-hard challenge due to the non-convex and discontinuous nature of both the rank function and the $l_0$-norm, which is often solved by the application of Principal Component Pursuit (PCP) \cite{zhou-2010-pcp}:
\begin{equation}
\setlength{\abovedisplayskip}{2pt}
\setlength{\belowdisplayskip}{2pt}
    \mathop {\min }\limits_{\mathbf{B},\mathbf{O}} {\left\| \mathbf{B} \right\|_*} + \lambda {\left\| \mathbf{O} \right\|_1}\quad s.t.\;\mathbf{D} = \mathbf{B} + \mathbf{O} \enspace,
\label{eq_PCP}
\end{equation}
where the term $\left\| \cdot \right\|_*$ represents the nuclear norm, which is the sum of the singular values of a matrix. Meanwhile, the notation $\left\| \cdot \right\|_1$ refers to the $l_1$-norm, defined as the sum of the absolute values of the matrix's entries. And such a model is widely adopted by traditional segmentation tasks \cite{gao-2013-ipi,tang-2016-WLRR}.

However, when facing complex scenarios, the background can exhibit varying degrees of complexity, rendering a solitary nuclear norm or rank function insufficient for encapsulating the practical constraints. Similarly, the sparsity of object elements can vary, making the exclusive use of the $l_0$ or $l_1$-norm potentially inadequate \cite{zhang-2021-srws}. Consequently, we propose a more generalized formulation of the problem. Here, we employ $\mathcal{R}(\mathbf{B})$ and $\mathcal{S}(\mathbf{O})$ as constraints that incorporate prior knowledge of the background and object images, individually:
\begin{equation}
\setlength{\abovedisplayskip}{2pt}
\setlength{\belowdisplayskip}{1pt}
    \min \limits_{\mathbf{B},\mathbf{O}} \mathcal{R}(\mathbf{B}) + \lambda \mathcal{S}(\mathbf{O}) \quad s.t.~\mathbf{D} = \mathbf{B} + \mathbf{O} \enspace.
\label{eq_relaxPCP}
\end{equation}

Furthermore, to alleviate the computational complexity associated with variable updates due to augmented Lagrange multipliers \cite{dai-2017-ript}, we opt for a more straightforward and intuitive approach. By employing the $l_2$-norm, convert the constrained optimization problem into an unconstrained format \cite{tseng-2008-convex}:
\begin{equation}
\setlength{\abovedisplayskip}{2pt}
\setlength{\belowdisplayskip}{2pt}
    \mathcal{L}(\mathbf{B},\mathbf{O}) = \mathcal{R}(\mathbf{B}) + \lambda \mathcal{S}(\mathbf{O}) + \frac{\mu}{2} \left\| \mathbf{D} - \mathbf{B} - \mathbf{O} \right\|^2_F \enspace,
\label{uncon}
\end{equation}
where penalty coefficient $\mu$ is introduced, and Frobenius norm (F-norm) ${\left\| \cdot \right\|_F}$ is utilized. 
The F-norm of a matrix $\mathbf{X}$ is given by: $\sqrt {\sum\limits_{i = 1}^m {\sum\limits_{j = 1}^n {{{\left| {{\mathbf{X}_{ij}}} \right|}^2}}}}$. 
Leveraging (\ref{uncon}), the background and object components are optimized iteratively.
\subsection{Iterative Solution}
\label{section:3.2}
\subsubsection{$\mathbf{B}^*$ Updates} For background updating, the sub-problem is articulated as:
\begin{equation}
\setlength{\abovedisplayskip}{3pt}
\setlength{\belowdisplayskip}{4pt}
    \mathbf{B}^* = \arg \min \limits_{\mathbf{B}} \mathcal{R}(\mathbf{B}) + \frac{\mu}{2} \left\| \mathbf{B} + \mathbf{O} - \mathbf{D}\right\|^2_F \enspace.
\label{eq_upbg}
\vspace{-0.2cm}
\end{equation}
As depicted in (\ref{eq_PCP}), traditional segmentation methods typically define $\mathcal{R}(\mathbf{B})$ as ${\left\| \mathbf{B} \right\|_*}$, thereby transform (\ref{eq_upbg}) to a combination of the nuclear norm and the $l_2$ norm. An analytical solution to this problem is known and can be expressed as follows:
\begin{equation}
\setlength{\abovedisplayskip}{2pt}
\setlength{\belowdisplayskip}{3pt}
    \mathbf{B}^* = \mathcal{\mathcal{D}}_{\mu}(\mathbf{D}-\mathbf{O}) \enspace,
    \vspace{-0.15cm}
\label{solv}
\end{equation}
where the operator $ \mathcal{D}_{\mu}(\cdot)$  represents the Singular Value Thresholding (SVT) \cite{ma-2011-fixed} with a threshold of $\mu$. Addressing (\ref{solv}) entails SVD, which is computationally intensive and impacts precision. However, once DUNs mimic this process via neural networks, which necessitates SVD at each iteration, it will lead to efficiency and accuracy concerns. The initialize method, proposed by \cite{cai-2021-lrpca}, involves decomposing a matrix into two matrices through best rank-$r$ approximation SVD, followed by separate updates. Despite this, SVD's computational challenges persist. Alternative approaches \cite{zhang-2022-lrcsnet} involve matrix decomposition under varying rank constraints, enhancing interpretability through ranks and dimensions. Yet, these require manual rank selection and may overlook the inherent features of images in neural layers.
\begin{figure*}[!ht]
\setlength{\abovecaptionskip}{1pt}
\setlength{\belowcaptionskip}{0pt}
\centering
	\includegraphics[width=0.88\linewidth]{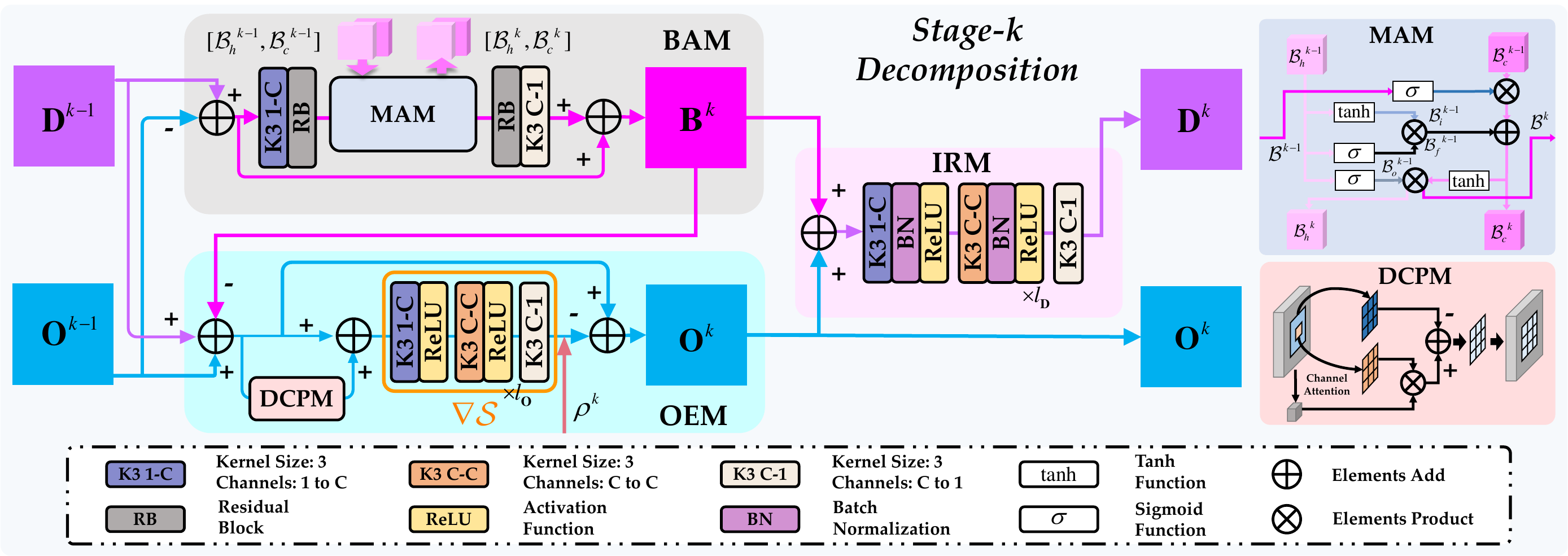}
   \caption{\textbf{Detail network structure} of a \textbf{single stage} RPCANet$^{++}$, consisting of background approximation module (\textbf{BAM}), object extraction module (\textbf{OEM}), and image restoration module (\textbf{IRM}). Memory augmented module (\textbf{MAM}) is injected in BAM while the deep contrast prior module (\textbf{DCPM}) is embedded in OEM. Notably, the channel number is set as 32 in this framework. [Zoom in for a better view]}
\label{fig:detail}
\vspace{-0.5cm}
\end{figure*}

Rather than using the nuclear norm and engaging in complex SVD computations, we simplify the process by implementing a constraint function $\mathcal{R}(\mathbf{B})$.
Additionally, we utilize a proximal operator ${\text{pro}}{{\text{x}}_\mu}( \cdot )$ to estimate the background’s closed-form solution. The equation is delineated as follows:
\begin{equation}
\setlength{\abovedisplayskip}{1pt}
\setlength{\belowdisplayskip}{0pt}
    \mathbf{B}^* = {\text{prox}}_{\mu}(\mathbf{D} - \mathbf{O})\enspace.  
\label{proxB}
\end{equation}

We utilize convolutional layers to approximate proximal functions, thereby simplifying complex matrix computations and tapping into the nonlinear processing of neural networks for deep feature extraction from images (detailed in Section \ref{sec:3.3}).
\vspace{-0.2cm}
\subsubsection{$\mathbf{O}^*$ Updates} Similar to (\ref{eq_upbg}), the sub-problem for optimizing the object is expressed as follows:
\begin{equation}
\setlength{\abovedisplayskip}{2pt}
\setlength{\belowdisplayskip}{2pt}
    \mathbf{O}^* = \arg \min \limits_{\mathbf{O}} \lambda \mathcal{S}(\mathbf{O}) + \frac{\mu}{2} \left\| \mathbf{O} + \mathbf{B} - \mathbf{D} \right\|^2_F \enspace.
\label{upta}
\end{equation}

In prevalent optimization strategies, the sparse object is usually constrained by an \( l_1 \) norm. Yet, integrating soft thresholding (ST) within neural networks poses difficulties \cite{zhang-2018-istanet}. Moreover, the nature of sparse constraints can fluctuate with varying detection contexts \cite{zhang-2021-srws}. We aim to formulate a more straightforward and intuitive representation for the closed-form resolution of (\ref{upta}).

To address these challenges, we apply the Taylor expansion to $\mathcal{S}(\mathbf{O})$. Given a function \(f(t)\) with a Lipschitz continuous gradient \(\nabla f(t)\), it can be approximated at a fixed point \(t_0\) via Taylor's series:
\begin{equation}
\setlength{\abovedisplayskip}{1pt}
\setlength{\belowdisplayskip}{1pt}
\hat f(t,{t_0})  \leftarrow \frac{L}{2}\left\| {t - {t_0} + \frac{1}{L}f({t_0})} \right\|^2 + C \enspace.
\end{equation}

In this case, $L$ is a constant and $C$ is defined as $C =  - \frac{1}{{2{L}}}{\left\| \nabla{f({t_0})} \right\|^2} + f({t_0})$ (For detailed derivations, please refer to the \textbf{Appendix A}). Consequently, we approximate $\mathcal{S}(\mathbf{O})$ at the previous iteration $\mathbf{O}^{k-1}$ as follows:
\begin{equation}
\setlength{\abovedisplayskip}{1pt}
\setlength{\belowdisplayskip}{2pt}
    \mathcal{\hat{S}}(\mathbf{O}, \mathbf{O}^{k-1}) \leftarrow \frac{L_{\mathcal{S}}}{2} \left\| \mathbf{O}\!-\!\mathbf{O}^{k-1}\!+\!\frac{1}{L_{\mathcal{S}}} \nabla \mathcal{S} \left( \mathbf{O}^{k-1} \right) \right\|^2_2\!+\!C_s \enspace,
\end{equation}
where $L_s$ denotes the Lipschitz constant for $\mathcal{S}(\mathbf{O})$, and $C_s$ is defined as a constant, represented by $C_s=-\frac{1}{{2{L_s}}}{\left\| {\nabla\mathcal{S}({\mathbf{O}^{k - 1}})} \right\|_2^2} + \mathcal{S}({\mathbf{O}^{k - 1}})$. This allows us to update the object matrix in a simplified manner.
\begin{equation}
\setlength{\abovedisplayskip}{1pt}
\setlength{\belowdisplayskip}{1pt}
\begin{aligned}
{\mathbf{O}^*} =& \arg \mathop {\min }\limits_\mathbf{O} \lambda \hat{\mathcal{S}}(\mathbf{O},{\mathbf{O}^{k - 1}})\!+\!\frac{\mu }{2}\left\| {\mathbf{O} + \mathbf{B} - \mathbf{D}} \right\|_F^2\\
 =& \arg \min \limits_{\mathbf{O}} \frac{\lambda{{L_\mathcal{S}}}}{2}\left\| {\mathbf{O} - {\mathbf{O}^{k - 1}} + \frac{1}{{{L_\mathcal{S}}}}\nabla \mathcal{S}({\mathbf{O}^{k - 1}})} \right\|_2^2 \\
 &+ \frac{\mu }{2}\left\| {\mathbf{O} + \mathbf{B} - \mathbf{D}} \right\|_F^2 \enspace.
\end{aligned}
\label{optT}
\end{equation}

Instead of the conventional \( l_1 \) norm constraint, this formulation incorporates only the sum of two $l_2$ norms. This simplification eliminates the need for conventional algorithms or soft thresholding simulations \cite{zhang-2018-istanet}. Deriving the equation's derivative and equating it to zero, we deduce a closed-form solution for the \( k \)-th iteration update of \(\mathbf{O}\):
\begin{equation}
\setlength{\abovedisplayskip}{1pt}
\setlength{\belowdisplayskip}{1pt}
\begin{aligned}
\mathbf{O}^{k} = & \frac{\lambda L_{\mathcal{S}}}{\lambda L_{\mathcal{S}} + \mu} \mathbf{O}^{k-1} + \frac{\mu}{\lambda L_{\mathcal{S}} + \mu} \left( \mathbf{D}^{k - 1} - \mathbf{B}^{k} \right)\\ 
&- \frac{\lambda}{\lambda L_{\mathcal{S}} + \mu} \nabla \mathcal{S}(\mathbf{O}^{k-1}) \enspace.
\end{aligned}
\label{closed}
\end{equation}

Thus, the object matrix update equation is succinctly reformulated as follows:
\begin{equation}
\setlength{\abovedisplayskip}{1pt}
\setlength{\belowdisplayskip}{1pt}
{\mathbf{O}^k}= \gamma {\mathbf{O}^{k-1}} + (1-\gamma )(\mathbf{D}^{k-1}-\mathbf{B}^k)-\rho \nabla S({\mathbf{O}^{k - 1}})\enspace,
\label{eq:upt}
\end{equation}
where two coefficients: $\gamma\!=\!\frac{{\lambda {L_S}}}{{\lambda {L_S}\!+\!\mu }}$ and $\rho\!=\!\frac{\lambda}{{\lambda {L_S}\!+\!\mu }}$. By learning the function $\nabla\mathcal{S}$ end-to-end, our method avoids intricate matrix operations such as ST while ensuring Lipschitz continuity.
\subsubsection{$\mathbf{D}^*$ Updates} 
The restoration update formula for $\mathbf{D}^{k}$ is efficiently obtained through:
\begin{equation}
\setlength{\abovedisplayskip}{-1pt}
\setlength{\belowdisplayskip}{0pt}
    {\mathbf{D}^k}={\mathbf{B}^k}+{\mathbf{O}^k}\enspace.
    \label{finald}
\end{equation}
\subsection{Unfolding into Deep Framework: RPCANet$^{++}$}
\label{sec:3.3}
This section outlines RPCANet$^{++}$'s architecture and modules, derived from Section \ref{section:3.2}'s optimization equations. 
Given an image denoted as $\mathbf{X} \in \mathbb{R}^{H\times W}$, where $H$ and $W$ represent the image's height and width, respectively, we initialize $\mathbf{D}^0 = \mathbf{X}$ while $\mathbf{O}^0 = 0$.
It undergoes $K$ decomposition stages, each simulating iterative low-rank sparse decomposition.

At each $k$-th stage, we input $\mathbf{D}^{k-1}$ and $\mathbf{O}^{k-1}$ to estimate $\mathbf{B}^k$, $\mathbf{O}^k$, and $\mathbf{D}^k$ using BAM, OEM, and IRM. We innovate beyond \cite{wu-2024-rpcanet} by incorporating a memory-augmented module (MAM) for background transmission loss, utilizing ConvLSTM \cite{shi-2015-lstm} for feature retention. Additionally, we introduce a deep contrast prior module named DCPM, inspired by local contrast \cite{yu-2020-cdc,ying-2022-moco,zhang2025irmamba,wu-2025-l2sknet}, to refine object extraction.
\subsubsection{Background Approximation Module (BAM)}
\begin{figure}[t]
\setlength{\abovecaptionskip}{-0.1cm}
\setlength{\belowcaptionskip}{-0.1cm}
    \centering
    \includegraphics[width=\linewidth]{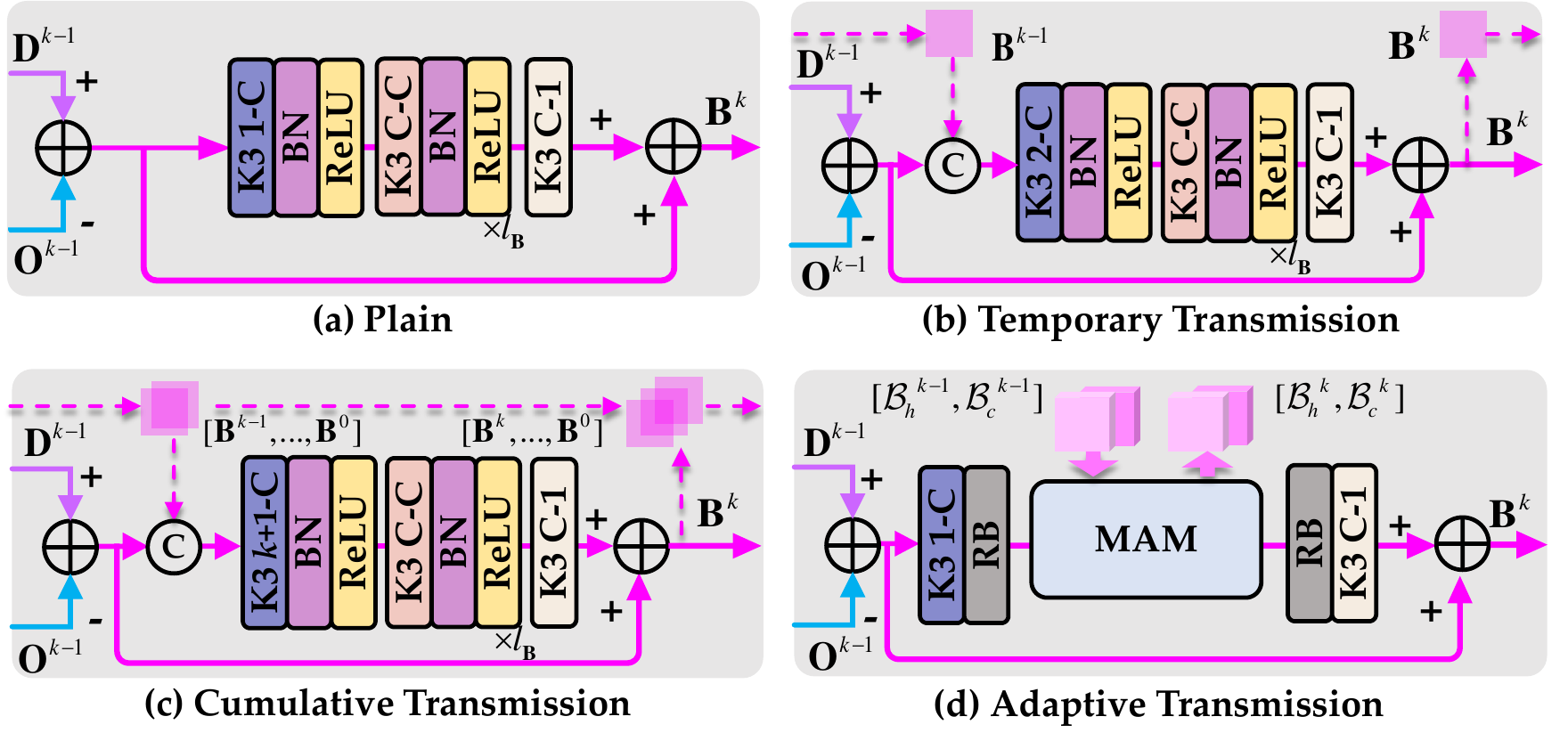}
    \caption{\textbf{Comparison} of different constructions of BAM, (a)~\textbf{Plain} construction \cite{wu-2024-rpcanet}; (b)~Temporary Transmission (\textbf{TT}) concatenates last stage background information;
 (c)~Cumulative Transmission (\textbf{CT}) \cite{zhang-2023-ctnet} concatenates all previous background features; (d) \textbf{Ours} select background features adaptively.}
    \vspace{-0.4cm}
    \label{fig:bam}
\end{figure}

Fig. \ref{fig:detail} illustrates using BAM for background estimation. The proximal operator $\text{prox}_\mu(\cdot)$ in (\ref{proxB}) is yet to be defined. Previous work used an ISTANet$^{++}$-like \cite{you-2021-istanetpp} residual structure, which only considered restored and object information, potentially causing feature loss. To address this, we introduce a memory-augmented module (MAM) for feature preservation, compatible with DUN. Drawing inspiration from recent studies \cite{song-2021-madun,song-2023-mapun}, instead of directly concatenating previous features \cite{zhang-2023-ctnet} (as shown in Fig. \ref{fig:bam}), we integrate ConvLSTM \cite{shi-2015-lstm} to manage past and current information flow, processed through an initial layer of $[Conv + BN + ReLU]$, an efficient ResBlock (widely utilized in \cite{zhang-2021-ampnet, wang-2023-indudonet+,mifdal-2023-pan}, etc.), and a symmetric decoder, collectively termed $\text{proxNet}(\cdot)$.

In detail, the MAM is sandwiched by two convolutional layers ($Conv_{B_1}(\cdot)$ and $Conv_{B_2}(\cdot)$) and two ResBlocks ($\mathcal{RB}_1(\cdot)$ and $\mathcal{RB}_2(\cdot)$) and it has three inputs: deep feature $\mathcal{B}^{k-1}$, hidden states $\mathcal{B}^{k-1}_h$, and cell outputs $\mathcal{B}^{k-1}_c$ (we use calligraphy to denote multi-channel features). We define the output of MAM as:
\begin{equation}
\setlength{\abovedisplayskip}{1pt}
\setlength{\belowdisplayskip}{1pt}
\centering
     [\mathcal{B}^{k}_h, \mathcal{B}^{k}_c] =\text{ConvLSTM}(\mathcal{B}^{k-1}, [\mathcal{B}^{k-1}_h, \mathcal{B}^{k-1}_c])  
\label{eq:mam}
\end{equation}
Given the deep inputs, the process is presented in the shadow blue part in the top-left of Fig. \ref{fig:detail} and formulated as:
\begin{equation}
\setlength{\abovedisplayskip}{1pt}
\setlength{\belowdisplayskip}{1pt}
\centering
\begin{aligned}
     &\mathcal{B}^{k}_i = \sigma(\mathbf{W}_{{\mathcal{B}}{\mathcal{B}_i}}*\mathcal{B}^{k-1} +\mathbf{W}_{{\mathcal{B}_h}{\mathcal{B}_i}}*{\mathcal{B}^{k-1}_h}+\textbf{b}_{\mathcal{B}_i});\\
     &\mathcal{B}^{k}_f = \sigma(\mathbf{W}_{{\mathcal{B}}{\mathcal{B}_f}}*\mathcal{B}^{k-1} +\mathbf{W}_{{\mathcal{B}_h}{\mathcal{B}_f}}*{\mathcal{B}^{k-1}_h}+\textbf{b}_{\mathcal{B}_f});\\
     &\mathcal{B}^{k}_c = \mathcal{B}^{k}_f \circ \mathcal{B}^{k-1}_{c} + \mathcal{B}^{k}_i \circ \text{tanh}(\mathbf{W}_{{\mathcal{B}_s}{\mathcal{B}_c}}*\mathcal{B}^{k-1} \\
     &~~~~~~~~~~~~~~~~~~~~~~~~~~~~~+\mathbf{W}_{{\mathcal{B}_h}{\mathcal{B}_c}}*{\mathcal{B}^{k-1}_h}+\textbf{b}_{\mathcal{B}_c});\\
     &\mathcal{B}^{k}_o = \sigma(\mathbf{W}_{{\mathcal{B}}{\mathcal{B}_o}}*\mathcal{B}^{k-1} +\mathbf{W}_{{\mathcal{B}_h}{\mathcal{B}_o}}*{\mathcal{B}^{k-1}_h}+\textbf{b}_{\mathcal{B}_o});\\
     &\mathcal{B}^{k}_h = \mathcal{B}^{k}_o \circ \text{tanh}(\mathcal{B}^{k}_c).
\end{aligned}
\label{eq:lstm}
\end{equation}
Here, the convolution operation is denoted by $*$, and the Hadamard product by $\circ$. The functions $\sigma(\cdot)$ and $\text{tanh}(\cdot)$ represent the sigmoid and tanh activations, respectively. The filter weights are represented by $\mathbf{W}_{\mathcal{B}\mathcal{B}_i}$, $\mathbf{W}_{\mathcal{B}\mathcal{B}_f}$, ..., $\mathbf{W}_{\mathcal{B}\mathcal{B}_o}$, and their corresponding biases by $\mathbf{b}_{\mathcal{B}\mathcal{B}_i}$, $\mathbf{b}_{\mathcal{B}\mathcal{B}_f}$, ..., $\mathbf{b}_{\mathcal{B}\mathcal{B}_o}$. The input, forget, and output gates are denoted as $\mathcal{B}^{k}_i$, $\mathcal{B}^{k}_f$, $\mathcal{B}^{k}_o$, respectively. We assign $\mathcal{B}^{k}$ to $\mathcal{B}^{k}_h$ for general use, which serves as the input to $\mathcal{RB}_2(\cdot)$. The pair $[\mathcal{B}^{k-1}_h, \mathcal{B}^{k-1}_c]$ is propagated through successive stages, ensuring the adaptive retention of background features throughout the process.

In summary, the total structure of BAM can be collected as:
\begin{equation}
\setlength{\abovedisplayskip}{1pt}
\setlength{\belowdisplayskip}{1pt}
\centering
\begin{aligned}
\mathbf{B}^k & = \text{proxNet}(\mathbf{D}^{k-1},\mathbf{O}^{k-1},[\mathcal{B}^{k-1}_h,\mathcal{B}^{k-1}_c])\\
&= \mathbf{D}^{k-1}\!-\!\mathbf{O}^{k-1}\!+\!Conv_{B_2}(\mathcal{RB}_2(\text{ConvLSTM}(\\
&~~~~\mathcal{RB}_1(Conv_{B_1}(\mathbf{D}^{k-1}\!-\!\mathbf{O}^{k\!-\!1})), [\mathcal{B}^{k\!-\!1}_h, \mathcal{B}^{k\!-\!1}_c]))),
\end{aligned}
\label{eq:bam}
\end{equation}
where we also incorporate residual construction to enhance training stability. Unlike \cite{song-2021-madun}, which relies heavily on intra-stage information for recovery, our BAM functions as a lightweight bridge between OEM and IRM. By selectively filtering redundant channel features, we streamline inter-module communication and reduce computational overhead.

\subsubsection{Object Extraction Module (OEM)} 
As Section \ref{section:3.2} deducts, we design the object extraction module for object updates, to treat three variables evenly, we assign $\gamma$ to 0.5. And (\ref{eq:upt}) are formulated as:
\begin{equation}
\setlength{\abovedisplayskip}{2pt}
\setlength{\belowdisplayskip}{2pt}
{\mathbf{O}^k}\!=\!{\mathbf{O}^{k - 1}}\!+\!{\mathbf{D}^{k - 1}}\!-\!{\mathbf{B}^k}\!-\!{\rho}\nabla\mathcal{S}({\mathbf{O}^{k - 1}}) \enspace.
\label{eq:tnet}
\end{equation}
We set $\rho$ as a stage-independent learnable parameter as $\rho^{k}$. And $\nabla\mathcal{S}$, a Lipschitz continuous gradient function, is simulated by a $[Conv + ReLU]$ construction, where the Lipschitz continuity can remain even within multistacked layers \cite{virmaux-2018-lipschitz}. As the orange box in Fig. \ref{fig:detail} shows, we denote such theory-guided networks as $\mathcal{G}(\cdot)$.

In traditional object iterations, the sparse constraint module is always assisted by a weighted prior module \cite{dai-2017-ript} for convergence acceleration. Such prior can be correlated with physical guidance such as saliency \cite{kong-2022-logtfnn}, which is generally written as:
\begin{equation}
\setlength{\abovedisplayskip}{2pt}
\setlength{\belowdisplayskip}{2pt}
{\bf{W}}^{k}  =  \frac{C_T}{{\left| {\bf{O}}^{k-1} \right| + \varepsilon }} \enspace,
\label{eq:orgw}
\end{equation}
$C_T$ is a manual constant, and $\varepsilon$ is set to avoid the "division by zero" issue. And the $\nabla\mathcal{S}(\bf{O})$ will be rewritten as $\nabla\mathcal{S}(\bf{W \circ O})$.  However, such collaboration still requires parameter fine-tuning, and introducing a division operation in neural networks will introduce issues such as "Inf" (object element approaching zero). 
\begin{figure}[!t]
\setlength{\abovecaptionskip}{0cm}
\setlength{\belowcaptionskip}{0cm}
    \centering
    \includegraphics[width=\linewidth]{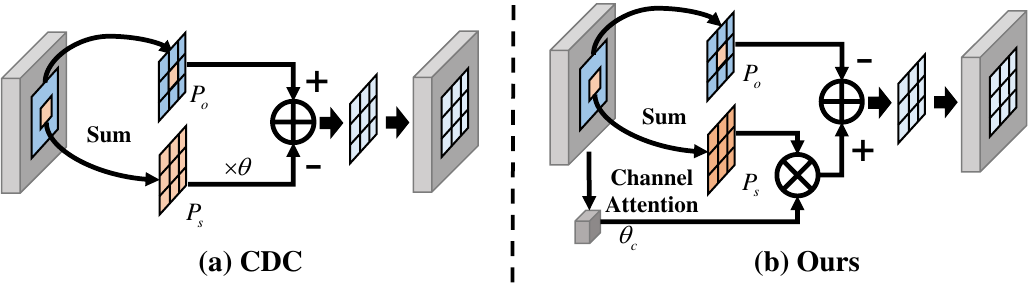}
    \caption{\textbf{Comparison} of different local contrast feature extraction methods: construction of CDC \cite{yu-2020-cdc} and our DCPM.}
    \vspace{-0.6cm}
    \label{fig:dcpm}
\end{figure}
Thus, instead of directly implementing the prior correlation conventionally, we pre-extract the saliency feature with a deep prior contrast module via an enhanced CDC. As shown in Fig. \ref{fig:dcpm}, unlike the original CDC \cite{yu-2020-cdc}, we employ a channel attention module to perceive deep features and adaptively adjust the manual $\theta$ in the general CDC. To be specific, a deep contrast prior extraction process consists of four steps: 1) sampling with a $s\times s$ kernel $P_o$ over the input feature $\mathcal{X}^{k-1}\in \mathbb{R}^{C\times H \times W}$; 2) summing the convolutional weights to central weight $P_s$; 3) making $\theta$ learnable by compressing the multi-channel information via simple channel attention \cite{hu-2018-senet} module $\theta_c$; 4) generalizing the network $\mathcal{P}(\cdot)$ in a central-difference scheme. We formulate this process as follows: 
\begin{equation}
\setlength{\abovedisplayskip}{2pt}
\setlength{\belowdisplayskip}{2pt}
\mathcal{P}(\mathcal{X}^{k-1}) = \theta_c(\mathcal{X}^{k-1}) \cdot \mathcal{P}_s(\mathcal{X}^{k-1}) - \mathcal{P}_o( \mathcal{X}^{k-1})
 \enspace,
\label{eq:cdc}
\end{equation}
$\mathcal{P}(\cdot)$ is sandwiched by two convolutional sections $Conv_{T_1}(\cdot)$ and $Conv_{T_2}(\cdot)$, and the overall DCPM network is written as:
\begin{equation}
\setlength{\abovedisplayskip}{2pt}
\setlength{\belowdisplayskip}{2pt}
{\mathbf{W}}^{k} =  Conv_{T_2}(\mathcal{P}(Conv_{T_1}(\mathbf{X}^{k-1})))
 \enspace.
\label{eq:w}
\end{equation}

To align with (\ref{eq:tnet}) and maintain the difference of previous reconstructed $\mathbf{D}^{k - 1}$ and current approximated $\mathbf{B}^{k}$, we assign $\textbf{X}^{k-1} = {\mathbf{O}^{k - 1}}\!+\!{\mathbf{D}^{k - 1}}\!-\!{\mathbf{B}^k}$. Similar to the prior assisted DUNs \cite{zhang-2021-prior, yang-2022-prior,wei-2022-prior}, we directly add the prior before ${{\mathcal{G}}^k}$, and the final update equation for OEM is collected as follows:
\begin{equation}
\setlength{\abovedisplayskip}{2pt}
\setlength{\belowdisplayskip}{2pt}
{{\bf{O}}^k\!=\!{{\bf{O}}^{k\!-\!1}}\!+\!{{\bf{D}}^{k\!-\!1}}\!-\!{{\bf{B}}^k}\!-\!{\rho ^k}{{\mathcal{G}}^k}({{\bf{O}}^{k\!-\!1}}\!+\!{{\bf{D}}^{k\!-\!1}}\!-\!{{\bf{B}}^k}\!+\!{{\bf{W}}^k})}.
\label{tnet}
\end{equation}

The overall OEM structure is demonstrated in Fig. \ref{fig:detail}, where $l_{\mathbf{O}}$ represents the number of middle layers. 

\subsubsection{Image Restoration Module (IRM)} 
We reconstruct the infrared image by merging the updated $\mathbf{B}^{k}$ and $\mathbf{O}^{k}$ via an simple yet efficient restoration module $\mathcal{M}(\cdot)$, as shown in the purple box in Fig. \ref{fig:detail} and formulated by:
\begin{equation}
\setlength{\abovedisplayskip}{2pt}
\setlength{\belowdisplayskip}{2pt}
{{\bf{D}}^k} = {\mathcal{M}^k}({{\bf{B}}^k} + {{\bf{O}}^k})\enspace,
\end{equation}
where $\mathcal{M}(\cdot)$ are comprised of one $[Conv + ReLU]$ block, $l_{\mathbf{D}}$ of $[Conv + BN + ReLU]$, and a $[Conv]$ layers.

\begin{figure}[!t]
\setlength{\abovecaptionskip}{0.1cm}
\setlength{\belowcaptionskip}{0cm}
    \centering
    \includegraphics[width=\linewidth]{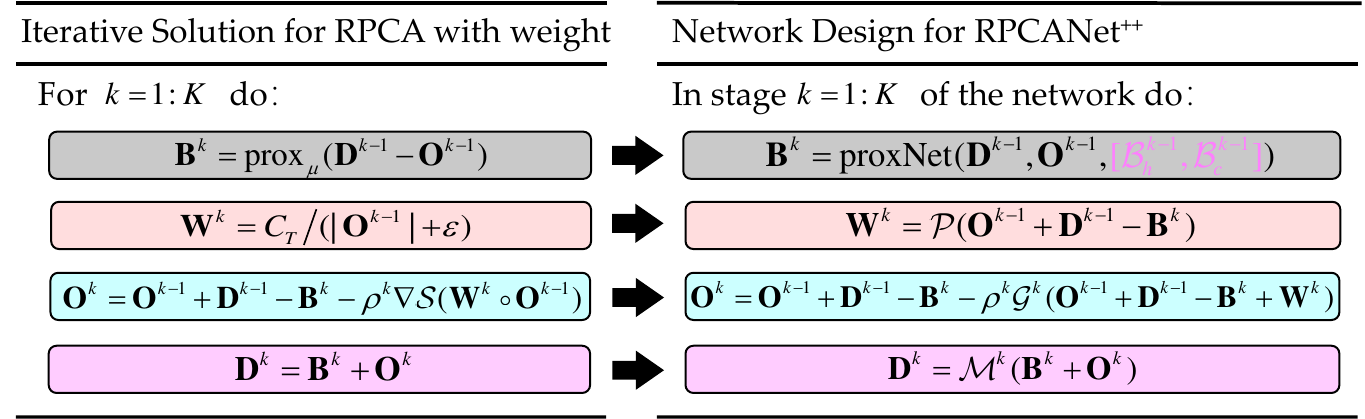}
    \caption{\textbf{Algorithm2Network}: Solutions of optimization algorithm and corresponding network architecture of RPCANet$^{++}$.}
    \vspace{-0.4cm}
    \label{fig:a2n}
\end{figure}

In summary, the three modules are orderly updated within a single stage and iterative for $K$-stages as Fig. \ref{fig:overall} shows. The overall algorithm-to-network diagram is demonstrated in Fig. \ref{fig:a2n}.

\subsection{Network Implementation} 
\subsubsection{Learnable Parameters}
The parameter set $\Theta$ of our RPCANet$^{++}$ is defined as $\Theta  = \left\{ {\Theta _{\mathbf{BAM}}^k,\Theta _{\mathbf{OEM}}^k,\Theta _{\mathbf{IRM}}^k,{\rho ^k}} \right\}_{k = 1}^K$. This encompasses the network parameters within BAM (with MAM), OEM (with DCPM), and IRM for each stage of decomposition, along with the trainable scalar $\rho^{k}$ in OEM. 

In \textbf{BAM}, the stride and padding for the $3\times 3$ convolutional layers, $Conv_{B_1}(\cdot)$ and $Conv_{B_2}(\cdot)$, are uniformly set to 1, with the channel expansion number $C$ specified as 32. The Residual Blocks, $\mathcal{RB}_1(\cdot)$ and $\mathcal{RB}_2(\cdot)$, adhere to a structured sequence: $[Conv + BN + ReLU + Conv + BN + Skip ~Connection]$, maintaining consistency in stride, padding, kernel size, and channel dimensions. Regarding \textbf{OEM}, the kernel size for the deep contrast prior $\mathcal{P}(\cdot)$ is established at $17\!\times\!17$. The channel attention module's ratio $r$ within $\theta_c$ is fixed at 4. Furthermore, all convolutional layers in this module are configured with a stride and padding of 1, a kernel size of $3\!\times\!3$, and an increased channel count of 32. The sparse layer parameter $l_{\bf{O}}$ is set to 6. Similarly, the reconstruction module $\mathcal{M}(\cdot)$ in \textbf{IRM} follows identical convolution setting, with the layer number $l_{\bf{D}}\!=\!3$.
\subsubsection{Training Loss}
Compared with conventional networks with singular segmentation tasks, our RPCANet$^{++}$ not only maps the object feature to masks but also fulfills the image restoration function. Specifically, we adopt Soft Intersection over Union (SoftIoU) \cite{huang-2020-softiou} as our segmentation loss, while a generalized mean squared error (MSE) is employed for reconstruction loss. The combined loss function is formulated as:
\begin{equation}
\setlength{\abovedisplayskip}{3pt}
\setlength{\belowdisplayskip}{2pt}
\begin{aligned}
&{{\mathcal{L}}_{{\text{All}}}} = \mathcal{L}_\text{SoftIoU} + \sigma  \cdot {\mathcal{L}_\text{MSE}}\enspace \\
&=1 \!-\!\frac{1}{{{N_t}}}\sum\limits_{i = 1}^{{N_t}}{\frac{{TP}}{{FP\!+\!TP\!+\!FN}}}\!+\!\frac{ \sigma }{{{N_t}N}}\sum\limits_{i = 1}^{{N_t}} {\left\| {{\mathbf{D}^K}\!-\!\mathbf{D}} \right\|_F^2} 
\end{aligned}
\end{equation}
where $N_t$ and $N$ are the total training number and total pixels per image. $\sigma$ represents the regularization parameter, which is set to 0.1 in our experiments (detailed in Section \ref{sec:4}).

\begin{figure*}[!ht]
\centering
\setlength{\abovecaptionskip}{1pt}
\setlength{\belowcaptionskip}{0pt}
   \includegraphics[width=\linewidth]{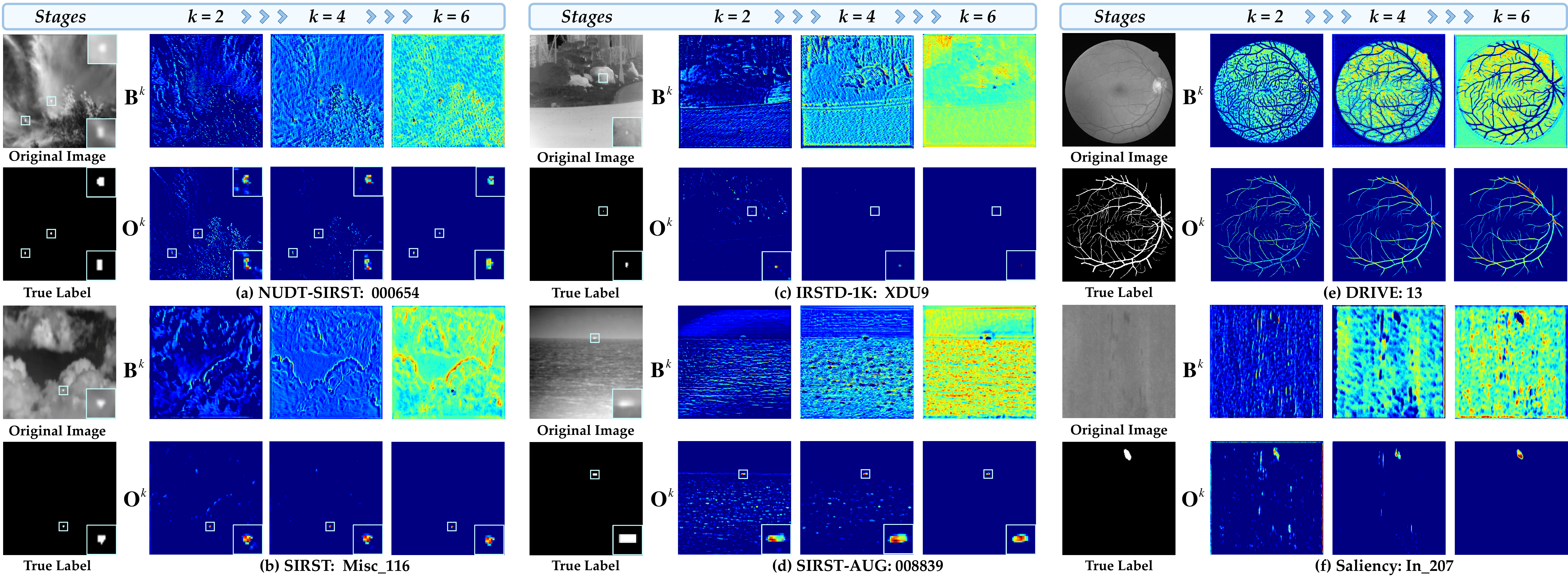}
   \caption{\textbf{Heatmaps} of different stages' $\textbf{B}^{k}$ and $\textbf{O}^{k}$ visualization results ($K=6$) of our RPCANet$^{++}$ on various scenarios from six different datasets (\textbf{IRSTD}, \textbf{VS}, and \textbf{DD} tasks). We can observe its gradual shaping process via iterative unfolding. [Zoom in with a \textcolor{cyan!30}{blue} box for tiny objects]}
   \label{fig:mv}
   \vspace{-0.3cm}
\end{figure*}

\begin{figure*}[!ht]
\centering
\setlength{\abovecaptionskip}{1pt}
\setlength{\belowcaptionskip}{0pt}
   \includegraphics[width=\linewidth]{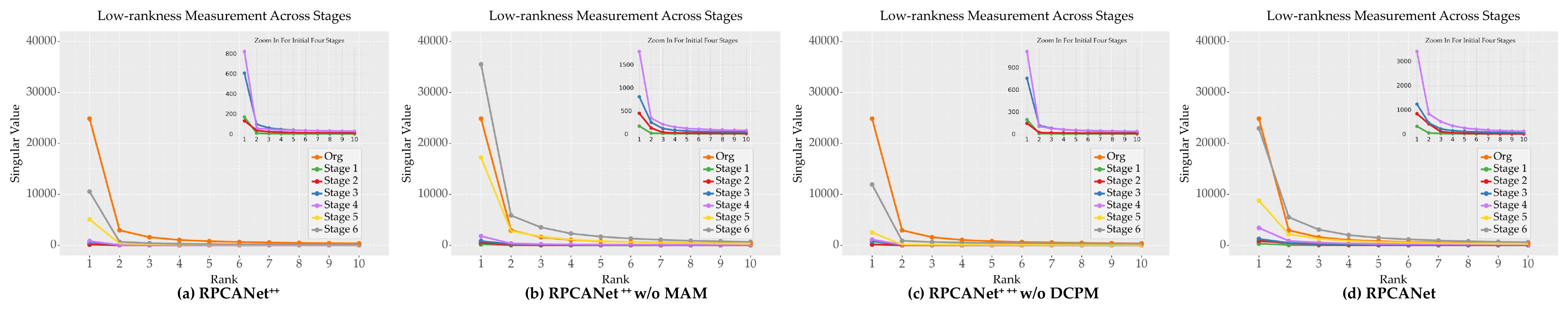}
   \caption{\textbf{Low-rankness verification} of different stage features (1st to 6th) in \textbf{(a)} RPCANet$^{++}$, compared to original images.  As well as its variants \textbf{(b)} without  MAM or \textbf{(c)} without DCPM, and the baseline \textbf{(d)} RPCANet \cite{wu-2024-rpcanet}. Verification is conducted on the IRSTD-1K test set \cite{zhang-2022-isnet}. Our RPCANet$^{++}$ progressively estimates background features satisfying low-rankness, step-by-step, without overestimation. [Zoom in for a better view]}
   \label{fig:interl}   
      \vspace{-0.4cm}
\end{figure*}

\section{Experiment and Results}
\label{sec:4}
\subsection{Implement Details}
\label{sec:4.1}
\subsubsection{Datasets}
\noindent \textbf{Infrared Small Target Segmentation (IRSTD):} We utilize four distinct open-source IRSTD datasets: NUDT-SIRST \cite{li-2023-dnanet}, IRSTD-1K \cite{zhang-2022-isnet},  SIRST \cite{dai-2021-acm}, and SIRST-Aug \cite{zhang-2023-agpc}. NUDT-SIRST comprises 1327 \(256 \times 256\) images, encapsulating diverse scenes such as cloud, city, sea, and field, featuring both natural and synthetic targets. Following the protocol of \cite{li-2023-dnanet}, we adopt a \(50:50\) split for training and testing. IRSTD-1K, documented in \cite{zhang-2022-isnet}, contains 1000 \(512 \times 512\) genuine images, spanning a variety of targets and scenarios. Although the original dataset division is \(50:30:20\) for training, validation, and testing, we align with the configuration of \cite{ying-2023-lesps}, opting for an \(80:20\) split. SIRST-Aug, introduced by \cite{zhang-2023-agpc}, enhances  SIRST \cite{dai-2021-acm} (which contains 427 realistic images with \(80:20\) training-testing split) through random cropping and rotation, amassing 9070 \(256 \times 256\) images, with 8525 designated for training and 545 for testing. In this study, due to computational restrictions and fair comparison, we resize images to \(256\times256\) when training.

\noindent\textbf{Vessel Segmentation (VS):} We evaluate performance on three open-source retinal vessel segmentation datasets: DRIVE \cite{staal-2004-drive}, CHASE\_DB1 \cite{fraz-2012-chasedb1}, and STARE \cite{hoover-2000-stare}. DRIVE includes 40 $584\times565$ retinal vessel images (7 with abnormal pathology), split evenly for training and testing. CHASE\_DB1 contains 28 $999\times960$ retina images from 14 subjects; we follow \cite{liu-2022-frunet} with a 20/8 train/test division. STARE provides 20 $700\times605$ blood vessel images from retinal fundus, using an 80/20 split.

\noindent \textbf{Defect Detection (DD):} We utilize two open-source metal defect detection datasets: NEU-Seg \cite{dong-2019-neuseg} and SD-saliency-900 \cite{song-2020-MCITF}. NEU-Seg dataset containing 4470 $200\times 200$ surface defects from strip steel plates with different illumination and material changes, divided into \(3630:840\) train and test split. For SD-saliency-900, it contains 900 $200\times 200$ defect images of three different types: scratches, inclusion, and patches. Following work \cite{song-2020-edrnet}, we set the train with 810 images (540 images with equally distributed three defect types, 270 salt-and-paper disturbed images with $\phi_{salt\&paper}\!=\!0.2$) and test split with 900 images.

\subsubsection{Learning Schedule}
Our experiments are conducted in a PyTorch environment, leveraging an Nvidia GeForce RTX 3090 GPU. We utilize an Adam \cite{kingma-2014-adam} optimizer with a polynomial decay policy. The learning rate is adjusted by a factor of $\left(1 - \frac{iter}{total\_iter}\right)^{0.9}$. 

For \textbf{IRSTD} tasks, our training regimen spans 800 epochs for NUDT-SIRST \cite{li-2023-dnanet}, IRSTD-1K \cite{zhang-2022-isnet}, and SIRST \cite{dai-2021-acm}, and 400 epochs for SIRST-Aug \cite{zhang-2023-agpc}, commencing with a learning rate of $10^{-4}$, and we proceed with a batch size of 8. In \textbf{VS} tasks, our training regimen spans 400 epochs for DRIVE \cite{staal-2004-drive}, CHASE\_DB1 \cite{fraz-2012-chasedb1}, and STARE \cite{hoover-2000-stare}, commencing with a learning rate of $5 \times 10^{-4}$ and proceeding with a batch size of 4. 
As to \textbf{DD}, our training regimen spans 200 epochs for NEU-Seg \cite{dong-2019-neuseg} and 400 epochs for SD-saliency-900 \cite{song-2020-MCITF} with a learning rate of $10^{-4}$. 
\begin{figure}[!t]
\setlength{\abovecaptionskip}{1pt}
\setlength{\belowcaptionskip}{1pt}
\centering
   \includegraphics[width=\linewidth]{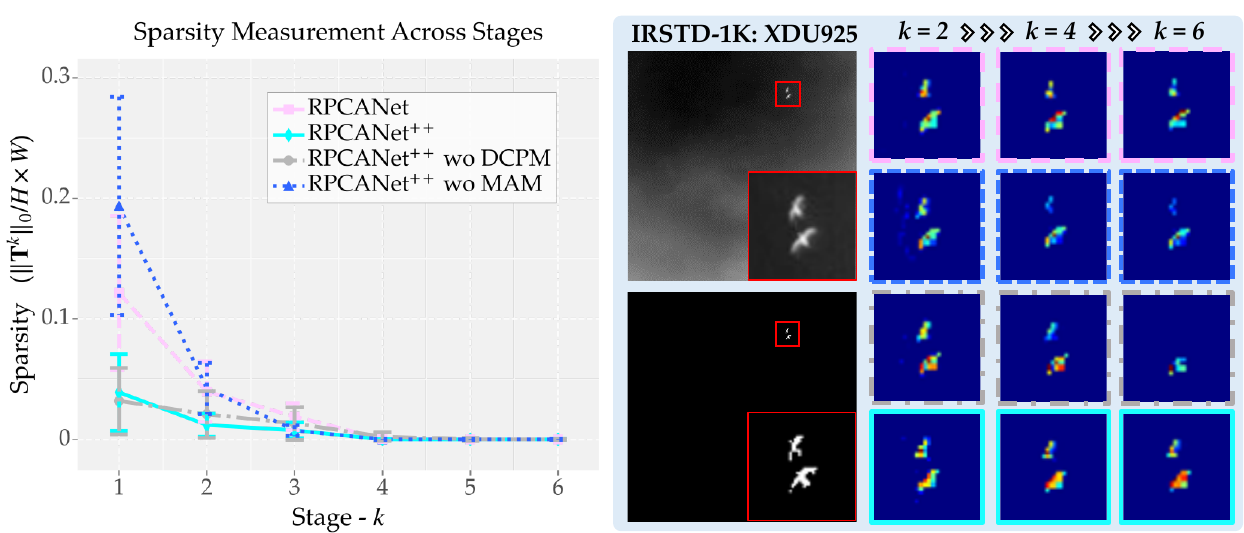}
   \caption{\textbf{Sparsity verification} of different stages our RPCANet$^{++}$ and its variants(without MAM or DCPM) vs RPCANet \cite{wu-2024-rpcanet} on IRSTD-1K \cite{zhang-2022-isnet}. \textbf{Left:} numerical verification. \textbf{Right:} heatmaps among different stages.}
   \label{fig:inters}
   \vspace{-0.35cm}
\end{figure}
\begin{table}[t]
\setlength{\abovecaptionskip}{1pt}
\setlength{\belowcaptionskip}{1pt}
\caption{The impact of different stage index $K$ on detection efficacy, quantified by IoU (\%), $F_1$ (\%), $P_d$ (\%), and $F_a$ ($10^{-5}$) on NUDT-SIRST dataset \cite{li-2023-dnanet} and SIRST-Aug \cite{zhang-2023-agpc}.}
\footnotesize
\centering
\renewcommand\arraystretch{1.1}
\resizebox{\linewidth}{!}{
\begin{tabular}{cIIcccccIcccc}
\hline\thickhline
\rowcolor{gray!20}&&\multicolumn{4}{cI}{\textbf{NUDT-SIRST \cite{li-2023-dnanet}}}& \multicolumn{4}{c}{\textbf{SIRST-Aug} \cite{zhang-2023-agpc}}\\\cline{3-6}\cline{7-10}
\rowcolor{gray!20} \multirow{-2}{*}{\textbf{Stages ($K$)}}&\multirow{-2}{*}{\textbf{Params (M)}} & IoU $\uparrow$ & $F_1$ $\uparrow$ & $P_d$ $\uparrow$ & $F_a$  $\downarrow$& IoU $\uparrow$ & $F_1$ $\uparrow$ & $P_d$ $\uparrow$ & $F_a$  $\downarrow$ \\ \hline\hline
\rowcolor{RoyalBlue!1}1    &0.447&   86.69  & 92.87 & 96.93&3.17&72.35&83.96&99.03&41.94  \\ 
\rowcolor{RoyalBlue!3}2     & 0.941 &  90.61    &  95.08 & 96.72  & 2.40&73.79&84.92&98.62&38.28   \\ 
\rowcolor{RoyalBlue!5}3      &1.435  &  92.70   & 96.21  & 98.31  & 1.67&73.90  &  85.01   & 98.90 &28.64  \\
\rowcolor{RoyalBlue!8}4      & 1.928 &  93.01  & 96.38 & 98.09   &3.35&74.49&85.38&96.84&26.81  \\ 
\rowcolor{RoyalBlue!10}5      & 2.422 &  93.52   & 96.65  &98.20  & 1.74&73.87&84.97&98.07&29.63 \\ 
\rowcolor{RoyalBlue!12}6      & 2.915   &  94.39   &  97.12 & 98.41    &  1.34&74.89 & 85.44    &  98.76    & 28.00 \\ 
\rowcolor{RoyalBlue!15}7      & 3.409 &  93.50   & 96.64  & 98.20 & 1.76&73.31&84.60&98.35&33.11 \\ 
\rowcolor{RoyalBlue!20}8      & 3.902 &   93.82  & 96.81  &  98.41  &  1.35&74.14&85.15&99.04&35.80  \\
\rowcolor{RoyalBlue!25}9      & 4.396 &   93.97  & 96.89  &  98.41  &  1.44&73.13&  84.48  &   97.66  &  32.18 \\
\hline\thickhline
\end{tabular}}
\label{tab:stages}
\vspace{-0.2cm}
\end{table}
\subsubsection{Metrics}
\label{sec:4.1.3}
For evaluation purposes, we primarily employ pixel-level metrics such as intersection over union (IoU) and the \(F\)-measure (\(F_1\)) for three different tasks. We also pick up field-adopted metrics: For \textbf{IRSTD} tasks, we adopt target-level metrics \cite{ying-2023-lesps}, including the probability of detection (\(P_d\)) and false alarm rate (\(F_a\)). Besides, the receiver operating characteristics (ROC) curves and the area under curves (AUC) are also introduced for performance measurement. In \textbf{VS} tasks \cite{liu-2022-frunet}, we utilize accuracy (Acc), sensitivity (Sen), specificity (Spe), and AUC for evaluation. For the \textbf{DD} tasks \cite{song-2020-MCITF}, we introduce structural similarity (S-measure) and mean absolute error (MAE) metrics.

Moreover, considering that the singular values of infrared images may approach zero within certain ranks, numerous optimization-based IRSTD methods integrate these low-rank properties during preliminary analysis, such as assessing the low-rankness across different modes. However, no prior works have incorporated this feature examination into the final verification stage. In this vein, we introduce a novel low-rankness metric to evaluate the background approximation process. Specifically, for a \(k\)-stage background \(\mathbf{B}^k\), we decompose it as \(\mathbf{B}^k = \mathbf{U} \Sigma \mathbf{V}^\top\), where \(\Sigma\) denotes the singular value matrix:
\begin{equation}
\setlength{\abovedisplayskip}{2pt}
\setlength{\belowdisplayskip}{2pt}
    \Sigma_{ij} = lr_i \cdot \delta_{ij}, \quad \delta_{ij} = 
    \begin{cases} 
    1 & \text{if } i = j, \\
    0 & \text{if } i \neq j\enspace,
    \end{cases}
\end{equation}
where \( \Sigma_{ij} \) signifies the element in the \( i \)-th row and \( j \)-th column of the diagonal matrix, and \( lr_i \) represents the \( i \)-th singular value, with \( i \) bounded by \(\min(H,W)\). And we adopt corresponding ranks' \( lr_i \) as our low-rankness index.

Furthermore, to quantify the sparse component of a target component, as inspired by \cite{yu-2024-whitebox}, we adopt the \(l_0\)-norm for sparsity evaluation, defining the sparsity rate \(r_s\) as:
\begin{equation}
\setlength{\abovedisplayskip}{1pt}
\setlength{\belowdisplayskip}{1pt}
    r_s = {\|\mathbf{O}^k\|_0}/{{H \times W}}.
\end{equation}

These metrics facilitate the evaluation of our framework's efficiency and interoperability. An easy-to-use toolkit containing these is open-sourced on our \href{https://github.com/fengyiwu98/RPCANet}{Github Repository}.
\begin{figure}[!t]
\setlength{\abovecaptionskip}{1pt}
\setlength{\belowcaptionskip}{1pt}
\centering
   \includegraphics[width=\linewidth]{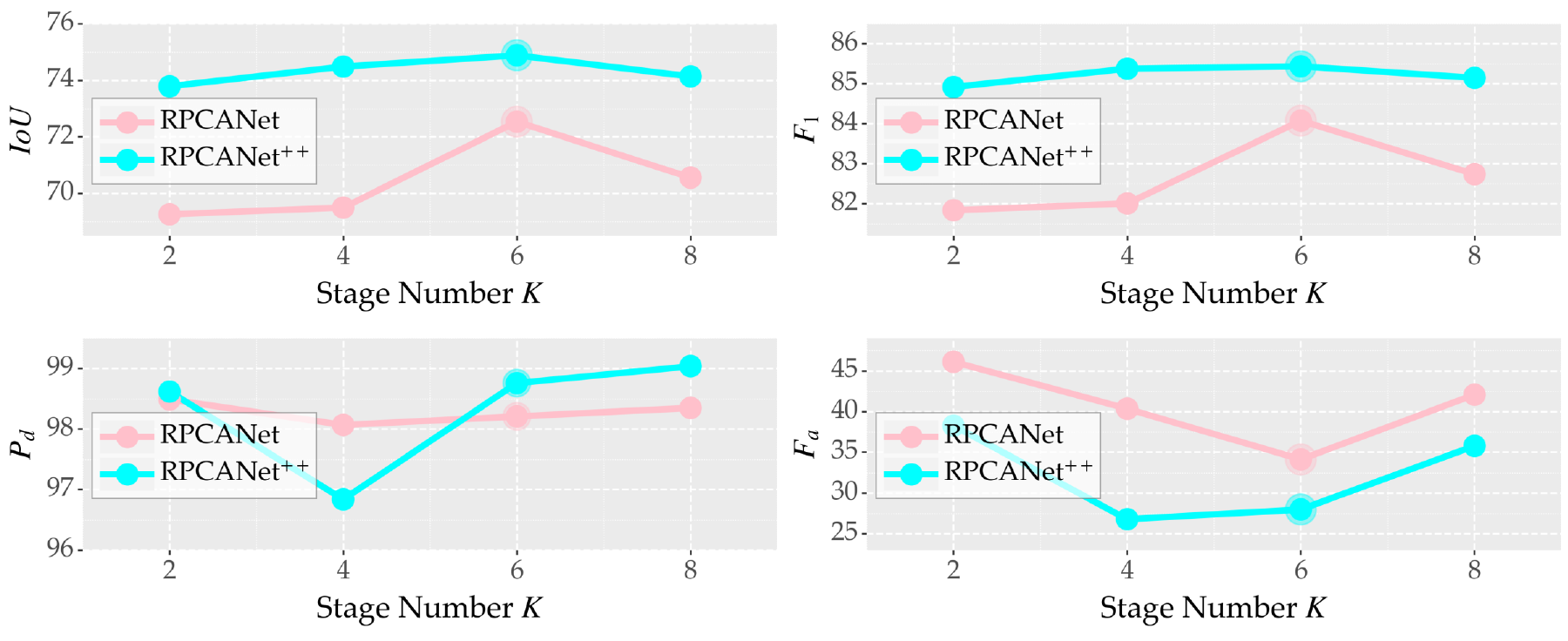}
   \caption{IoU (\%), $F_1$ (\%), $P_d$ (\%), and $F_a$ ($10^{-5}$) of different stages number $K$ on RPCANet$^{++}$ and RPCANet \cite{wu-2024-rpcanet} on SIRST-Aug \cite{zhang-2023-agpc}. }
   \label{fig:stages}
   \vspace{-0.4cm}
\end{figure}
\begin{figure}[!t]
\setlength{\abovecaptionskip}{2pt}
\setlength{\belowcaptionskip}{2pt}
\centering
   \includegraphics[width=0.98\linewidth]{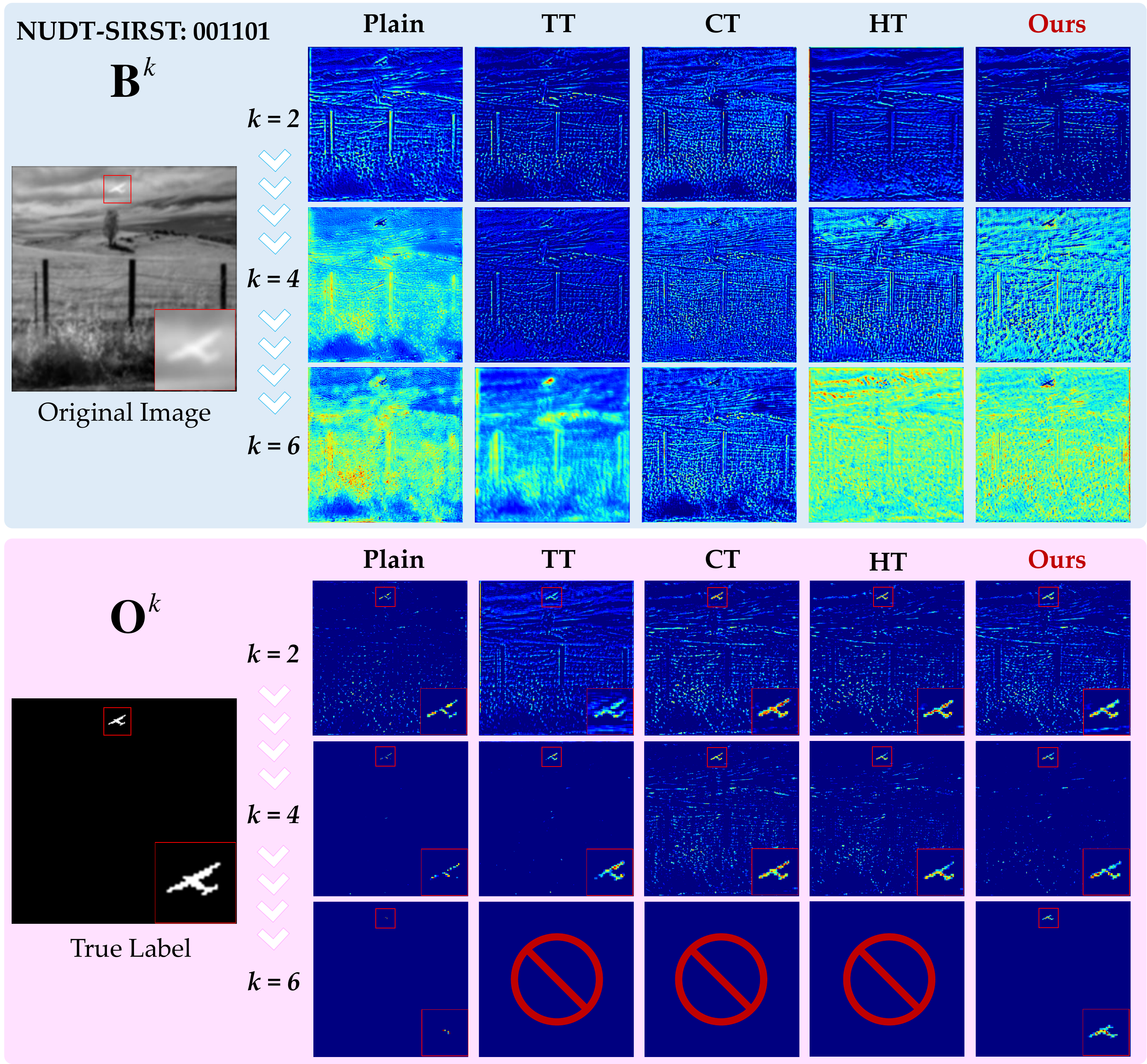}
   \caption{Target and background heatmaps comparison in different stages $k=2,4,6$ from different memory-augmented strategies on a NUDT-SIRST \cite{li-2023-dnanet} example. MAM can gradually assist background feature learning while maintaining sparse targets.}
   \label{fig:abl_mam}
   \vspace{-0.4cm}
\end{figure}
\setlength{\abovecaptionskip}{2pt}
\setlength{\belowcaptionskip}{2pt}
\begin{table}[t]
\caption{The impact of different memory augmented modules on detection efficacy, quantified by IoU (\%), $F_1$ (\%), $P_d$ (\%), and $F_a$ ($10^{-5}$) on NUDT-SIRST dataset \cite{li-2023-dnanet}.}
\footnotesize
\renewcommand\arraystretch{1.1}
\centering
\begin{tabular}{rIIccccc}
\hline\thickhline
\rowcolor{gray!20}\textbf{Method} &\textbf{Params (M)}& IoU $\uparrow$ & $F_1$ $\uparrow$ & $P_d$ $\uparrow$ & $F_a$  $\downarrow$ \\ \hline\hline
Plain   &  2.518 &  92.33   & 96.02 & 97.35 &1.70  \\ 
\rowcolor{gray!10}TT \cite{zhang-2023-ctnet}   &  2.518 &  91.52 &95.57  & 96.85  & 1.93      \\ 
CT  \cite{zhang-2023-ctnet}   &  2.524 &  91.70   & 95.67  &  97.67 & 1.95  \\
\rowcolor{gray!10}HT \cite{song-2021-madun}    & 2.970 &  91.87  & 95.76 &  97.67  & 2.31 \\ \hline
w/o MAM     &  2.518   &  92.62  & 96.17 & 96.93    &  1.97    \\
\rowcolor{RoyalBlue!12}w MAM     &  2.915   &  94.39   &  97.12 & 98.41    &  1.34    \\ 
\hline\thickhline
\end{tabular}
\label{tab:mam}
\vspace{-0.35cm}
\end{table}

\subsection{Model Verification}
In the context of deep unfolding, the network is architected to iteratively yield guided results congruent with an algorithm’s unrolled stages. Demonstrating outcomes at each stage is vital for model validation. Fig. \ref{fig:mv} showcases heatmaps from a six-stage RPCANet$^{++}$ at $k= 2,4,6$. Initially, the background predominantly reflects low-level edge information. With advancing stages, it progressively encapsulates detailed features, including intensity and non-local details. In contrast, sparse maps reveal a gradual reduction of residuals, where elements such as edges and target-mimicking false alarms are incrementally suppressed, steered by the guided mask. Consequently, the target contour sharpens, and our results—more aligned with the original images—furnish an enhanced approximation to the ground truth compared to preliminary stages, as illustrated in Fig. \ref{fig:mv}(d).

As outlined in Section \ref{sec:4.1}, the proposed model’s effectiveness is substantiated by the progressive validation of its low-rankness and sparsity. Evaluations conducted on IRSTD-1K \cite{zhang-2022-isnet} affirm this assertion. Fig. \ref{fig:interl} illustrates the stages of low-rankness evolution, with the orange line depicting the singular value distribution of the original images.
Our advanced RPCANet$^{++}$ demonstrates consistent feature enhancement as processing depth increases, with all stages converging to zero, signifying achieved low-rankness. In contrast, the DCPM’s isolated application tends to overestimate the background, as indicated by the 6$_{\text{th}}$ layer’s mean singular value (shown in grey) surpassing that of the original image. Conversely, the MAM addresses this by orderly increasing learned features through cascaded enhancement, and the integrated RPCANet$^{++}$ preserves this advantage. A comparative analysis with RPCANet, which also exhibits background overestimation as depicted in Fig. \ref{fig:interl}(d), highlights the importance of MAM in correcting background transmission errors.

Additionally, the right side of Fig. \ref{fig:inters} presents a sparsity comparison, detailing the mean and standard deviation across various stages. The memory-augmented module yields a markedly sparser outcome than RPCANet, swiftly pinpointing potential target areas. Despite the DCPM-enhanced model achieving greater sparsity after three stages relative to RPCANet, its initial stage measurements reveal considerable uncertainty.
Our RPCANet$^{++}$ effectively integrates the strengths of MAM and DCPM, enabling swift and reliable target segmentation, and achieving significant sparsity that approaches convergence rapidly. Moreover, as shown in the left of Fig. \ref{fig:inters}, RPCANet$^{++}$ demonstrates minimal noise initially and superior shape retention in later stages, unlike its counterparts.
In summary, our model, supported by both visual and quantitative evidence, conforms to theoretical expectations, manifesting low-rank and sparse properties. We envisage our method setting the stage for an interpretable paradigm in the IRSTD field.

\subsection{Ablation Studies}
\subsubsection{Impact of Stage Numbers}
The number of iterative stages is crucial to the performance of our algorithm. Similar to optimization problems, an insufficient number of stages can lead to non-convergence, while an excessive number of iterations can result in computational redundancy. Therefore, this section examines the impact of the number of stages on RPCANet$^{++}$, as illustrated in Table. \ref{tab:stages}. Due to computational memory constraints, we limit the maximum number of stages to nine. Results indicate that the overall performance improves with an increasing number of stages, showing a linear enhancement in parameters and peaking at $K=6$. We adopt this setting for the subsequent experiments. 

In addition, we also compare different stage number comparisons on RPCANet$^{++}$ and RPCANet in Fig. \ref{fig:stages}. We can observe that both DUN methods have detection performance increases as the stage climbs, where both of them have an optimal value when $K=6$. To be specific, despite a slight off in $P_d$ when $K=4$, RPCANet$^{++}$ has all stages better IoU, $F_1$, and $F_a$ than RPCANet, proving the enhancement proposed in this study.


\begin{table}[t]
\caption{The impact of different feature extractors on detection efficacy, quantified by IoU (\%), $F_1$ (\%), $P_d$ (\%), and $F_a$ ($10^{-5}$) on NUDT-SIRST dataset \cite{li-2023-dnanet}.}
\footnotesize
\renewcommand\arraystretch{1.1}
\centering
\begin{tabular}{rIIccccc}
\hline\thickhline
\rowcolor{gray!20}\textbf{Method} &\textbf{Params (M)}& IoU $\uparrow$ & $F_1$ $\uparrow$ & $P_d$ $\uparrow$ & $F_a$  $\downarrow$ \\ \hline\hline
 $l_\mathbf{B}=1$   & 2.805 &  92.26&  95.97  & 98.09 & 1.91  \\ 
\rowcolor{gray!10}$l_\mathbf{B}=2$    & 2.917 &  92.27    &  95.98 & 98.09  &  2.23  \\ 
$l_\mathbf{B}=3$     &  3.029&  92.02   &  95.84 &  97.14 & 2.19  \\
\rowcolor{gray!10} two RBs    &  3.137 &   93.65 & 96.72 &  97.98  & 1.41 \\ 
\rowcolor{RoyalBlue!12}single RB   &  2.915   &  94.39   &  97.12 & 98.41    &  1.34    \\ 
\hline\thickhline
\end{tabular}
\label{tab:fe}
\vspace{-0.25cm}
\end{table}
\begin{table}[t]
\caption{The impact of different DCPM configurations on detection efficacy, quantified by IoU (\%), $F_1$ (\%), $P_d$ (\%), and $F_a$ ($10^{-5}$) on NUDT-SIRST dataset \cite{li-2023-dnanet}.}
\footnotesize
\centering
\renewcommand\arraystretch{1.1}
\resizebox{\linewidth}{!}{
\begin{tabular}{rIIccccc}
\hline\thickhline
\rowcolor{gray!20}\textbf{Method} &\textbf{Params (M)}& IoU $\uparrow$ & $F_1$ $\uparrow$ & $P_d$ $\uparrow$ & $F_a$  $\downarrow$ \\ \hline\hline
base (w/o DCPM)  & 1.132&  90.99  &  95.24& 96.51& 2.50 \\ 
\rowcolor{gray!10}base + CDC \cite{yu-2020-cdc}  & 2.911 &  91.74    & 95.69  &  97.35 & 1.76   \\ 
base + SE \cite{hu-2018-senet}  &  1.138&   90.58   & 95.06  & 97.46  &   2.39 \\ 
\rowcolor{gray!10}base + CBAM \cite{woo-2018-cbam}  & 1.139 & 91.78     &  95.71 &  98.09 &2.56    \\ \hline
base + DCPM (s = 3)    & 1.195 &   93.15  & 96.46  & 97.35  & 1.58  \\
\rowcolor{gray!10}base + DCPM (s = 5)    &  1.293&  93.15   &  96.46 & 98.41  &  2.01 \\
base + DCPM (s = 9)    & 1.637 & 93.43    &  96.60 &  98.20 & 1.62  \\
\rowcolor{RoyalBlue!12} base + DCPM (Ours)   & 2.915   &  94.39   &  97.12 & 98.41    &  1.34  \\ 
\hline\thickhline
\end{tabular}}
\label{tab:dcpm}
\vspace{-0.25cm}
\end{table}
\begin{figure}[!t]
\setlength{\abovecaptionskip}{0pt}
\setlength{\belowcaptionskip}{2pt}
\centering
   \includegraphics[width=\linewidth]{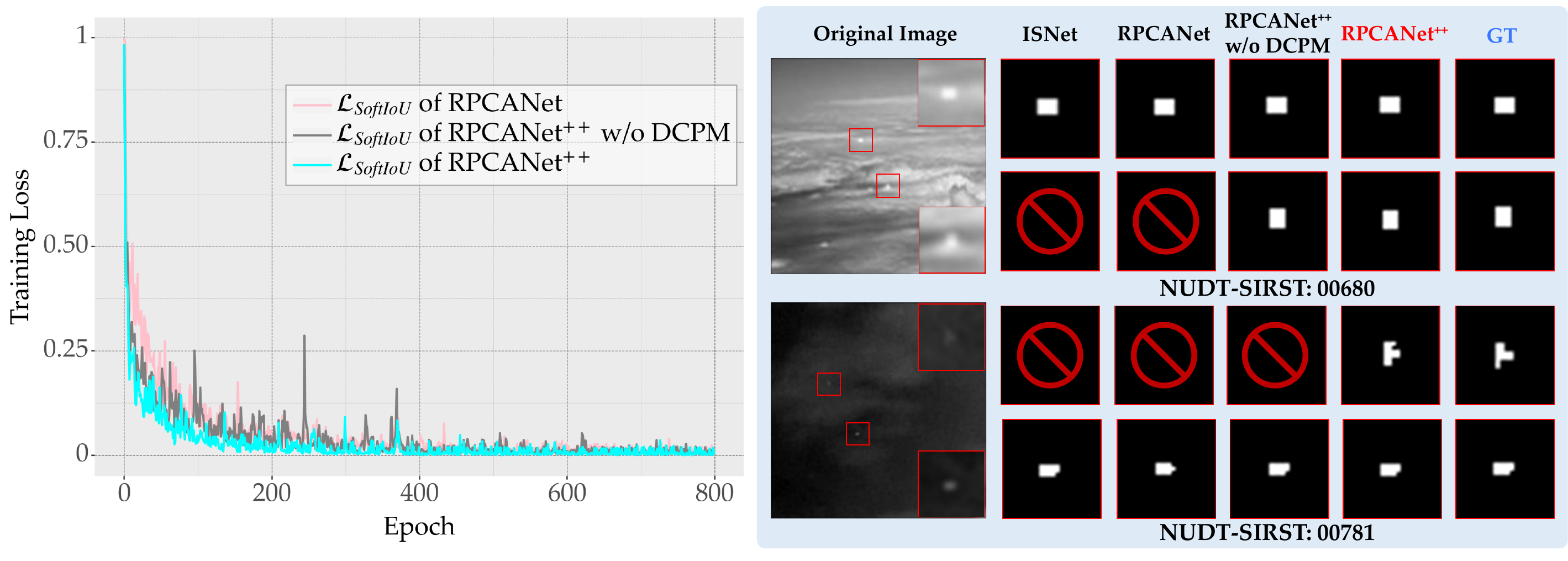}
   \caption{\textbf{Left:} Traning loss comparison of RPCANet and RPCANet$^{++}$ w/wo DCPM. \textbf{Right:} Comparison of target results with different methods in tricky environments from NUDT-SIRST \cite{li-2023-dnanet}.}
   \label{fig:abl_dcpm}
   \vspace{-0.5cm}
\end{figure}
\begin{table}[t]
\caption{The impact of different OEM configurations on detection efficacy, quantified by IoU (\%), $F_1$ (\%), $P_d$ (\%), and $F_a$ ($10^{-5}$) on NUDT-SIRST dataset \cite{li-2023-dnanet}.}
\footnotesize
\renewcommand\arraystretch{1.1}
\centering
\begin{tabular}{rIIccccc}
\hline\thickhline
\rowcolor{gray!20} \textbf{Method} &\textbf{Params (M)}& IoU $\uparrow$ & $F_1$ $\uparrow$ & $P_d$ $\uparrow$ & $F_a$  $\downarrow$ \\ \hline\hline
 $l_\mathbf{O}=2$   &2.693 & 92.27  &95.98  &97.99 &1.96 \\ 
\rowcolor{gray!10} $l_\mathbf{O}=4$  &2.804 &  92.89 & 96.31 &97.46 &1.50 \\ 
 $l_\mathbf{O}=8$  &3.026 &  92.68 & 96.20 &97.35 &1.64 \\ 
\rowcolor{gray!10} $l_\mathbf{O}=10$  & 3.137&  92.60 &96.16  &98.10 &1.94 \\  \hline
$\mathbf{O}^k$   & 2.915 & 92.49  &96.10  &97.67 &1.59  \\
\rowcolor{gray!10}Concate   &2.917 & 93.27  & 96.52 & 98.31& 1.58\\ \hline
\rowcolor{RoyalBlue!12} Ours   & 2.915   &  94.39   &  97.12 & 98.41    &  1.34  \\ 
\thickhline
\end{tabular}
\label{tab:tem}
\vspace{-0.3cm}
\end{table}
\begin{table}[t]
\caption{The impact of different Loss configurations on detection efficacy, quantified by IoU (\%), $F_1$ (\%), $P_d$ (\%), and $F_a$ ($10^{-5}$) on NUDT-SIRST dataset \cite{li-2023-dnanet}.}
\footnotesize
\renewcommand\arraystretch{1.1}
\centering
\resizebox{\linewidth}{!}{
\begin{tabular}{rIIccccIcccc}
\hline\thickhline
\rowcolor{gray!20}& \multicolumn{4}{cI}{\textbf{RPCANet }\cite{wu-2024-rpcanet}}& \multicolumn{4}{c}{\textbf{RPCANet$^{++}$}} \\ \cline{2-9}
\rowcolor{gray!20}\multirow{-2}{*}{\textbf{Method}}&IoU $\uparrow$ & $F_1$ $\uparrow$ & $P_d$ $\uparrow$ & $F_a$  $\downarrow$ &IoU $\uparrow$ & $F_1$ $\uparrow$ & $P_d$ $\uparrow$ & $F_a$  $\downarrow$\\ \hline\hline
 $\sigma=0$   &88.23  &93.75  &96.40 &3.23 &93.78&96.79&98.10&1.42\\ 
\rowcolor{gray!10} $\sigma=0.01$  &  87.37 & 93.26 &96.40 &3.81 &93.51&96.64&98.52&1.75\\ 
 $\sigma=0.05$  & 88.58 & 93.94 &96.30 &3.04&93.75&96.78&97.99&1.41 \\ 
\rowcolor{gray!10}$\sigma=0.5$   &  86.84  &92.96  &95.66 &3.53 &93.25&96.51&97.88&1.61 \\
$\sigma=1$   & 86.67  & 92.84 & 96.19& 3.75&94.08&96.95&98.20&1.12\\
\rowcolor{RoyalBlue!12} $\sigma=0.1$&89.31&94.35& 97.14&2.87&  94.39   &  97.12 & 98.41    &  1.34  \\ 
\hline\thickhline
\end{tabular}}
\label{tab:loss}
\vspace{-0.25cm}
\end{table}

\begin{table}[t]
\setlength{\abovecaptionskip}{0pt} 
\setlength{\belowcaptionskip}{0pt} 
\caption{Studies on different component of RPCANet$^{++}$ on the detection performance in IoU ($\%$), $F_1$ ($\%$), $P_d$ ($\%$), and $F_a$ ($10^{-5}$) on SIRST-Aug \cite{zhang-2023-agpc}.}
\centering
\renewcommand\arraystretch{1.1}
\scriptsize{
\resizebox{\linewidth}{!}{
\begin{tabular}{cIIcccccIccccc}
\hline\thickhline
\rowcolor{gray!20}\textbf{Config.}&\textbf{OEM}&\textbf{BAM}&\textbf{IRM}&\textbf{MAM}&\textbf{DCPM}&\textbf{Params}& IoU $\uparrow$ & $F_1$ $\uparrow$ & $P_d$ $\uparrow$ & $F_a$ $\downarrow$  \\ \hline\hline
1&$\checkmark$ & \bcancel{\checkmark} &$\usym{2717}$&$\usym{2717}$& $\usym{2717}$ & 0.507 &60.48 & 75.38& 92.98& 53.55 \\
\rowcolor{gray!10}2&$\checkmark$ & $\checkmark$ & $\usym{2717}$&$\usym{2717}$& $\usym{2717}$ & 0.507&61.99 &76.53 & 98.21&37.50  \\
3&$\checkmark$ & $\checkmark$ & $\checkmark$&$\usym{2717}$& $\usym{2717}$ &0.734 & 68.75 & 81.48& 96.15& 33.78  \\
\rowcolor{gray!10}4&$\checkmark$ & $\checkmark$ & $\checkmark$&$\checkmark$& $\usym{2717}$ & 1.132& 72.30& 83.92&96.42& 40.42\\
5&$\checkmark$ & $\checkmark$ & $\checkmark$&$\usym{2717}$& $\checkmark$ & 2.518&73.60& 84.79& 96.42& 27.66 \\
\rowcolor{RoyalBlue!12} 6&$\checkmark$ & $\checkmark$ & $\checkmark$&$\checkmark$&$\checkmark$ & 2.915 &\textbf{74.89} &\textbf{85.44}   &  98.76    & 28.00  \\ 
\hline\thickhline
\end{tabular}
}}
\label{tab:ind}
\vspace{-0.3cm}
\end{table}
\subsubsection{Different Construction of MAM}
In this section, we discuss the implementation of handling transmission loss for $\mathbf{B}$ during iteration. Section \ref{sec:3.3} outlines various enhancement strategies, with visual representations provided in Fig. \ref{fig:bam}. This ablation study contrasts different memory augmentation methods: Temporary Transmission (TT) \cite{zhang-2023-ctnet}, Cumulative Transmission (CT) \cite{zhang-2023-ctnet}, High-Throughput (HT) -injected MAM \cite{song-2021-madun}, and our optimized MAM, detailed in Table. \ref{tab:mam}.

Compared to the baseline, alternative methodologies negatively impact the IoU and \(F_1\) scores. Although there is a slight improvement in \(P_d\) for CT and HT, the overall advancements remain suboptimal, underscoring the efficacy of our MAM module's configuration. Additionally, we offer a stage-by-stage visual representation of the background/target evolution in Fig. \ref{fig:abl_mam}. The background enhancements indicate that the direct addition of antecedent elements, as seen in TT and CT, limits the approximation to elementary features such as edges and dispersed points. This limitation can inadvertently trigger incorrect target-background delineations, resulting in potential target omission, as shown at the bottom of Fig. \ref{fig:abl_mam}. On the other hand, incorporating HT elements could lead to similar issues. In contrast, MAM in our RPCANet$^{++}$ methodically integrates background details while maintaining target fidelity, leading to a significant increase in IoU by 2.06 and reducing $F_a$ by 0.36 compared to the baseline.

\subsubsection{Discussion on Feature Extractor in MAM}
Given that the deep extractor within the proximal network, denoted as $\text{proxNet}(\cdot)$, determines the information flow to the MAM, selecting an appropriate feature extractor is crucial. In our investigation (as summarized in Table. \ref{tab:fe}), we explore various extractors, including $l_\mathbf{B}$ layers of $3\times 3$ convolutional layers (where $l_\mathbf{B}=1,2,3$); two residual blocks (RB); and our single RB configuration. The results reveal that residual construction outperforms single CNN layer construction, consistent with \cite{wu-2024-rpcanet}. Additionally, an excessive number of extraction layers may compromise feature representation. Specifically, the performance of $l_{\mathbf{B}}= 3$ or two RBs is slightly weaker than that of the single RB construction. Consequently, we adopt a single RB before and after the MAM in our architecture.

Meanwhile, as shown in Table. \ref{tab:dcpm}, considering the impact of the receptive field on feature extraction and computational efficiency, we investigate different kernel sizes ($s=3,5,9$) with our 17-setting. As kernel size increases, overall performance improves, reaching its peak at $s\!=\!17$. Specifically, compared to $s\!=\!9$, we select $s\!=\!17$ for its +0.94 IoU and 17.28\% fewer $F_a$.  

In addition, to demonstrate the acceleration of the “convergence" process by our DCPM, we compare training loss and performance among three variants: RPCANet, our RPCANet$^{++}$ without DCPM, and RPCANet$^{++}$. While RPCANet$^{++}$ without DCPM exhibits faster loss reduction, it suffers from unstable fluctuations. In contrast, RPCANet$^{++}$ with DCPM achieves robustness and steady convergence. Additionally, under strong noisy conditions (as depicted Fig. \ref{fig:abl_dcpm}), our DCPM-enhanced RPCANet$^{++}$ captures targets with minimal edge information, outperforming ISNet \cite{zhang-2022-isnet} and RPCANet \cite{wu-2024-rpcanet}. Notably, only our DCPM preserves clear target shapes despite reduced lightness, highlighting its efficacy and overall architectural robustness.

\subsubsection{Discussion on Basic Construction of OEM}
In this section, we perform ablation studies on the inner construction of $\nabla\mathcal{S}$. Specifically, we explore different numbers of layers for OEM, denoted as $l_{\mathbf{O}}$, where $l_{\mathbf{O}}\!=\!\{2, 4, 6, 8, 10\}$. As summarized in Table. \ref{tab:tem}, detection performance improves with increasing layers in the initial stages. However, beyond $l_{\mathbf{O}}\!=\!6$, overall detectability begins to decline. Thus, we select the optimal configuration of $l_{\mathbf{O}}\!=\!6$ for our RPCANet$^{++}$.

Meanwhile, as discussed in Section \ref{sec:3.3}, the flexibility of DUNs allows us to leverage $\mathbf{D}^{k-1}$ and update $\mathbf{B}^{k-1}$ to enhance adjacent stage information while avoiding target transmission loss. However, evaluating the performance is essential when using only $\mathbf{O}^{k-1}$. Table. \ref{tab:tem} illustrates that this simplified construction achieves satisfactory results but experiences a -1.90 IoU performance compared to our enhanced approach. Furthermore, we compare our “Plus" operation with feature concatenation (“Concate") in Table. \ref{tab:tem}, demonstrating that the former strategy maintains superior performance. Nevertheless, the excellent performance of both approaches underscores the robustness and effectiveness of our overall framework.

\subsubsection{Discussion on Loss Functions}
Instead of solely introducing a constraint to the target and background with $\lambda$, we also introduce a $\sigma$ value in the loss function. As $\lambda$ has become a learnable variable across stages, it's essential to discuss the value of $\sigma$ regarding the effectiveness of introducing background constraints and how to weigh it. In Table. \ref{tab:loss}, we compare the value of $\sigma\!=\!0$ (no $\mathcal{L}_{\text{MSE}}$), 0.01, 0.05, 0.1, 0.5, and 1 on RPCANet and RPCANet$^{++}$. We can observe that solely adopting segmentation loss can achieve relatively excellent results, and properly constraining the background can assist detection performance, e.g., $\sigma\!=\!0.05$ for RPCANet, $\sigma\!=\!1$ for RPCANet$^{++}$. In contrast, when $\sigma\!=\!0.1$, both methods achieve their best, with +1.08 and +0.61 in IoU; thus, we choose $\sigma\!=\!0.1$ as our loss constraint.

\subsubsection{Discussion on Individual Module in RPCANet$^{++}$}
Lastly, we present a step-by-step ablation study on each component in RPCANet$^{++}$ on SIRST-Aug \cite{zhang-2023-agpc}, similar to \cite{wu-2024-rpcanet}. As depicted in Table. \ref{tab:ind}, we list six different configurations with five different components. Here, OEM means without DCPM, and BAM means without MAM. Check and cross marks mean with and without the corresponding module, and a half checkmark in Config. 1 means without loss constraint in background regularization as discussed in Table. \ref{tab:loss}. Typically, we do not eliminate the background module as the RPCA formulation requires an integration. Also, in Config. 1 and 2, the image restoration is realized by $\bf{D} = \bf{B} + \bf{O}$. We can see that increasingly implementing each module can improve the detection metrics by 1.51, 8.27, 11.82, 13.12, and 14.41 in IoU (\%), proving the effectiveness of each module. Also, we can see that individually introduced MAM and DCPM can assist the detection, by 3.55 and 4.85, where solely adopting the local prior can receive a relatively low false alarm.

\begin{figure*}[t]
\setlength{\abovecaptionskip}{2pt} 
\setlength{\belowcaptionskip}{0pt} 
    \centering
    \includegraphics[width=\linewidth]{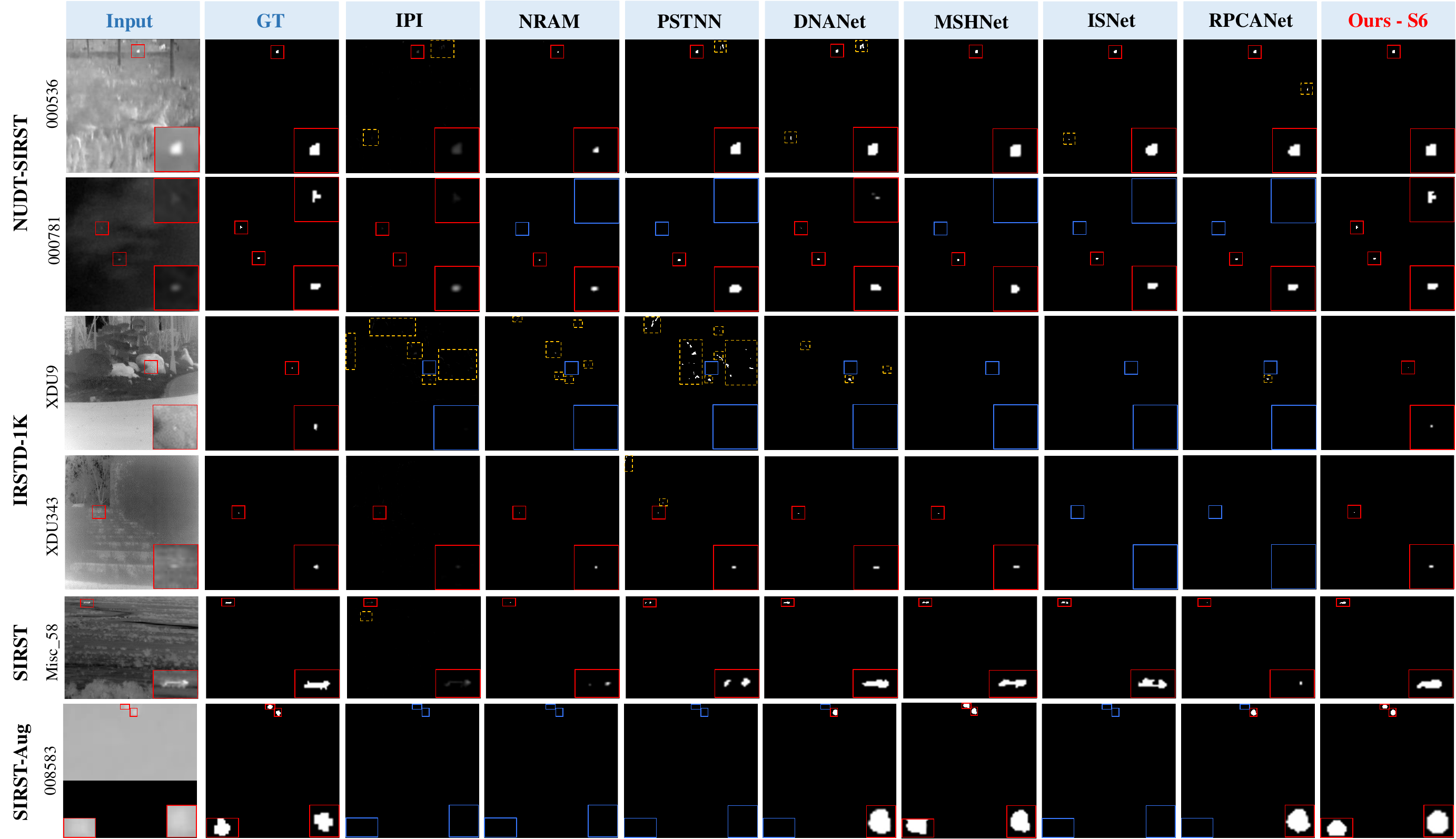}
    \caption{Visual comparisons of diverse IRSTD techniques from NUDT-SIRST \cite{li-2023-dnanet}, IRSTD-1K \cite{zhang-2022-isnet}, SIRST \cite{dai-2021-acm}, and SIRST-Aug \cite{zhang-2023-agpc} are presented, with correctly identified targets, undetected targets, and false positives delineated by \textcolor{red}{red}, \textcolor{blue}{blue}, and \textcolor{yellow}{yellow} dot boxes, respectively.}
    \label{fig:pres1}
    \vspace{-0.2cm}
\end{figure*}
\begin{table*}[!ht] 
\setlength{\abovecaptionskip}{1pt} 
\setlength{\belowcaptionskip}{1pt} 
\caption{Performance metrics including IoU (\%), $F_1$ (\%), $P_d$ (\%), $F_a$ ($10^{-5}$), and runtime are evaluated for various methods on datasets NUDT-SIRST \cite{li-2023-dnanet}, IRSTD-1K \cite{zhang-2022-isnet}, SIRST \cite{dai-2021-acm}, and SIRST-Aug \cite{zhang-2023-agpc}. Venue. means the publication and year. Params. (M) is the parameter statistic for data-driven approaches within the second-to-last column. [Zoom in for a better view]} 
\centering 
\renewcommand\arraystretch{1.1} 
\scriptsize{ 
\resizebox{\linewidth}{!}{ 
\setlength\tabcolsep{3.pt} 
\begin{tabular}{rIIcIccccIccccIccccIccccIcc} 
\hline\thickhline 
\rowcolor{gray!20} 
 & & \multicolumn{4}{cI}{\textbf{NUDT-SIRST} \cite{li-2023-dnanet}} 
& \multicolumn{4}{cI}{\textbf{IRSTD-1K} \cite{zhang-2022-isnet}} 
& \multicolumn{4}{cI}{\textbf{SIRST}\cite{dai-2021-acm}} 
& \multicolumn{4}{cI}{\textbf{SIRST-Aug} \cite{zhang-2023-agpc}} 
& \textbf{Params} & \textbf{Time} (s) \\ 
\cline{3-6} \cline{7-10} \cline{11-14} \cline{15-18} 
\rowcolor{gray!20} 
\multirow{-2}{*}{\textbf{Methods}} 
& \multirow{-2}{*}{\textbf{Venue}} 
& IoU $\uparrow$ & $F_1$ $\uparrow$ & $P_d$ $\uparrow$ & $F_a$ $\downarrow$ 
& IoU $\uparrow$ & $F_1$ $\uparrow$ & $P_d$ $\uparrow$ & $F_a$ $\downarrow$ 
& IoU $\uparrow$ & $F_1$ $\uparrow$ & $P_d$ $\uparrow$ & $F_a$ $\downarrow$ 
& IoU $\uparrow$ & $F_1$ $\uparrow$ & $P_d$ $\uparrow$ & $F_a$ $\downarrow$ 
& (M) & CPU/GPU \\ 
\hline \hline 
\multicolumn{20}{l}{\textcolor{gray!60}{\textit{Model-Based Methods}}} \\ 
MPCM \cite{wei-2016-mpcm} & PR$^{\textbf{16}}$ & 9.26 & 16.95 & 70.58 & 32.72 & 15.70 & 27.13 & 62.54 & 12.47 & 22.30 & 36.47 & 90.83 & 5.56 & 19.76 & 33.00 & 93.39 & 3.14 & - & 0.0441/- \\ 
\rowcolor{gray!10} 
IPI \cite{gao-2013-ipi} & TIP$^{\textbf{13}}$ & 34.62 & 51.43 & 92.38 & 7.54 & 25.27 & 40.35 & 83.51 & 20.93 & 44.83 & 61.90 & 97.25 & 3.76 & 21.93 & 35.97 & 80.05 & \textbf{1.62} & - & 5.0583/- \\ 
NRAM \cite{zhang-2018-nram} & RS$^{\textbf{18}}$ & 11.44 & 20.52 & 65.71 & 2.35 & 17.97 & 30.46 & 59.79 & 5.71 & 26.39 & 41.75 & 88.07 & \textbf{1.60} & 10.41 & 18.86 & 76.06 & 2.26 & - & 1.5439/- \\ 
\rowcolor{gray!10} 
PSTNN \cite{zhang-2019-pstnn} & RS$^{\textbf{19}}$ & 25.17 & 40.22 & 80.21 & 7.61 & 24.39 & 39.22 & 65.29 & 15.77 & 35.17 & 52.04 & 88.99 & \underline{2.57} & 12.38 & 22.04 & 60.79 & \underline{1.98} & - & 0.2863/- \\ 
\hdashline 

\multicolumn{20}{l}{\textcolor{gray!60}{\textit{Deep Learning-Based Methods}}} \\ 
ACM \cite{dai-2021-acm} & WACV$^{\textbf{21}}$ & 69.46 & 81.98 & 97.14 & 13.11 & 54.16 & 70.25 & 83.84 & 7.32 & 66.77 & 80.07 & 98.16 & 11.03 & 67.79 & 80.80 & 95.87 & 32.52 & 0.398 & -/0.0146 \\ 
\rowcolor{gray!10} 
AGPCNet \cite{zhang-2023-agpc} & TAES$^{\textbf{23}}$ & 85.02 & 91.90 & 97.88 & 4.34 & 60.44 & 75.34 & 90.72 & 6.10 & 68.19 & 81.08 & \underline{99.08} & 12.09 & 72.07 & 83.77 & 98.62 & 29.39 & 12.36 & -/0.1173 \\ 
DNANet \cite{li-2023-dnanet} & TIP$^{\textbf{23}}$ & 87.73 & 93.47 & 97.67 & 5.52 & 62.23 & 76.72 & \underline{92.44} & 5.64 & 73.12 & 84.48 & \textbf{100.0} & 10.26 & 70.56 & 82.74 & 96.56 & 36.49 & 4.697 & -/0.0437 \\ 
\rowcolor{gray!10} 
UIUNet \cite{wu-2023-uiunet} & TIP$^{\textbf{23}}$ & 88.94 & 94.15 & 95.23 & 1.53 & 63.08 & 77.36 & 92.10 & 6.09 & 72.15 & 83.82 & 98.16 & 7.90 & 70.76 & 82.88 & 96.70 & 34.30 & 50.54 & -/0.0541 \\ 
MSHNet \cite{liu-2024-mshnet} & CVPR$^{\textbf{24}}$ & 75.96 & 86.34 & 95.23 & 8.28 & \underline{64.83} & \underline{78.67} & \textbf{92.78} & 7.14 & 64.64 & 78.52 & 98.16 & 15.66 & 71.93 & 83.67 & \textbf{99.03} & 32.06 & 4.065 & -/0.0483 \\ 
\rowcolor{gray!10} 
ALCNet \cite{dai-2021-alcnet} & TGRS$^{\textbf{21}}$ & 80.75 & 89.35 & 97.56 & 9.56 & 60.69 & 75.53 & 85.57 & 5.17 & 70.31 & 82.57 & \underline{99.08} & 11.36 & 67.38 & 80.51 & 97.94 & 28.68 & 0.378 & -/0.0134 \\ 
ISNet \cite{zhang-2022-isnet} & CVPR$^{\textbf{22}}$ & 87.05 & 93.08 & 96.83 & 4.05 & 63.00 & 77.29 & 88.66 & 5.52 & 66.96 & 80.20 & 96.33 & 9.85 & 71.10 & 83.11 & 97.66 & 30.33 & 0.967 & -/0.0270 \\\hline
\multicolumn{20}{l}{\textcolor{gray!60}{\textit{Deep Unfolding-Based Methods}}} \\
RPCANet \cite{wu-2024-rpcanet} & WACV$^{\textbf{24}}$ & 89.31 & 94.35 & 97.14 & 2.87 & 63.21 & 77.45 & 88.31 & 4.39 & 70.37 & 82.61 & 95.41 & 7.42 & 72.54 & 84.08 & 98.21 & 34.14 & 0.680 & -/0.0217 \\ 
\rowcolor{RoyalBlue!5} 
\textbf{Ours - S3} & - & 92.70{\tiny{\textcolor{VioletRed}{+3.39}}} & 96.21{\tiny{\textcolor{VioletRed}{+1.86}}} & \underline{98.31} & 1.67 & 64.40{\tiny{\textcolor{VioletRed}{+1.19}}} & 78.34{\tiny{\textcolor{VioletRed}{+0.98}}} & \textbf{92.78} & 5.56 & \textbf{75.52}{\tiny{\textcolor{VioletRed}{+5.15}}} & \textbf{86.06}{\tiny{\textcolor{VioletRed}{+3.45}}} & \textbf{100.0} & 8.70 & \underline{73.90}{\tiny{\textcolor{VioletRed}{+1.36}}} & \underline{85.01}{\tiny{\textcolor{VioletRed}{+0.93}}} & \underline{98.90} & 28.64 & 1.435 & -/0.0262 \\ 
\rowcolor{RoyalBlue!12} 
\textbf{Ours - S6} & - & \textbf{94.39}{\tiny{\textcolor{VioletRed}{+5.08}}} & \textbf{97.12}{\tiny{\textcolor{VioletRed}{+2.77}}} & \textbf{98.41} & \textbf{1.34} & \textbf{64.93}{\tiny{\textcolor{VioletRed}{+1.72}}} & \textbf{78.73}{\tiny{\textcolor{VioletRed}{+1.28}}} & 89.70 & \underline{4.35} & \underline{74.76}{\tiny{\textcolor{VioletRed}{+4.39}}} & \underline{85.47}{\tiny{\textcolor{VioletRed}{+2.86}}} & \textbf{100.0} & 10.77 & \textbf{74.89}{\tiny{\textcolor{VioletRed}{+2.35}}} & \textbf{85.44}{\tiny{\textcolor{VioletRed}{+1.36}}} & 98.76 & 28.00 & 2.915 & -/0.0471 \\ 
\rowcolor{RoyalBlue!25} 
\textbf{Ours - S9} & - & \underline{93.97}{\tiny{\textcolor{VioletRed}{+4.66}}} & \underline{96.89}{\tiny{\textcolor{VioletRed}{+2.54}}} & \textbf{98.41} & \underline{1.44} & 63.23{\tiny{\textcolor{VioletRed}{+0.02}}} & 77.48{\tiny{\textcolor{VioletRed}{+0.03}}} & 89.35 & \textbf{4.28} & 72.62{\tiny{\textcolor{VioletRed}{+2.25}}} & 84.14{\tiny{\textcolor{VioletRed}{+1.53}}} & \textbf{100.0} & 9.57 & 73.13{\tiny{\textcolor{VioletRed}{+0.59}}} & 84.48{\tiny{\textcolor{VioletRed}{+0.04}}} & 97.66 & 32.18 & 4.396 & -/0.0627 \\ 
\hline\thickhline 
\end{tabular} }} 
\label{tab:baseline} 
\vspace{-0.4cm}
\end{table*}
\subsubsection{Discussion and Comparison on DCPM}
In this study, to fit into domain knowledge, we introduce a novel deep contrast prior module (DCPM) to enhance target allocation while accelerating the “convergence" process of optimization. To validate the effectiveness of our DCPM, we compare it with the original CDC \cite{yu-2020-cdc} (kernel size of 17 $\times$ 17) and related efficient feature extractors (SE \cite{hu-2018-senet} and CBAM \cite{woo-2018-cbam}). Our findings in Table. \ref{tab:dcpm} indicates that the “central difference" strategy slightly improves detection performance (+0.75 in IoU) while reducing false alarms by 30\% ($F_a$). However, our learnable parameter $\theta$, despite a slight increase in parameters, significantly enhances overall IoU by 2.65. Although SE and CBAM have fewer parameters, they do not match the effectiveness of our DCPM.

\begin{table}[]
\setlength{\abovecaptionskip}{0pt} 
\setlength{\belowcaptionskip}{0pt} 
    \centering
    \caption{AUC performance (\%) of various methods on four datasets.}
    \renewcommand\arraystretch{1.1}
    \scriptsize{
    \resizebox{\linewidth}{!}{
    \begin{tabular}{rIIcIcIcIc}
\hline\thickhline
\rowcolor{gray!20}        \textbf{Method} & \textbf{NDUT-SIRST} \cite{li-2023-dnanet} & \textbf{IRSTD-1K} \cite{zhang-2022-isnet} & \textbf{SIRST} \cite{dai-2021-acm} & \textbf{SIRST-Aug} \cite{zhang-2023-agpc} \\ \hline\hline
         NRAM \cite{zhang-2018-nram}&60.71 &68.69 &73.96 & 60.21\\
\rowcolor{gray!10}         PSTNN \cite{zhang-2019-pstnn} &74.73&82.48&81.44&60.47\\
         AGPCNet \cite{zhang-2023-agpc}&96.88&90.62&92.90&92.81\\
\rowcolor{gray!10}         DNANet \cite{li-2023-dnanet} &97.96 & 81.10& 89.51& 89.25 \\
         ALCNet \cite{dai-2021-alcnet}& 97.88& 82.80& 94.27&88.90  \\
\rowcolor{gray!10}         ISNet \cite{zhang-2022-isnet}&96.36& 89.90&85.12 &70.00 \\ \hline
         \multicolumn{5}{l}{\textcolor{gray!60}{\textit{Deep unfolding-Based Methods}}} \\
         RPCANet \cite{wu-2024-rpcanet} & 96.50& 88.58& 91.58& 96.01 \\
         
         \rowcolor{RoyalBlue!5} \textbf{Ours-S3}&\underline{98.83}{\tiny{\textcolor{VioletRed}{+2.33}}} & \textbf{96.73}{\tiny{\textcolor{VioletRed}{+8.15}}}& 
         \textbf{99.99}{\tiny{\textcolor{VioletRed}{+8.41}}}&\textbf{99.43}{\tiny{\textcolor{VioletRed}{+3.42}}}  \\ 
         
         \rowcolor{RoyalBlue!12} \textbf{Ours-S6}&\textbf{99.27}{\tiny{\textcolor{VioletRed}{+2.77}}} & \underline{95.04}{\tiny{\textcolor{VioletRed}{+6.46}}}& \underline{99.31}{\tiny{\textcolor{VioletRed}{+7.73}}}&95.52{\tiny{\textcolor{VioletRed}{-0.49}}} \\
         
         \rowcolor{RoyalBlue!25} \textbf{Ours-S9}&98.60{\tiny{\textcolor{VioletRed}{+2.10}}} & 90.19{\tiny{\textcolor{VioletRed}{+1.61}}}& 
         \textbf{99.99}{\tiny{\textcolor{VioletRed}{+8.41}}}&\underline{97.05}{\tiny{\textcolor{VioletRed}{+1.04}}} \\
\thickhline       
    \end{tabular}}}
    \label{tab:auc}
    \vspace{-0.4cm}
\end{table}
\begin{figure*}[htb]
\vspace{-0.3cm}
\setlength{\abovecaptionskip}{0pt} 
\setlength{\belowcaptionskip}{0pt} 
	\begin{minipage}{0.245\linewidth}
		\vspace{0pt}
		\centerline{\includegraphics[width=\textwidth]{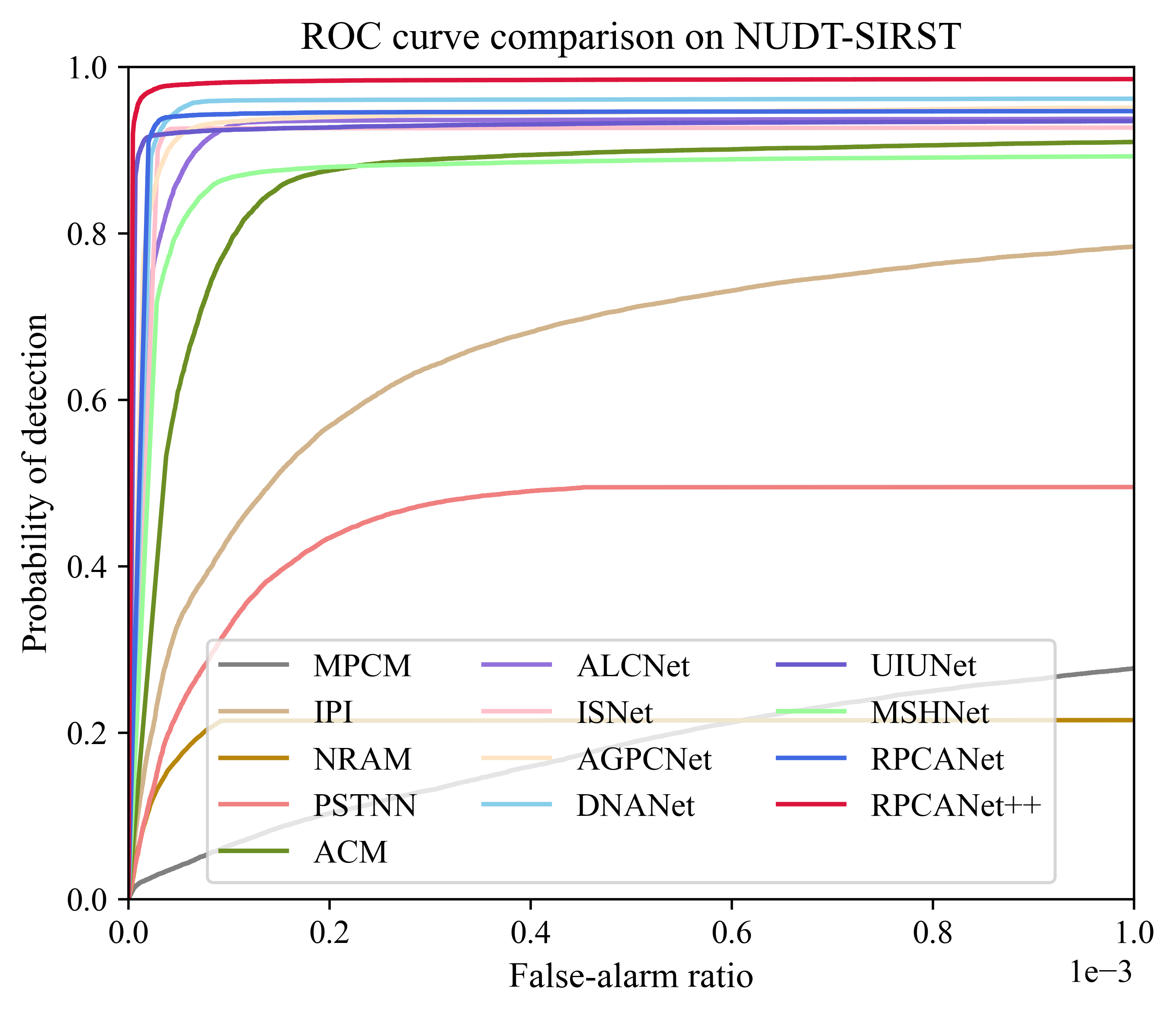}}
\subcaption{NUDT-SIRST \cite{li-2023-dnanet}}
	\end{minipage}
    \begin{minipage}{0.245\linewidth}
		\vspace{0pt}
		\centerline{\includegraphics[width=\textwidth]{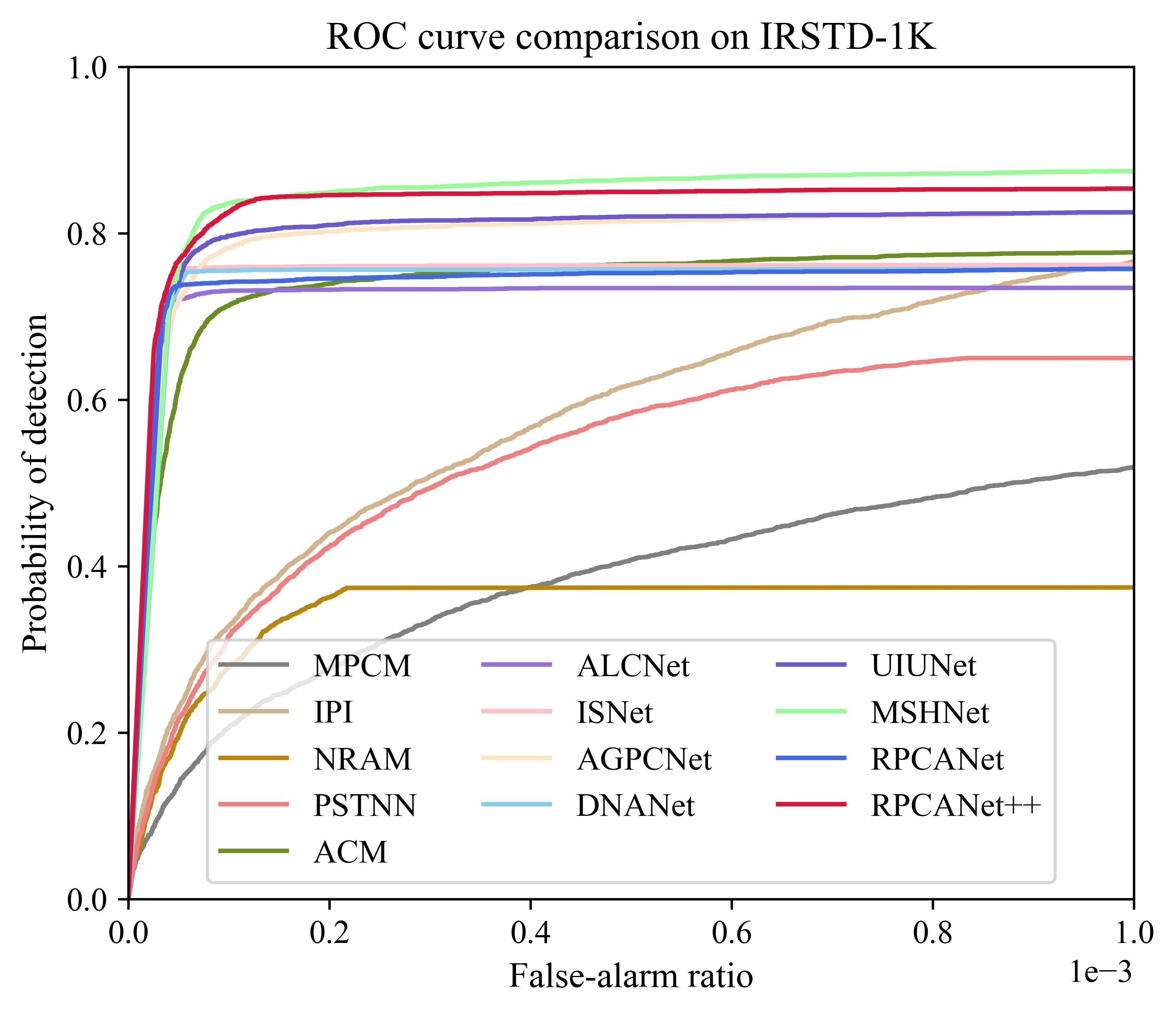}}
	\subcaption{IRSTD-1K \cite{zhang-2022-isnet}}
	\end{minipage}
	\begin{minipage}{0.245\linewidth}
		\vspace{0pt}
		\centerline{\includegraphics[width=\textwidth]{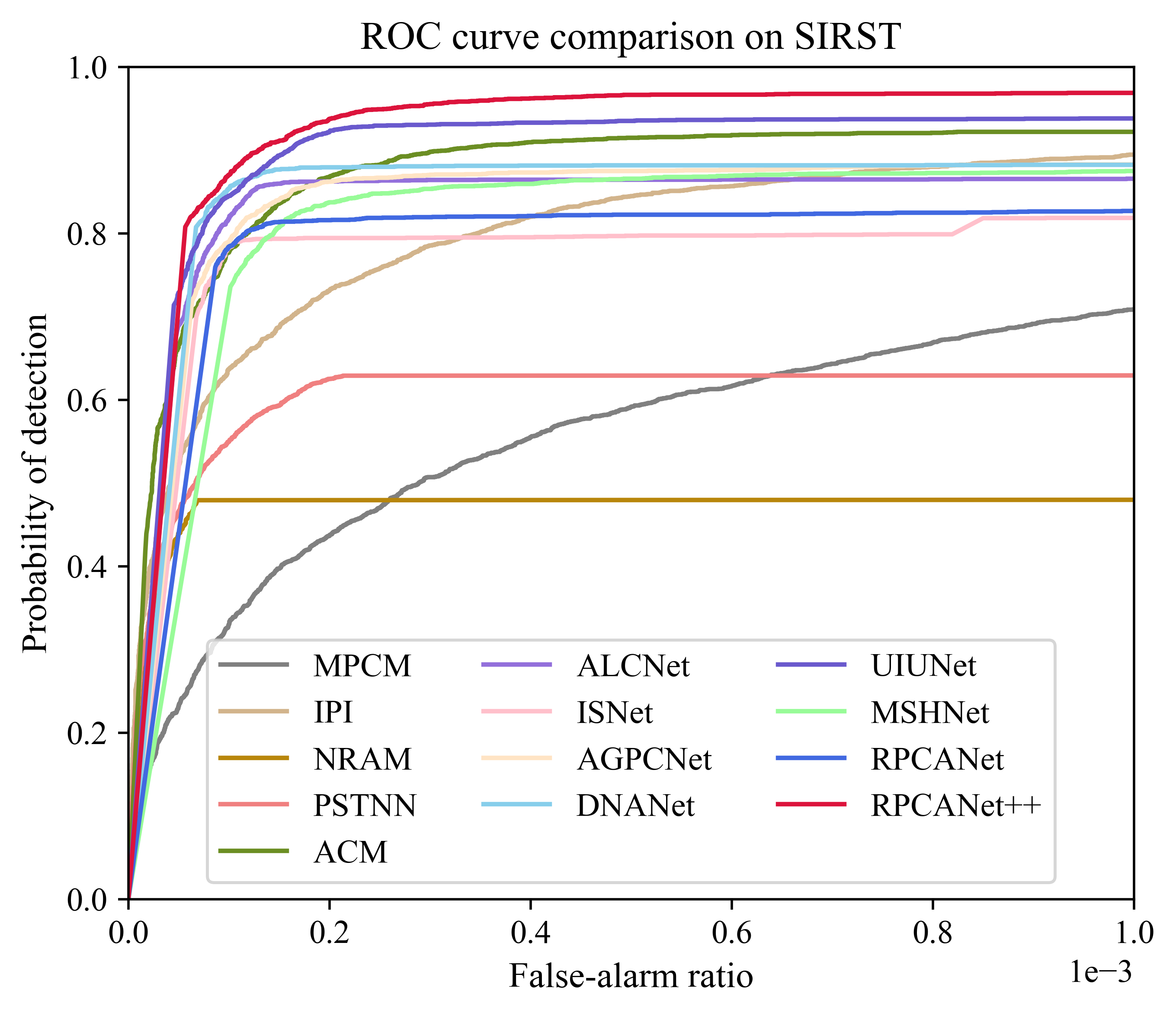}}	 	
\subcaption{ SIRST \cite{dai-2021-acm}}
	\end{minipage}
	\begin{minipage}{0.245\linewidth}
		\vspace{0pt}
		\centerline{\includegraphics[width=\textwidth]{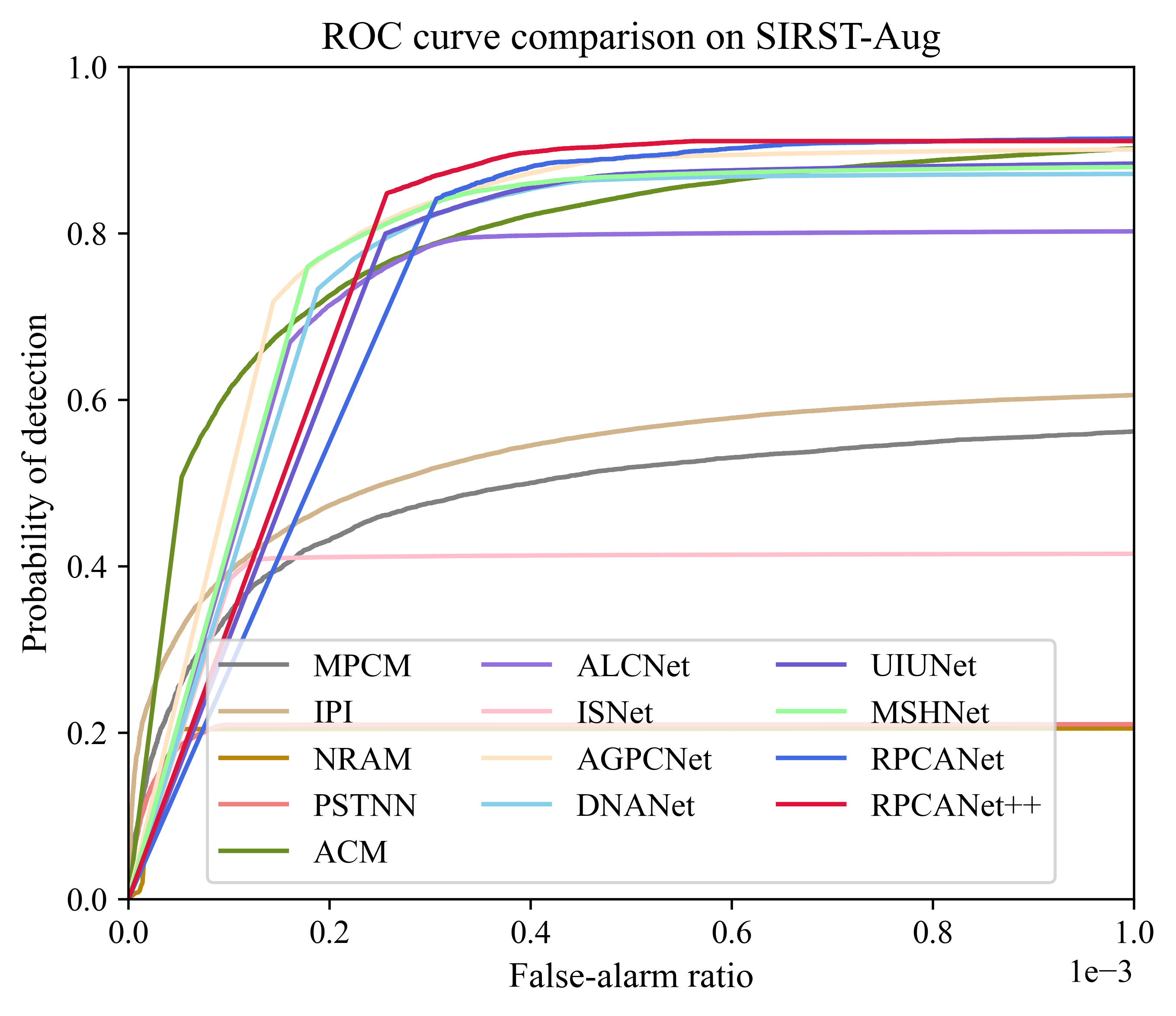}}	 	
\subcaption{SIRST-Aug \cite{zhang-2023-agpc}}
	\end{minipage}
	\caption{ROC curves for various algorithms across four datasets are depicted. RPCANet$^{++}$ is represented in \textcolor{red}{red} and RPCANet \cite{wu-2024-rpcanet} in \textcolor{blue!70}{blue}.}
	\label{fig_roc}
 \vspace{-0.3cm}
\end{figure*}
\begin{figure*}[t]
\setlength{\belowcaptionskip}{0pt}
\setlength{\abovecaptionskip}{1pt}
\centering
\includegraphics[width=\linewidth]{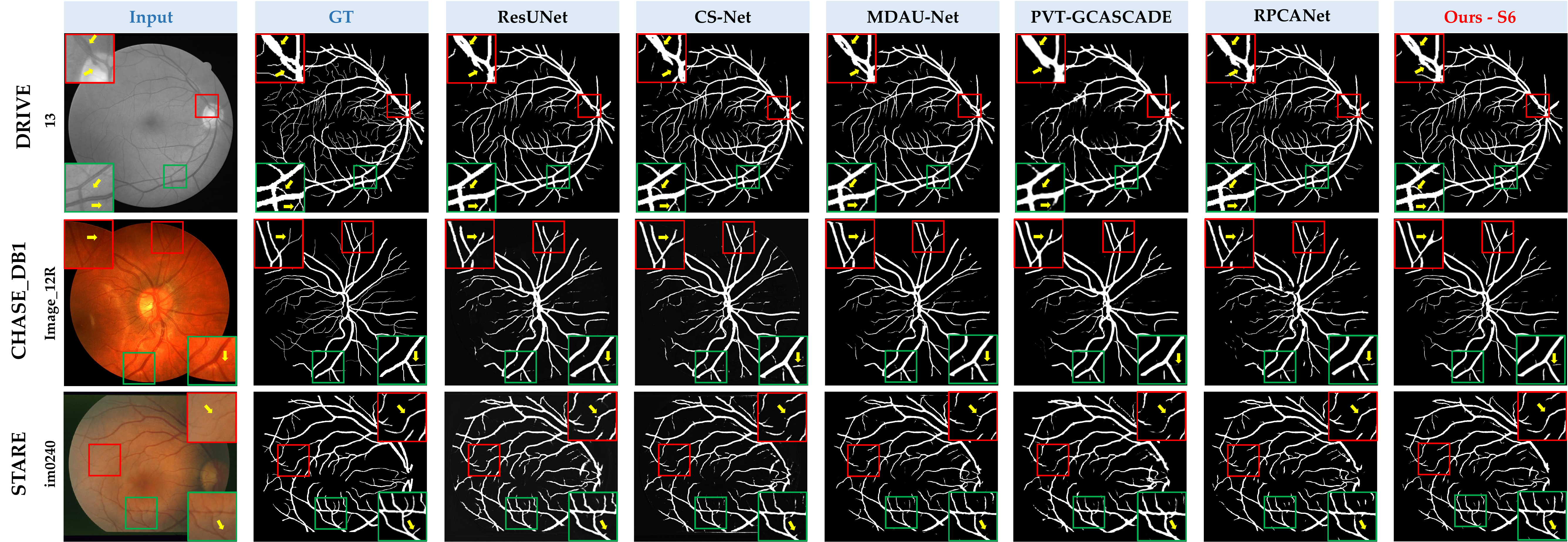}
\caption{Visual comparisons of diverse medical segmentation techniques from DRIVE \cite{drive-2014}, STARE \cite{hoover-2000-stare}, and CHASE$\_$DB1 \cite{chase_db1-2020}. We zoom in local challenging regions with \textcolor{green}{green} and \textcolor{red}{red} boxes. \textcolor{yellow}{Yellow} arrows indicate vital details of corresponding vessel features. [Zoom in for a better view]}
    \label{fig:medical}
     \vspace{-0.2cm}
\end{figure*}
\begin{table*}[!ht]
\setlength{\belowcaptionskip}{-2pt}
\setlength{\abovecaptionskip}{0pt}
\caption{Performance metrics including Acc (\%), Sen (\%), Spe (\%), AUC (\%), $F_1$ (\%), IoU (\%), and runtime are evaluated for various methods on datasets DRIVE \cite{drive-2014}, STARE \cite{hoover-2000-stare}, and CHASE$\_$DB1 \cite{chase_db1-2020}. Venue means the publication and year. Params. (M) is the parameter statistic for data-driven approaches within the second-to-last column. [Zoom in for a better view]}
\label{tab:medical}
\begin{threeparttable}
\centering
\renewcommand\arraystretch{1.1}
\small{
\resizebox{\linewidth}{!}{
\begin{tabular}{rIIcIccccccIccccccIccccccIcc}
\hline\thickhline
 \rowcolor{gray!20}&  & \multicolumn{6}{cI}{\textbf{DRIVE \cite{staal-2004-drive}}} & \multicolumn{6}{cI}{\textbf{STARE \cite{hoover-2000-stare}}} & \multicolumn{6}{cI}{\textbf{CHASE$\_$DB1 \cite{fraz-2012-chasedb1}}} &\textbf{Params} & \textbf{Time (s)} \\ \cline{3-20}
\rowcolor{gray!20}\multirow{-2}{*}{\textbf{Methods}}     &     \multirow{-2}{*}{\textbf{Venue}}            &Acc$\uparrow$ &Sen$\uparrow$&Spe$\uparrow$&AUC$\uparrow$& $F_1$$\uparrow$&IoU$\uparrow$&Acc$\uparrow$&Sen$\uparrow$&Spe$\uparrow$& AUC$\uparrow$&$F_1$$\uparrow$&IoU$\uparrow$&Acc$\uparrow$&Sen$\uparrow$&Spe$\uparrow$&AUC$\uparrow$&$F_1$$\uparrow$&IoU$\uparrow$&(M) & CPU/GPU\\ \hline\hline
\multicolumn{22}{l}{\textcolor{gray!60}{\textit{Deep Learning-Based Methods}}}\\
U-Net  \cite{ronneberger-2015-unet} & MICCAI$^{\textbf{15}}$   &  96.85   & 81.39&  \underline{82.73} &95.88 &81.85 & 69.30& 95.96&79.16 &71.22 &    \underline{96.69}  &  74.89  &59.91&    97.17&    84.48&   74.31& 96.04&79.02 &65.37 & 34.526 & -/0.0217  \\  
\rowcolor{gray!10}U-Net++ \cite{zhou-2018-unetplusplus}   & DLMIA$^{\textbf{18}}$     &   96.66 & \underline{84.42}&   79.13 & 95.56& 81.49&68.78 &95.27 &82.42 & 64.85&     94.43 & 72.25   & 56.70   &    96.28&  \textbf{90.80} & 64.66& \underline{97.47}& 75.49& 60.67& 2.294 & -/0.0154  \\ 
ResUNet \cite{li-2019-resunet}   & ICIP$^{\textbf{19}}$    &    96.83  &  82.87  &81.54 &95.98&82.01 &69.53 &96.33 & 78.26& 73.96&  95.07    &  75.97  &   61.30 &  96.95  & 84.61  & 72.07& 97.16&77.77 & 63.67& 0.905 & -/0.0125  \\  
\rowcolor{gray!10}Attention UNet \cite{oktay-2018-attnunet}  & arXiv$^{\textbf{18}}$      & 96.82    & 82.30&  81.84  & 94.05& 81.86& 69.31& 89.71& \textbf{91.30}& 40.98&    94.35  &  56.35  &    39.41&   96.32 & \underline{89.20}  & 65.27& 96.71&75.33 &60.47 & 2.184 & -/0.0137  \\ 
CS-Net  \cite{mou-2019-csnet} & MICCAI$^{\textbf{19}}$    &   96.70  & \textbf{84.68}&   79.34 & 95.23& 81.70& 69.08& 96.35& 74.02&76.25 &     95.62 &   75.03 &   60.04 &   96.92 &  85.03 & 71.55&96.49 & 77.64& 63.48& 8.400 & -/0.0159  \\ 
\rowcolor{gray!10}FR-UNet \cite{liu-2022-frunet}  & JBHI$^{\textbf{22}}$    &  96.73   & 82.82&   80.69 & 95.55& 81.53& 68.85&  92.48& \underline{85.70}&    50.04  & 94.42  & 62.96&   46.05 &   97.04 &  80.41 & 74.72& 93.43& 77.36& 63.13& 5.719 & -/0.0347  \\ 
MCDAU-Net \cite{zhou-2023-MCDAUNet}  & CIBM$^{\textbf{23}}$   & \textbf{96.91}   & 81.44&   \textbf{83.22} & 95.75& 82.10& 69.67&96.38 &75.75 & 75.65&  95.32    &   75.58 &   60.77 &   \underline{97.26} &  84.35 &75.26 &95.49 & \underline{79.46}& \underline{65.96}& 12.979 & -/0.0343  \\ 
\rowcolor{gray!10}PVT-GCASCADE \cite{rahman-2024-gcascade}  & WACV$^{\textbf{24}}$    &   95.88  &79.71 &   75.07 & \underline{97.78}& 77.12& 62.79&96.41 &77.45 &75.53 &     \textbf{97.31} & 76.23   &61.60    &  96.94  &  85.69 &71.57 & \textbf{98.74}& 77.94&63.87 & 26.628 & -/0.0411  \\\hline
\multicolumn{22}{l}{\textcolor{gray!60}{\textit{Deep Unfolding-Based Methods}}}\\
RPCANet  \cite{wu-2024-rpcanet}  & WACV$^{\textbf{24}}$   &   \underline{96.87}  & 82.81&  81.91  & 97.56&82.15 & 69.74&96.61 &77.08 &77.71 &   95.68   & 77.15  & 62.82   & 97.11   &  80.88 &\underline{75.46} &95.49 & 78.00& 64.00& 0.680 & -/0.0217  \\ 
\rowcolor{RoyalBlue!5}\textbf{Ours - S3}  & -  &   96.86 & 83.42  &   81.47 & 95.69 &  \underline{82.24}{\tiny{\textcolor{VioletRed}{+0.09}}}  &  \underline{69.86}{\tiny{\textcolor{VioletRed}{+0.12}}}&\textbf{96.80} &  75.86& \underline{80.19} & 93.41 &     \textbf{77.83}{\tiny{\textcolor{VioletRed}{+0.68}}}  &     \textbf{63.71}{\tiny{\textcolor{VioletRed}{+0.89}}}& 97.13 & 81.22 &  75.27&    94.94  &   78.06{\tiny{\textcolor{VioletRed}{+0.06}}}   &    64.09{\tiny{\textcolor{VioletRed}{+0.09}}}& 1.435 & -/0.0262  \\
\rowcolor{RoyalBlue!12} \textbf{Ours - S6}  & -  &  \textbf{96.91}   & 84.11 & 81.45   & 96.23& \textbf{82.58}{\tiny{\textcolor{VioletRed}{+0.43}}}  & \textbf{70.35}{\tiny{\textcolor{VioletRed}{+0.61}}}&\underline{96.79} &74.25 &\textbf{80.83} &   92.90   &  77.33{\tiny{\textcolor{VioletRed}{+0.18}}}  & 63.04{\tiny{\textcolor{VioletRed}{+0.22}}}    &   \textbf{97.35} & 84.41  & \textbf{76.23}&94.90 &\textbf{80.07}{\tiny{\textcolor{VioletRed}{+2.07}}}  & \textbf{66.81}{\tiny{\textcolor{VioletRed}{+2.81}}}& 2.915 & -/0.0471  \\
\rowcolor{RoyalBlue!25} \textbf{Ours - S9}  &-  &  96.85  & 82.95  & 81.71   & 91.87 & 82.12{\tiny{\textcolor{VioletRed}{-0.03}}}   & 69.69{\tiny{\textcolor{VioletRed}{-0.05}}}  & 96.73 & 75.82& 79.74 &   88.89   & \underline{77.59}{\tiny{\textcolor{VioletRed}{+0.44}}}   &  \underline{63.40}{\tiny{\textcolor{VioletRed}{+0.58}}}  &  97.03  &  81.56 & 74.11 & 90.68 & 77.59{\tiny{\textcolor{VioletRed}{-0.41}}}  & 63.45{\tiny{\textcolor{VioletRed}{-0.55}}}& 4.396 & -/0.0627  \\
\hline\thickhline
\end{tabular}}
}
\begin{tablenotes} 
\tiny
\item[*]Due to memory constraints, we adopt the initial channel number of 16 for UNet++, ResUNet, and Attention UNet. Similar to \cite{liu-2022-frunet}, we convert the color retinal images from three datasets to grayscale images.
\end{tablenotes}
\end{threeparttable}
\vspace{-0.4cm}
\end{table*}
\subsection{Compared to State-of-the-art Studies}
\subsubsection{Comparison on Infrared Small Target Detection Task}
In the comparison section, we benchmark our approach against several state-of-the-art IRSTD methodologies. For model-driven methods, we incorporate baselines: MPCM \cite{wei-2016-mpcm}, IPI \cite{gao-2013-ipi}, NRAM \cite{zhang-2018-nram}, and PSTNN \cite{zhang-2019-pstnn}. In the realm of data-driven approaches, we evaluate frameworks including ACM \cite{dai-2021-acm}, DNANet \cite{li-2023-dnanet}, AGPCNet \cite{zhang-2023-agpc}, UIUNet \cite{wu-2023-uiunet}, and MSHNet \cite{liu-2024-mshnet}; alongside model-inspired networks such as ALCNet \cite{dai-2021-alcnet}, ISNet \cite{zhang-2022-isnet}, and the DUN-based RPCANet \cite{wu-2024-rpcanet}. We retrained listed DL methods based on their original codes.

\noindent \textbf{Visual Comparison:}
Fig.~\ref{fig:pres1} displays mask results from four datasets under conditions of extremely low SCR, high false-alarm rates, and multiple targets in both terrestrial and aerial scenes. In the first row, dense target-like clusters cause most methods to produce numerous false alarms. In contrast, both the optimization-based NRAM and RPCANet\textsuperscript{++} maintain clean backgrounds, with RPCANet\textsuperscript{++} rendering the most precise target contours. In low-SCR scenarios (e.g., NUDT-SIRST 000781 and SIRST-Aug 008583), while many approaches either miss or merge targets, RPCANet\textsuperscript{++} consistently detects all targets with accurate shapes. For diminutive targets such as IRSTD-1K XUD9, where interference poses a significant challenge, our method markedly enhances detectability. Moreover, RPCANet\textsuperscript{++} more effectively distinguishes closely spaced targets (e.g., XDU406) compared to the original RPCANet (e.g., Misc\_58 in SIRST), and significantly improves shape preservation in cluttered backgrounds.

\noindent \textbf{Numerical Comparison:}  
Tables.~\ref{tab:baseline} and~\ref{tab:auc} summarize performances on four compared datasets. Model-based methods (e.g., MPCM, IPI, NRAM) generally yield low IoU and high false alarms, whereas deep learning approaches (e.g., ACM, AGPCNet, DNANet) improve performance at the cost of larger models. Our baseline RPCANet achieves 89.31\% IoU and 94.35\% $F_1$ on NUDT-SIRST. The RPCANet$^{++}$ series further boosts performance: the \textbf{RPCANet$^{++}$-S3} increases IoU to 92.70\% (+3.39\%) and $F_1$ to 96.21\% (+1.86\%), with similar trends on other datasets. Notably, the \textbf{RPCANet$^{++}$-S6} model reaches 94.39\% IoU and 97.12\% $F_1$ on NUDT-SIRST, while reducing false alarms (1.34 vs. 2.87). AUC values also improve significantly; for example, \textbf{RPCANet$^{++}$-S3} attains 98.83\% on NUDT-SIRST and 99.99\% on SIRST, and \textbf{S6} records 99.27\% on NUDT-SIRST. Overall, the RPCANet$^{++}$ series achieves enhanced sensitivity, accuracy, and lower false-alarm rates across IRSTD benchmarks while maintaining a compact and efficient model.

\noindent \textbf{Computational Efficiency:}  
As summarized in Table.~\ref{tab:baseline}, RPCANet\textsuperscript{++} incurs a moderate increase in parameters due to the added prior and memory-augmented modules, yet its model size remains substantially lower than UIUNet (50.54 M) and DNANet (4.697 M). Its execution time is under 0.05 s on a GPU, demonstrating high efficiency.

\noindent Collectively, the visual and quantitative results affirm the precision, false-alarm suppression, and efficiency of RPCANet$^{++}$, highlighting the advantages of integrating memory augmentation and deep contrast priors within a deep unfolding framework for IRSTD.

\begin{figure*}[t]
\setlength{\abovecaptionskip}{1pt} 
\setlength{\belowcaptionskip}{0pt} 
    \centering
    \includegraphics[width=\linewidth]{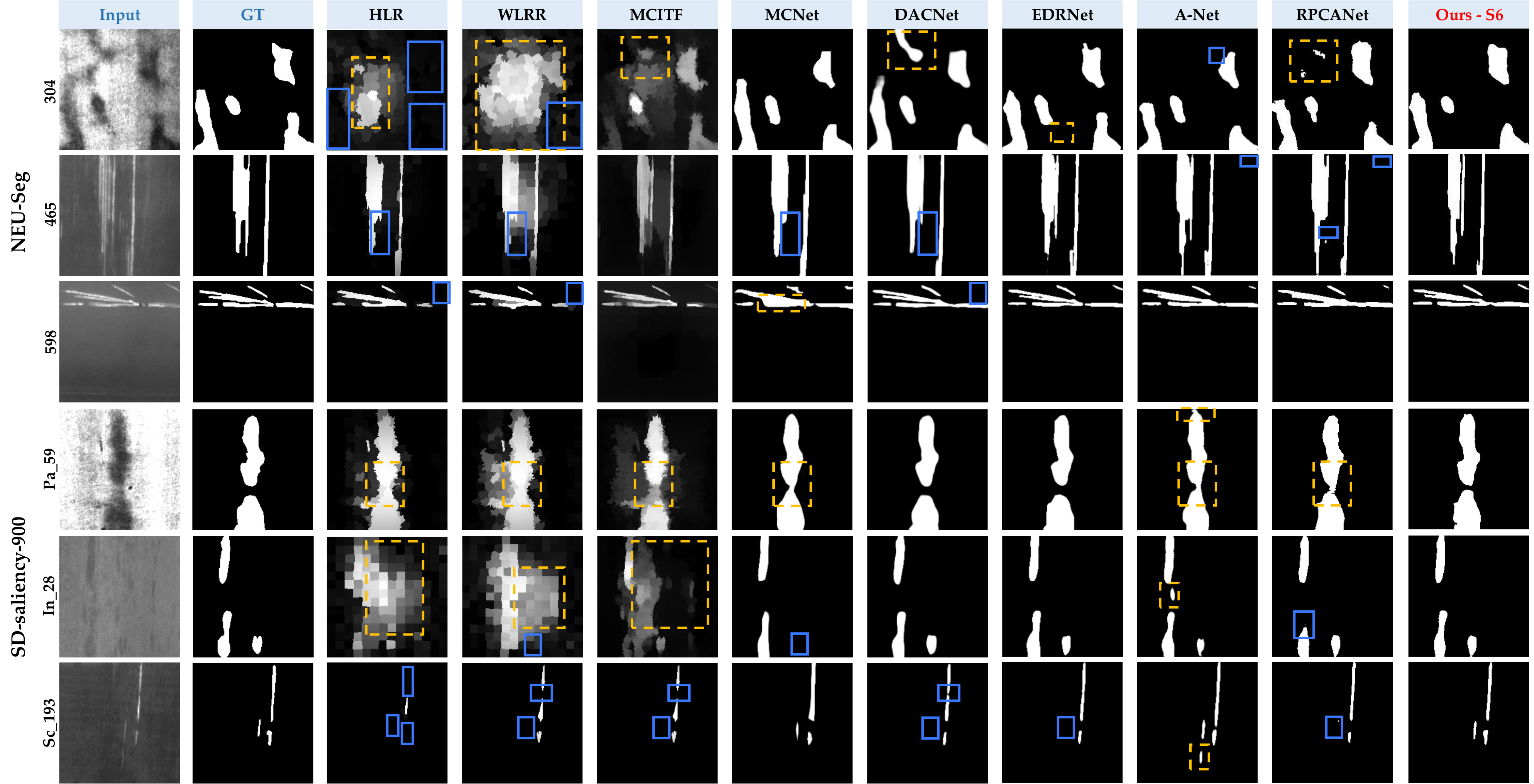}
    \caption{Visual comparisons of diverse defect detection techniques results from NEU-Seg \cite{dong-2019-neuseg} and SD-saliency-900 \cite{song-2020-edrnet},  with correctly identified targets, undetected targets, and false positives delineated by \textcolor{blue!70}{blue} and \textcolor{orange!50}{yellow} boxes, respectively. [Zoom in for a better view]}
    \label{fig:defect}
\vspace{-0.2cm}
\end{figure*}
\begin{table*}[!ht]
\setlength{\abovecaptionskip}{0pt} 
\setlength{\belowcaptionskip}{0pt} 
\caption{Performance metrics including S-measure (\%), MAE, IoU (\%), $F_1$ (\%), and runtime are evaluated for various methods on datasets NEU-Seg \cite{dong-2019-neuseg} and SD-saliency-900  \cite{song-2020-edrnet}. Parameter statistics Params (M) for data-driven approaches are encapsulated within the second-to-last column. [Zoom in for a better view]}
\centering
\begin{threeparttable}
\scriptsize{
\begin{tabular}{>{\raggedleft\arraybackslash}p{2.4cm}II>{\centering}p{1.7cm}I>{\centering}p{0.7cm}>{\centering}p{0.9cm}>{\centering}p{0.8cm}>{\centering}p{0.8cm}I>{\centering}p{0.8cm}>{\centering}p{0.9cm}>{\centering}p{0.8cm}>{\centering}p{0.8cm}I>{\centering}p{0.8cm}c}
\hline\thickhline
\rowcolor{gray!20} &  & \multicolumn{4}{cI}{\textbf{NEU-Seg \cite{dong-2019-neuseg}}} & \multicolumn{4}{cI}{\textbf{SD-saliency-900 \cite{song-2020-edrnet}}} & \textbf{Params} &\textbf{Time (s)} \\ \cline{3-10}
 \rowcolor{gray!20} \multirow{-2}{*}{\textbf{Methods}} & \multirow{-2}{*}{\textbf{Venue}} & $S_{m}$ $\uparrow$ & MAE $\downarrow$ &  IoU $\uparrow$ & $F_1$ $\uparrow$ &  $S_{m}$ $\uparrow$ & MAE $\downarrow$ &IoU $\uparrow$ & $F_1$ $\uparrow$ &(M) & CPU/GPU\\ 
\hline\hline
\multicolumn{12}{l}{\textcolor{gray!60}{\textit{Model-Based Methods}}} \\
WLRR \cite{tang-2016-WLRR}&  SPL$^{\textbf{17}}$  & 43.79   & 0.1249  &  27.62&  43.29   &44.13 &0.1181&  25.92  & 41.17 &  -  & 0.3324  \\ 
\rowcolor{gray!10} HLR  \cite{zheng-2020-HLR}& Neurocomput.$^{\textbf{20}}$  & 43.79  &  0.1247  &29.75 & 45.86 &44.13 & 0.1179&  27.19  & 42.75 & -  & 0.2047  \\ 
MCITF  \cite{song-2020-MCITF}&OLE$^{\textbf{20}}$  & 43.86 &0.1250 &39.79  &   56.93  & 44.22  & 0.1182 &  40.86   &   58.02& - & 17.8710  \\ 
 \hdashline
\multicolumn{12}{l}{\textcolor{gray!60}{\textit{Deep Learning-Based Methods}}} \\
U-Net \cite{ronneberger-2015-unet} & MICCAI$^{\textbf{15}}$  &   85.43  & 0.0297  & 79.25    & 88.42   &  88.02  & 0.0203 &  84.82 &  91.78   & 34.526 & 0.0141  \\ 
\rowcolor{gray!10}U-Net++ \cite{zhou-2018-unetplusplus}  & DLMIA$^{\textbf{18}}$  &   85.62  &  0.0298 &  79.35   & 88.49    &  88.21  &  0.0200  & 85.04   &  91.92 & 2.294  & 0.0139  \\ 
Attention UNet \cite{oktay-2018-attnunet}  & arXiv$^{\textbf{18}}$  &   85.33  &  0.0305  &    78.80 & 88.15    &   87.70 & 0.0213&   84.10 &  91.36  & 2.184 & 0.0134  \\ 
\rowcolor{gray!10}ResUNet \cite{diakogiannis-2020-resunet} &  ISPRS$^{\textbf{20}}$ &  85.28    &  0.0304 &   79.09  & 88.32    &   89.00 &  0.0183 &  85.27  &   92.05 & 0.905  & 0.0117  \\ 
U2Net  \cite{qin-2020-u2net}& PR$^{\textbf{20}}$ &  \underline{86.44}   &  0.0289  &   78.64  & 88.04   &  88.87 &  0.0188 & 86.08   &  92.52 & 1.130 & 0.0344  \\ 
\rowcolor{gray!10}EDRNet \cite{song-2020-edrnet} & TIM$^{\textbf{20}}$ &  85.88   &    0.0283  &  \textbf{80.45}   &\textbf{89.17}   & 89.13   &\underline{0.0180} &  \underline{86.66} & \underline{92.85} & 39.307  & 0.0487  \\
MCNet \cite{zhang-2021-mcnet}& TIM$^{\textbf{21}}$  &  85.95   &  0.0301 &  79.06   & 88.30    &  85.38  & 0.0240 & 82.28   & 90.28 & 38.437  & 0.0604  \\
\rowcolor{gray!10}DACNet \cite{zhou-2022-dacnet}& TIM$^{\textbf{22}}$  & 85.88      &  0.0301 & 79.61 &  88.65   &  \textbf{90.54}  &  \textbf{0.0172} & \textbf{87.58} &  \textbf{93.38} & 98.390 & 0.0444  \\ 
A-Net \cite{chen-2024-anet} & TIM$^{\textbf{24}}$ &   85.40  &0.0297 & 79.26   & 88.43      &  85.90  &  0.0247&   81.96 &  90.08 & 0.390  & 0.0265  \\ 
\hline
\multicolumn{12}{l}{\textcolor{gray!60}{\textit{Deep Unfolding-Based Methods}}} \\
RPCANet \cite{wu-2024-rpcanet} & WACV$^{\textbf{24}}$  &  84.19    &  0.0346 &  76.22   & 86.51   &  87.92  &   0.0243 & 81.28   &  89.68  & 0.680 & 0.0200  \\ 
\rowcolor{RoyalBlue!5} \textbf{Ours - S3}& - &  85.31  & 0.0295 &  78.99{\tiny{\textcolor{VioletRed}{+2.77}}}  &88.26{\tiny{\textcolor{VioletRed}{+1.65}}}  &87.76 &0.0235&  81.81{\tiny{\textcolor{VioletRed}{+0.53}}} & 90.00{\tiny{\textcolor{VioletRed}{+0.32}}}  & 1.435 & 0.0247  \\
\rowcolor{RoyalBlue!12} \textbf{Ours - S6}& - & 85.75    &\textbf{0.0273} &\underline{80.32}{\tiny{\textcolor{VioletRed}{+4.10}}}    &\underline{89.09}{\tiny{\textcolor{VioletRed}{+2.58}}}    &  88.45  &0.0215 &83.30{\tiny{\textcolor{VioletRed}{+2.02}}}    &90.89{\tiny{\textcolor{VioletRed}{+1.21}}} & 2.915 & 0.0357  \\
\rowcolor{RoyalBlue!25} \textbf{Ours - S9}& - & \textbf{86.84} & \underline{0.0276} & 80.10{\tiny{\textcolor{VioletRed}{+3.88}}}  &88.95{\tiny{\textcolor{VioletRed}{+2.44}}} &\underline{89.28} &0.0208 &  83.80{\tiny{\textcolor{VioletRed}{+2.52}}} & 91.18{\tiny{\textcolor{VioletRed}{+1.50}}}   & 4.396 & 0.0499   \\ 
\hline\thickhline
\end{tabular}}
\begin{tablenotes} 
\tiny
\item[*]Due to memory constraints, we adopt the initial channel number of 16 for UNet++, ResUNet, and Attention UNet.
\end{tablenotes}
\end{threeparttable}
\label{tab:defect}
\vspace{-0.4cm}
\end{table*}

\subsubsection{Comparison on Vessel Segmentation Task}
In the comparison section for the vessel segmentation task, we benchmark our approach against several state-of-the-art vessel segmentation methods. We incorporate baselines: U-Net \cite{ronneberger-2015-unet}, U-Net++ \cite{zhou-2018-unetplusplus}, ResUNet \cite{li-2019-resunet}, Attention UNet \cite{oktay-2018-attnunet}, CS-Net \cite{mou-2019-csnet}, FR-UNet \cite{liu-2022-frunet}, MCDAU-Net \cite{zhou-2023-MCDAUNet}, PVT-GCASCADE \cite{rahman-2024-gcascade}, and RPCANet \cite{wu-2024-rpcanet}. We retrained all the listed DL methods with 256$\times$256 crop size and tested for full size for a fair comparison.

\noindent\textbf{Visual Comparison:} 
Fig.~\ref{fig:medical} presents segmentation results from three datasets under challenging conditions. Methods such as ResUNet, as demonstrated on the DRIVE and CHASE\_DB1 datasets, tend to overlook many microvascular structures. Similarly, the baseline RPCANet exhibits intermittent vessel predictions—particularly evident in CHASE\_DB1—indicating its limited capability in capturing local features. A comparable shortcoming is observed with CS-Net on the STARE dataset. In contrast, our proposed RPCANet\textsuperscript{++} effectively captures finer vessels and boundary details, thereby preserving both global and local vascular structures.

\noindent \textbf{Numerical Comparison:}  
Table.~\ref{tab:medical} reports the performance of various segmentation techniques on the three vessel segmentation datasets using six different metrics. Among the deep learning-based methods, conventional networks like U-Net, U-Net++, and ResUNet achieve competitive results, but they come with large parameter counts (e.g., U-Net has 34.526 M parameters) and varying sensitivity to fine structures in medical images. In contrast, the deep unfolding-based methods show an attractive balance between performance and compactness. Our earlier RPCANet, with only 0.680 M parameters, already demonstrates respectable performance. The enhanced RPCANet$^{++}$ series further elevates these results. Specifically, the \textbf{RPCANet$^{++}$-S3} gains modest improvements in $F_1$ (82.24\% on DRIVE) and IoU (69.86\% on DRIVE) compared to RPCANet. The six-stage RPCANet$^{++}$ model achieves the best overall performance on DRIVE with an Acc of 96.91\% and leads on CHASE\_DB1 with notable enhancements of +2.07\% in $F_1$ and +2.81\% in IoU. While the \textbf{RPCANet$^{++}$-S9} shows a slight drop in some metrics, it still outperforms most conventional approaches, demonstrating the robustness of our framework.

\noindent \textbf{Computational Efficiency:} Despite a moderate increase in parameters (from 1.435 M for S3 to 4.396 M for S9), the RPCANet$^{++}$ series maintains competitive runtimes (typically, our \textbf{S3} is under 0.027 s), making it relatively efficient. 

Overall, these results underscore the merits of the RPCANet$^{++}$ series, which consistently achieves state-of-the-art segmentation performance and remarkable computational efficiency across diverse medical imaging datasets.

\subsubsection{Comparison on Defect Detection Task}
In the comparison section, we benchmark our approach against several state-of-the-art DD methodologies. For model-driven baselines, we incorporate: WLRR \cite{tang-2016-WLRR}, HLR \cite{zheng-2020-HLR}, and MCITF \cite{song-2020-MCITF}. For data-driven framework, we evaluate frameworks U-Net\cite{ronneberger-2015-unet}, U-Net++ \cite{zhou-2018-unetplusplus}, Attention UNet \cite{oktay-2018-attnunet}, ResUNet \cite{li-2019-resunet},  U2Net \cite{qin-2020-u2net}, EDRNet \cite{song-2020-edrnet}, MCNet \cite{zhang-2021-mcnet}, DACNet \cite{zhou-2022-dacnet}, A-Net \cite{chen-2024-anet}, and the DUN-based RPCANet \cite{wu-2024-rpcanet}. Despite RPCANet, we retrained listed DL methods referring to their public codes.

\noindent \textbf{Visual Comparison:}  Fig.~\ref{fig:defect} presents mask segmentation results from two defect datasets under challenging conditions. Many optimization-based methods, such as MCITF, effectively capture salient features, although they sometimes overlook small defects (e.g., HLR in Sc\_193). In contrast, deep learning methods can reliably predict defect shapes by leveraging ground truth labels; however, unexpected adhesion issues can occur when capturing fine details (e.g., Pa\_59 in MCNet and A-Net). Notably, RPCANet\textsuperscript{++} consistently detects all defect targets with high shape accuracy. It delineates the background from the target even when boundaries are blurred (e.g., 598 in NEU-Seg and In\_28 from SD-saliency-900), thereby demonstrating its superior capability in preserving shape details in cluttered backgrounds.

\noindent \textbf{Numerical Comparison:}  
Table.~\ref{tab:defect} illustrates the performance on two defect detection datasets. Among the deep unfolding-based methods, our baseline RPCANet achieves an IoU of 76.22\% and an $F_1$ score of 86.51\% on NEU-Seg, and 81.28\% IoU and 89.68\% $F_1$ on SD-saliency-900. RPCANet$^{++}$ series consistently improves upon these results. For instance, \textbf{S3} variant raises the IoU on NEU-Seg by 2.77\% (to 78.99\%) and boosts the $F_1$ score by 1.65\%, while yielding modest gains on SD-saliency-900. \textbf{RPCANet$^{++}$-S6} further enhances performance, achieving an IoU of 80.32\% (+4.10\%) and an $F_1$ score of 89.09\% on NEU-Seg, with similar improvements on SD-saliency-900. Although the nine-stage RPCANet$^{++}$ demonstrates comparable trends with a satisfying $S_m$ value, its overall gains are slightly lower than those achieved by the six-stage RPCANet$^{++}$.

\noindent\textbf{Computational Efficiency:} RPCANet$^{++}$ achieves strong performance with 2.915M parameters (\textbf{S6}), significantly fewer than DACNet’s 98.390M. It also offers faster inference—0.0247s (\textbf{S3}) vs. 0.0604s (MCNet) and 0.0487s (EDRNet)—highlighting its efficiency–effectiveness balance for defect detection.

Notably, in the defect detection task, the RPCANet$^{++}$ series does not achieve the pronounced advantage observed in other scenarios (e.g., VS and IRSTD). A plausible explanation is that the defects in these datasets are typically larger and less sparse, making the sparse representation assumption—central to our framework—less effective when target regions occupy a substantial portion of the image.

\begin{figure}[t]
\setlength{\abovecaptionskip}{1pt} 
\setlength{\belowcaptionskip}{1pt} 
	\begin{minipage}{0.47\linewidth}
		\centerline{\includegraphics[width=\textwidth]{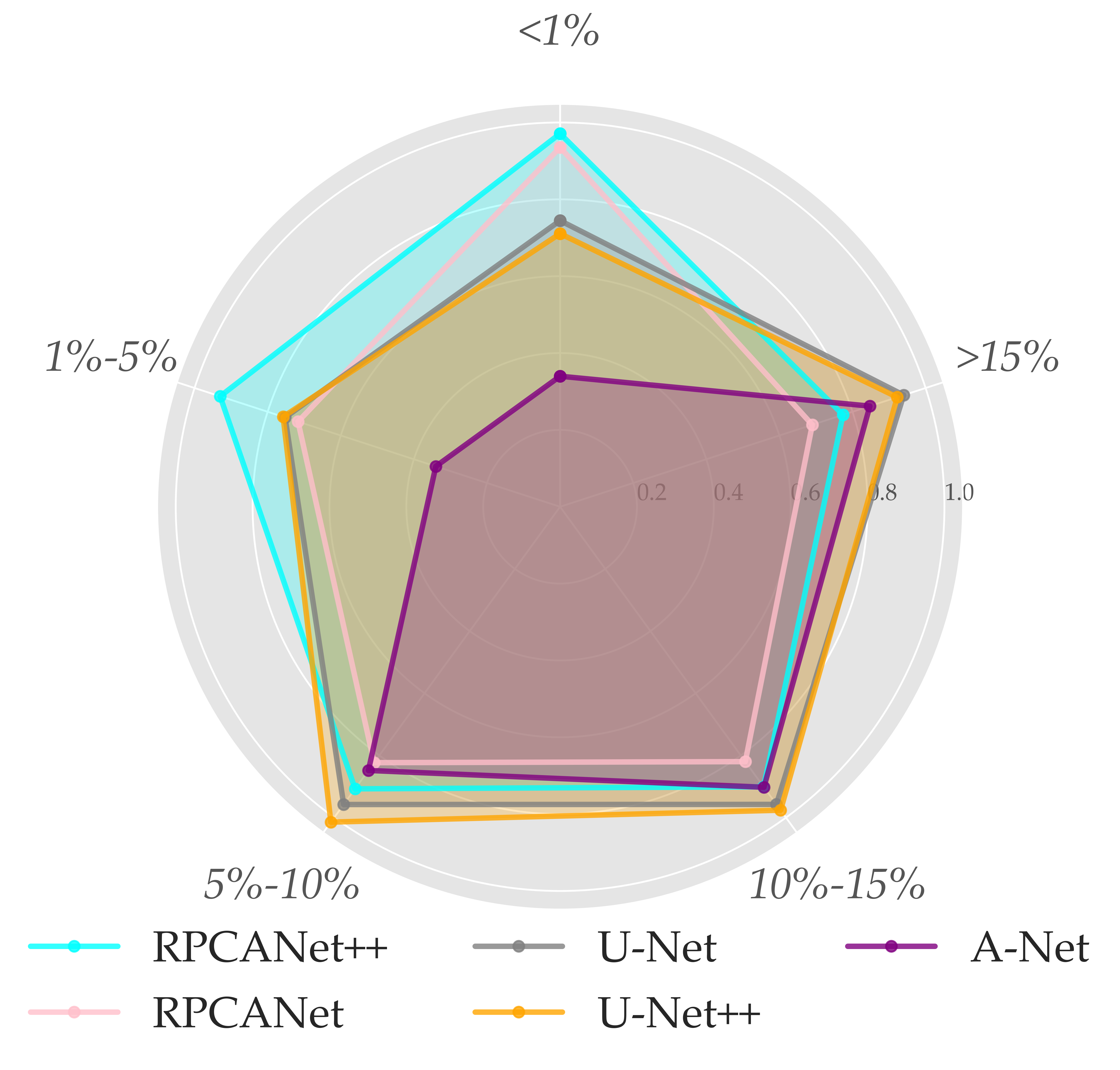}}
\subcaption{IoU Performance}
	\end{minipage}
    \begin{minipage}{0.47\linewidth}
		\centerline{\includegraphics[width=\textwidth]{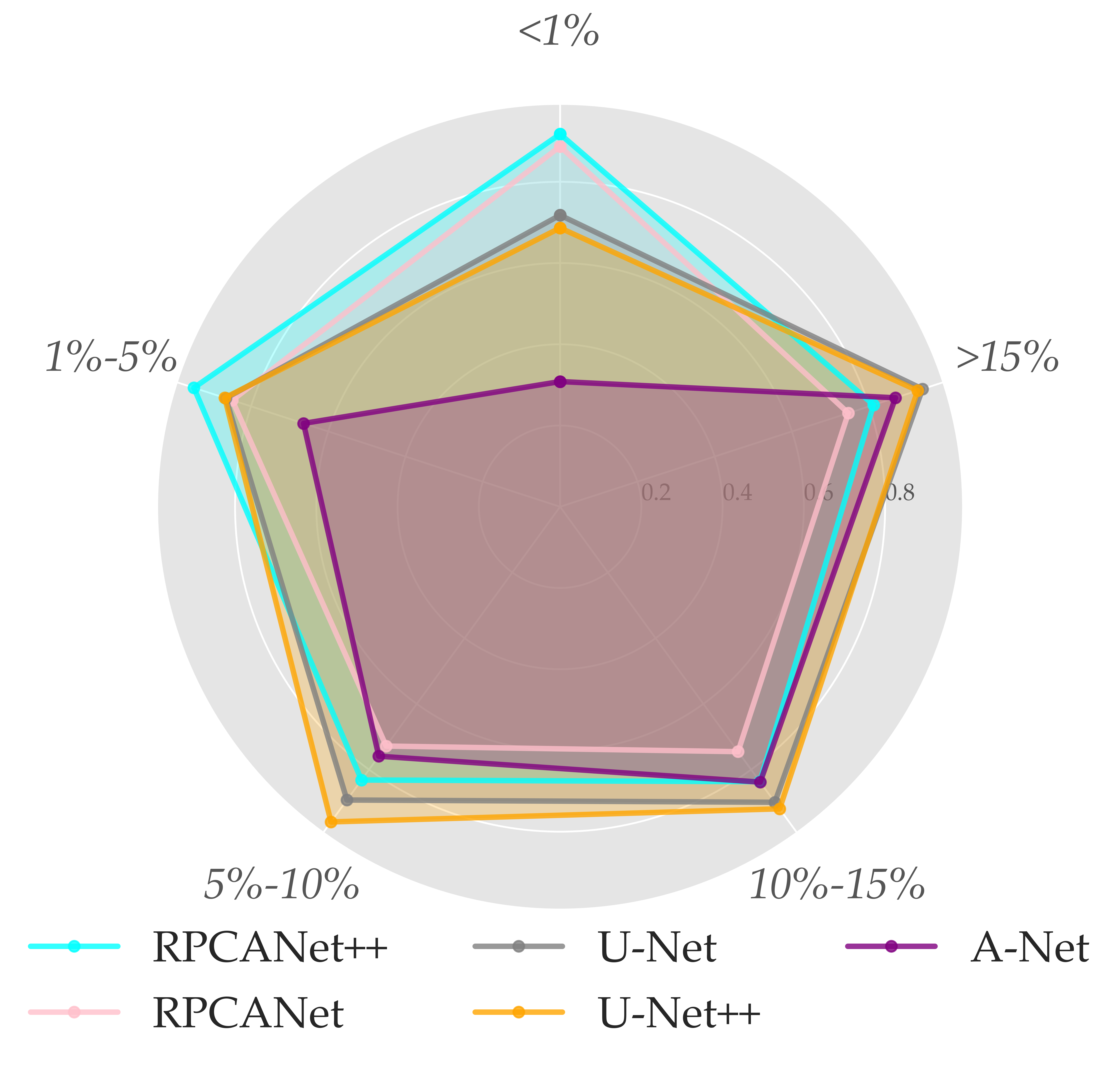}}
	\subcaption{$F_1$ Performance}
	\end{minipage}
	\caption{Performance comparison of RPCANet$^{++}$, RPCANet, U-Net, and A-Net across varying target area distributions in the SD-saliency-900 dataset \cite{song-2020-edrnet} for the \textbf{DD} task. (Target size distribution: $<$1\%—106 images (11.78\%), 1\%-5\%—305 images (33.89\%), 5\%-10\%—238 images (26.44\%), 10\%-15\%—99 images (11.00\%), $>$15\%—152 images (16.89\%)).}
	\label{fig_dis}
\end{figure}
\begin{figure}[t]
\setlength{\abovecaptionskip}{0pt}
\setlength{\belowcaptionskip}{0pt}
    \centering
    \includegraphics[width=1\linewidth]{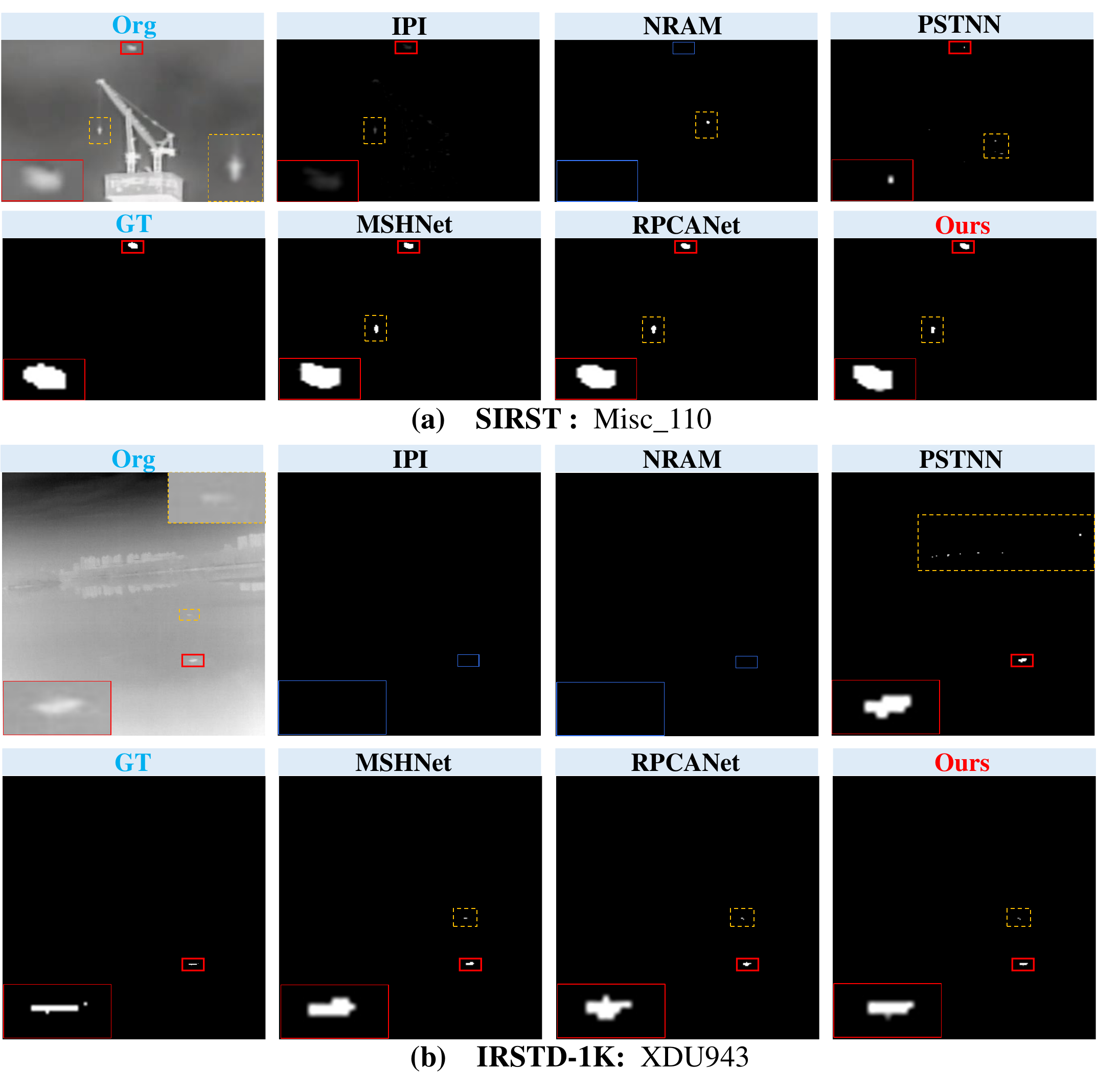}
    \caption{Failure examples that both compared methods and RPCANet$^{++}$ has encountered in  SIRST \cite{dai-2021-acm} and IRSTD-1K \cite{zhang-2022-isnet} in \textbf{IRSTD} tasks.}
    \label{fig:fail}
\end{figure}

\subsection{Limitations and Future Work}
In this section, we discuss the limitations and corresponding solutions or future work of RPCANet$^{++}$.

\noindent\textbf{1) Focus on sparse object segmentation:} As shown in Fig.~\ref{fig_dis}, we performed a detailed analysis of target area distributions and their influence on network performance. The results reveal that the RPCANet family excels at segmenting extremely small objects (target area $<$ 1\%). However, as object size increases—especially beyond 15\%—both RPCANet variants exhibit diminished competitiveness compared to U-Net and its derivatives, such as A-Net. These findings highlight the strength of RPCANet-based approaches in handling sparse or tiny object segmentation. Future work will explore applying the algorithm to scenarios where target size plays a critical role.

\noindent\textbf{2) Sparse-like target elimination}: On one hand, as presented in Fig. \ref{fig:fail}, due to the inherent sparsity of the target in the RPCA model, it may lead to falsely distinguishing target and sparse and target-like false alarms, which is also true for both data and model-driven methods. We plan to introduce \textbf{spatial-temporal} or \textbf{multi-modal} information in future work. For instance, temporal information could better assist the separation of moving targets and still prevent false alarms. Moreover, the possible injection of a \textbf{background awareness} (such as pre-segmentation priors \cite{xiao2024background}) can reduce the false alarm rate.

\noindent\textbf{3) Balancing parameter efficiency and interpretability:} RPCANet series adopts a fully convolutional architecture without downsampling to preserve interpretable operations (e.g., OEM construction). However, this design results in suboptimal inference speed. Compared to dense networks such as UIUNet and DACNet, RPCANet$^{++}$ achieves a relatively low parameter count. A promising future direction is to develop more efficient architectures that maintain mathematical rigor while

\section{Conclusion}
\label{sec:5}
\noindent In this work, we introduce RPCANet$^{++}$—an interpretable deep unfolding network that bridges optimization theory with deep learning for sparse object segmentation. By unfolding a relaxed RPCA model into three modular stages—background approximation (BAM), object extraction (OEM), and image restoration (IRM)—our approach combines theoretical rigor with neural efficiency. To counteract degradation of background features across unfolding stages, we incorporate memory augmentation (MAM), which enhances latent background estimation and improves sparse object detection. Furthermore, we boost target sensitivity by introducing a local contrast-driven prior (DCPM), extending the original RPCA formulation.

We validate the alignment of RPCANet$^{++}$ with RPCA principles using novel low-rankness and sparsity metrics that quantify stage-wise decomposition quality, supplemented by interpretability analyses. Our results confirm that RPCANet$^{++}$ iteratively learns low-rank backgrounds and sparse components following theoretical constraints, validating the architecture’s effectiveness. Extensive experiments across a range of sparse object segmentation tasks—including infrared small target detection (IRSTD), vessel segmentation (VS), and defect detection (DD)—demonstrate the superior performance of our method.

By systematically integrating model-driven constraints with data-driven optimization, the RPCANet series offers a principled and interpretable alternative to conventional black-box architectures. This work paves the way for more trustworthy segmentation frameworks and contributes a solid foundation for future research in interpretable deep learning.

\section*{Acknowledgments}
The authors acknowledge the insightful critiques from the editorial team and anonymous reviewers, which have substantially refined this manuscript.

\appendices
\setcounter{table}{0}   
\setcounter{figure}{0}
\setcounter{section}{0}
\setcounter{equation}{0}
\renewcommand{\thefigure}{A.\arabic{figure}}
\renewcommand{\thesection}{A.\arabic{section}}
\renewcommand{\theequation}{A.\arabic{equation}}
\section{}
\label{appendix}
In this section, the feasibility of Taylor's second-order expansion under the Lipschitz continuous condition is analyzed and derived.

\noindent
{\bf{Theorem 1.}} For a function $f(t)$, with its Lipschitz continuous gradient function $\nabla f(t)$ (i.e., $\forall {t_1},{t_2}:\left\| {\nabla f({t_1}) - \nabla f({t_2})} \right\| \leqslant L\left\| {{t_1} - {t_2}} \right\|$, where  $L$ is a constant), then $f(t)$ can be Taylor approximated at the a fix point $t_0$ by:
\begin{equation}
\setlength{\abovedisplayskip}{2pt}
\setlength{\belowdisplayskip}{2pt}
\begin{aligned}
\hat f(t,{t_0}) & \leqslant f({t_0}) + \left \langle {\nabla f({t_0}),t - {t_0}} \right\rangle  + \frac{L}{2}{\left\| {t - {t_0}} \right\|^2}\\
& \leftarrow \frac{L}{2}\left\| {t - {t_0} + \frac{1}{L}f({t_0})} \right\|^2 + C.
\end{aligned}
\end{equation}

\noindent
{\bf{Proof.}} For a given continuously differentiable function $\nabla f(t)$, the Taylor second-order expansion at the point $t_0$ is\footnote{Please refer to \url{http://www.seas.ucla.edu/~vandenbe/236C/lectures/gradient.pdf}, page 14-15.}:
\begin{equation}
\setlength{\abovedisplayskip}{2pt}
\setlength{\belowdisplayskip}{2pt}
\hat f(t,{t_0}) \leqslant f({t_0}) + \left \langle {\nabla f({t_0}),t - {t_0}} \right\rangle  + \frac{L}{2}{\left\| {t - {t_0}} \right\|^2}.
\end{equation}
This leads to the following derivation: 
\begin{equation}
\setlength{\abovedisplayskip}{2pt}
\setlength{\belowdisplayskip}{2pt}
\begin{aligned}
\hat f&(t,{t_0}) \leqslant f({t_0}) + \left \langle {\nabla f({t_0}),t - {t_0}} \right\rangle  + \frac{L}{2}{\left\| {t - {t_0}} \right\|^2} \\
&= \frac{L}{2} \left( {{\left\| {t - {t_0}} \right\|}^2} + 2\left\langle {\frac{1}{L}f({t_0}),t - {t_0}} \right\rangle 
 \right.\\
&~~~~\left. + {{\left\| {\frac{1}{L}f({t_0})} \right\|}^2} - {{\left\|{\frac{1}{L}f({t_0})} \right\|}^2} + \frac{2}{L}f({t_0}) \right)\\
& = \frac{L}{2}{\left\| {t - {t_0} + \frac{1}{L}f({t_0})} \right\|^2} - \frac{1}{{2L}}{\left\| {f({t_0})} \right\|^2} + f({t_0}).
\end{aligned}
\end{equation}
And in optimization problems, we often have a fixed point $t_0$, such that:
\begin{equation}
\setlength{\abovedisplayskip}{2pt}
\setlength{\belowdisplayskip}{2pt}
\hat f(t,{t_0})  \leftarrow \frac{L}{2}\left\| {t - {t_0} + \frac{1}{L}f({t_0})} \right\|^2 + C,
\end{equation}
where $L$ is a constant, $C =  - \frac{1}{{2{L}}}{\left\| \nabla{f({t_0})} \right\|^2} + f({t_0})$. 

Thus, Theorem 1 is proved.
\ifCLASSOPTIONcompsoc

\ifCLASSOPTIONcaptionsoff
  \newpage
\fi



%
\bibliographystyle{IEEEtran}
\bibliography{IEEEabrv,reference}

%

\newcommand{\addPhoto}[1]{\includegraphics[width=1in,height=1.15in,clip,keepaspectratio]{figures/biography/#1}}

\begin{IEEEbiography}[{\includegraphics[width=1in,height=1.15in,clip,keepaspectratio]{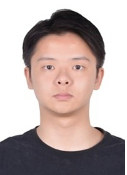}}]
{Fengyi Wu}
received his B.E. degree in Electronic Information Engineering from both the University of Electronic Science and Technology of China (UESTC) in 2021 and is currently chasing a PhD degree at the School of Information and Communication Engineering, UESTC. His current interests include computer vision, pattern recognition, deep unfolding, and interpretable object detection.
\end{IEEEbiography}

\begin{IEEEbiography}[{\includegraphics[width=1in,height=1.15in,clip,keepaspectratio]{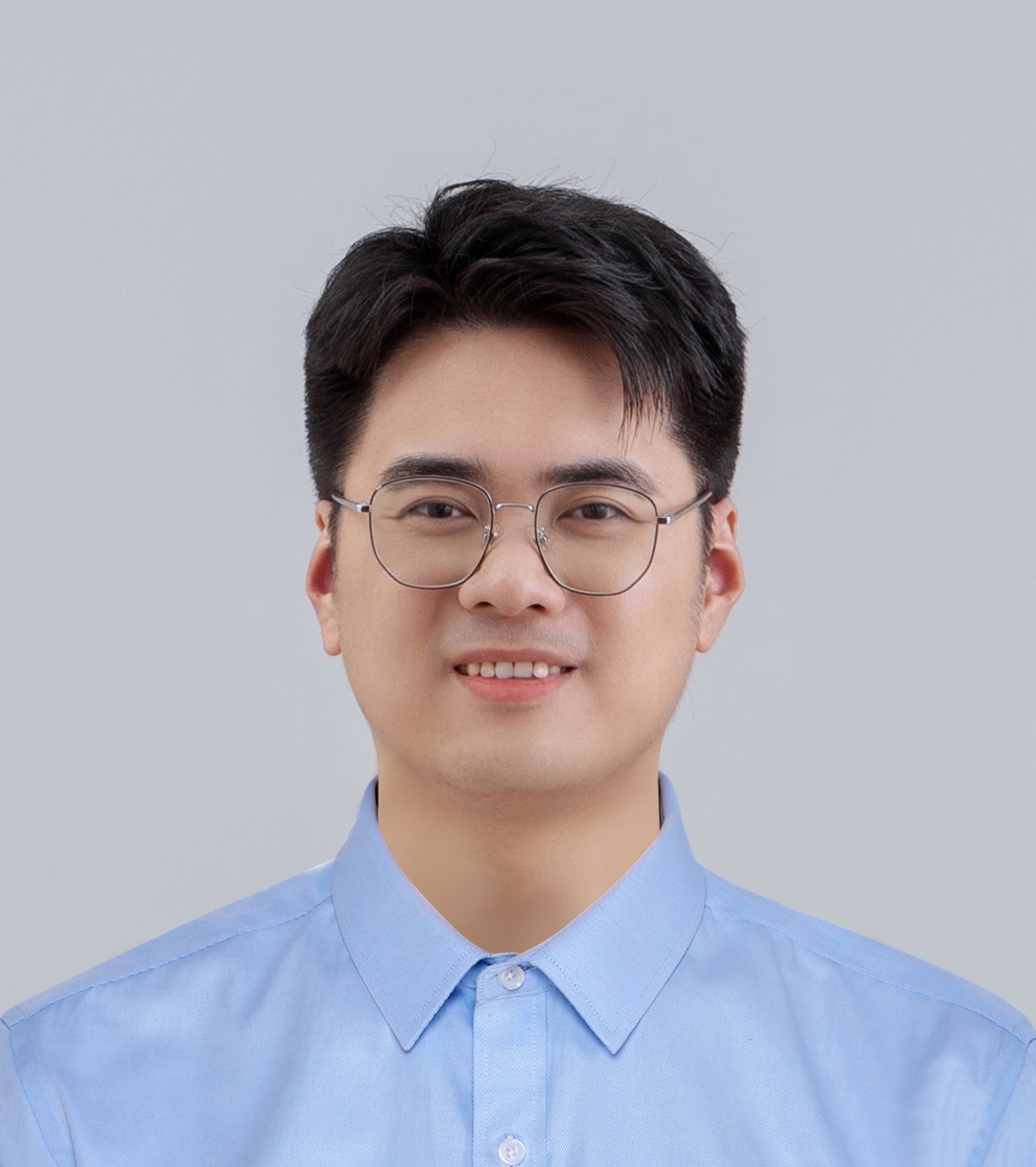}}]
{Yimian Dai}
(Member, IEEE) received the B.E. degree in information engineering and the Ph.D. degree in signal and information processing from Nanjing University of Aeronautics and Astronautics, Nanjing, China, in 2013 and 2020, respectively.
From 2021 to 2024, he was a Postdoctoral Researcher with the School of Computer Science and Engineering, Nanjing University of Science and Technology, Nanjing, China. 
He is currently an Associate Professor with the College of Computer Science, Nankai University, Tianjin, China.
His research interests include computer vision, deep learning, and their applications in remote sensing.
For more information, please visit the link (\href{https://yimian.grokcv.ai/}{https://yimian.grokcv.ai/}).
\end{IEEEbiography}

\begin{IEEEbiography}[{\includegraphics[width=1in,height=1.15in,clip,keepaspectratio]{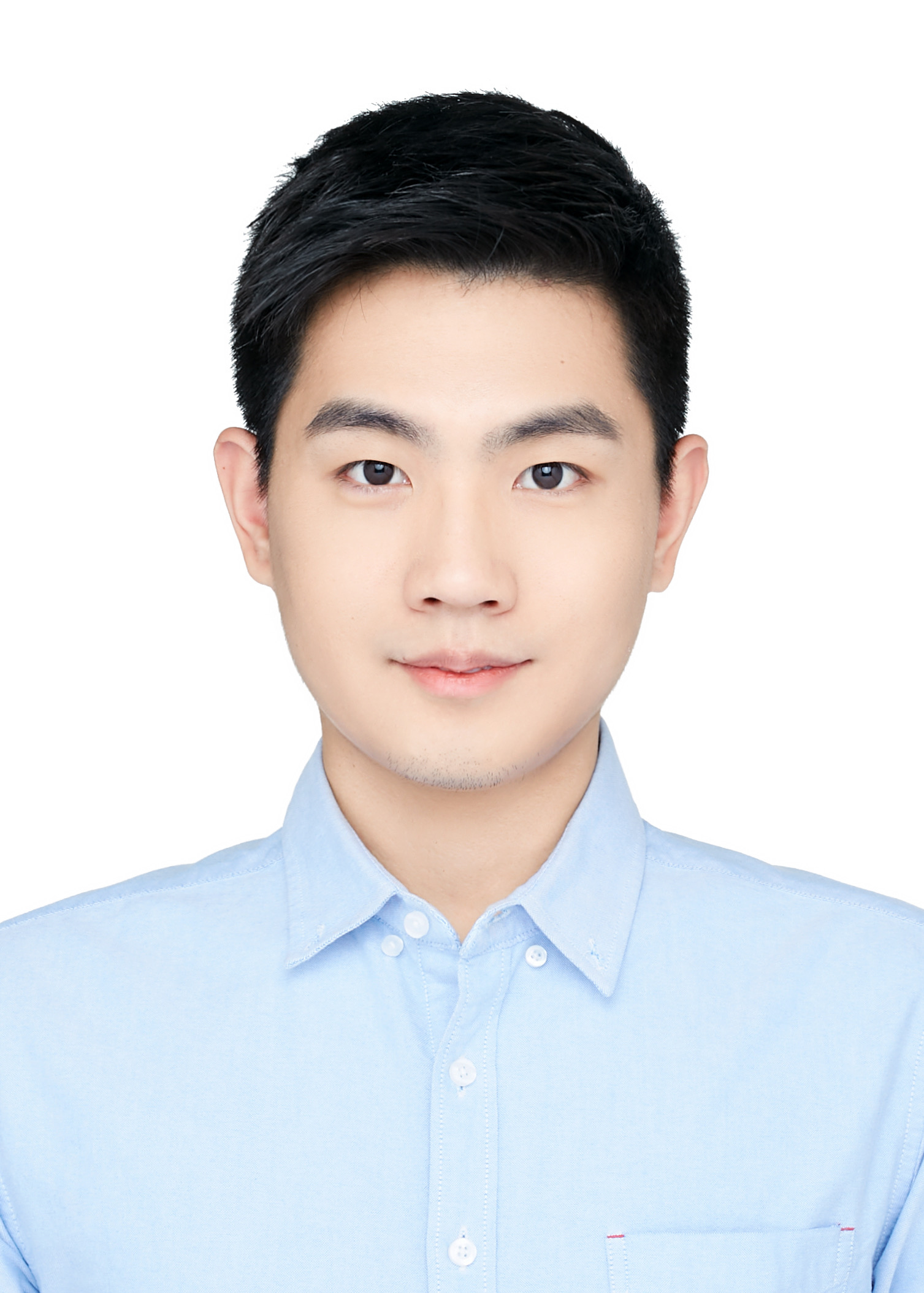}}]
{Tianfang Zhang}
received his PhD degree from the School of Information and Communication Engineering, University of Electronic Science and Technology of China (UESTC) in 2023. Currently, he is working as a postdoctoral fellow in the Department of Automation, Tsinghua University. His main research interests are computer vision, efficient vision transformers, and multi-modal models.
\end{IEEEbiography}

\begin{IEEEbiography}[{\includegraphics[width=1in,height=1.15in,clip,keepaspectratio]{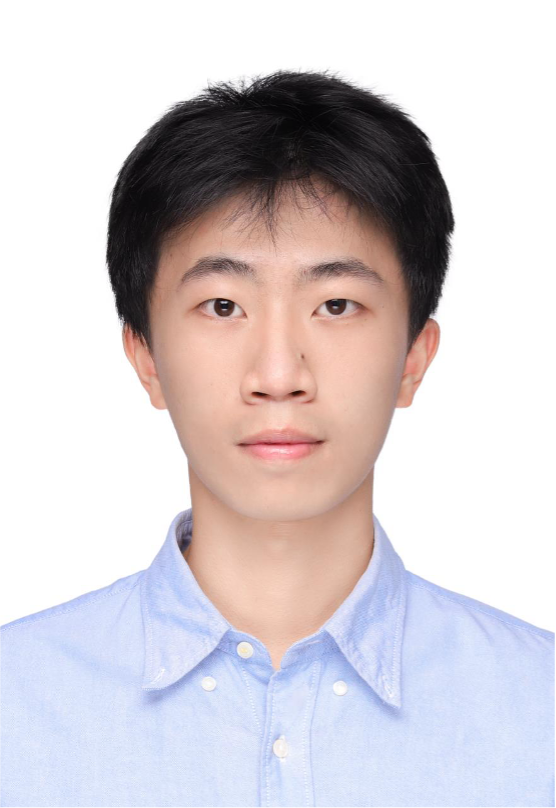}}]
{Yixuan Ding}
is currently pursuing a bachelor’s degree in communication engineering from the College of Glasgow, the University of Electronic Science and Technology of China (UESTC), Chengdu, China. His recent research interests include target detection, compressive sensing, and deep unfolding. 

\end{IEEEbiography}

\begin{IEEEbiography}[{\includegraphics[width=1in,height=1.15in,clip,keepaspectratio]{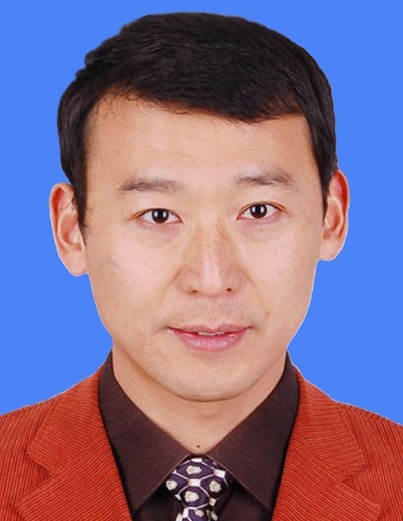}}]
{Jian Yang} received the PhD degree from Nanjing University of Science and Technology (NJUST) in 2002, majoring in pattern recognition and intelligence systems. From 2003 to 2007, he was a Postdoctoral Fellow at the University of Zaragoza, Hong Kong Polytechnic University and New Jersey Institute of Technology, respectively. From 2007 to present, he is a professor in the School of Computer Science and Technology of NJUST. Currently, he is also a visiting distinguished professor in the College of Computer Science of Nankai University. His papers have been cited over 50000 times in the Scholar Google. His research interests include pattern recognition and computer vision. Currently, he is/was an associate editor of Pattern Recognition, Pattern Recognition Letters, IEEE Trans. Neural Networks and Learning Systems, and Neurocomputing. He is a Fellow of IAPR. 
\end{IEEEbiography}

\begin{IEEEbiography}[{\includegraphics[width=1in,height=1.15in,clip,keepaspectratio]{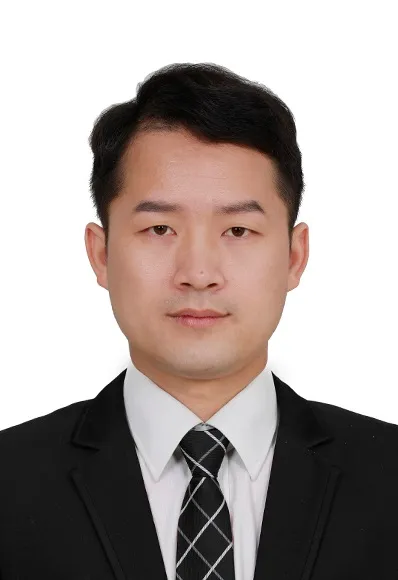}}]
{Ming-Ming Cheng} received his PhD degree from Tsinghua University in 2012.
Then, he did 2 years research fellow, with Prof. Philip Torr in Oxford.
He is now a professor at Nankai University, leading the Media Computing Lab.
His research interests include computer graphics, computer vision, and image processing. 
He received research awards, including the National Science Fund for Distinguished Young Scholars and the ACM China Rising Star Award.
He is on the editorial boards of IEEE TPAMI and IEEE TIP.
\end{IEEEbiography}

\begin{IEEEbiography}[{\includegraphics[width=1in,height=1.15in,clip,keepaspectratio]{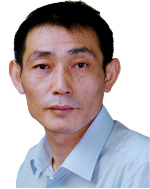}}]
{Zhenming Peng}
(Member, IEEE) received a Ph.D. degree in geodetection and information technology from the Chengdu University of Technology, Chengdu, China, in 2001.
From 2001 to 2003, he was a Post-Doctoral Researcher with the Institute of Optics and Electronics, Chinese Academy of Sciences, Chengdu. He is a Professor with the University of Electronic Science and Technology of China, Chengdu. His research interests include image processing, signal processing, and target recognition and tracking.
\end{IEEEbiography}
\end{document}